\documentclass[opre]{informs3} 

\OneAndAHalfSpacedXI 

\usepackage{endnotes}

\renewenvironment{proof}[1][Proof]{%
\par\noindent{\itshape #1.} \ignorespaces
}{%
\hfill\Halmos\par
}

\usepackage{booktabs}
\usepackage{tikz}

\usetikzlibrary{arrows.meta,shadows,positioning,automata, arrows,shapes.geometric}
\usepackage{subcaption}
\usepackage[utf8]{inputenc}
\usepackage{amsfonts}
\usepackage{amsmath}
\usepackage{xspace} 

\usepackage{thmtools}

\usepackage{mathtools}
\usepackage{tikz}
\usepackage{graphicx}
\usepackage{breqn}
\IfPackageLoadedTF{hyperref}{}{\usepackage[colorlinks=true,breaklinks=true,bookmarks=true,urlcolor=magenta,citecolor=magenta,linkcolor=magenta,bookmarksopen=false,draft=false]{hyperref}}
\usepackage{color}
\usepackage{comment}
\usepackage{enumerate}
\usepackage{url}
\usepackage{placeins}

\usepackage{algorithm}
\usepackage{algpseudocode}
\usepackage[capitalize]{cleveref}
\usepackage{enumitem}
\usepackage[normalem]{ulem}
\usepackage{multirow}
\usepackage{blkarray, bigstrut}
\usepackage{xparse}
\usepackage[version=4]{mhchem}
\usepackage[toc,page]{appendix}
\usepackage{dsfont}
\usepackage{graphicx}
\usepackage{rotating}
\usepackage{enumitem} 

\newenvironment{rproof}{ \ifdefined\opre \proof{\it Proof.}
	\else \begin{proof} \fi }{ \ifdefined\opre 
		\Halmos \endproof \else \end{proof} \fi  }

\usepackage{color-edits}
\addauthor{jw}{olive}
\addauthor{srs}{orange}
\addauthor{ds}{red}
\addauthor{mw}{red}
\addauthor{rev}{blue}
\addauthor{min}{blue}

\usepackage{xcolor}
\definecolor{edit}{rgb}{0,0,0} 

\ifdefined \argmin
\else \DeclareMathOperator*{\argmin}{arg\,min}
\fi
\ifdefined \argmax
\else \DeclareMathOperator*{\argmax}{arg\,max}
\fi
\usepackage{Jia_macros}

\usepackage{algpseudocode}
\usepackage{tikz}

\usepackage{natbib}
\bibpunct[, ]{(}{)}{,}{a}{}{,}%
\def\bibfont{\small}%

\TheoremsNumberedThrough     

\ECRepeatTheorems

\EquationsNumberedThrough    

\begin{document}

\RUNAUTHOR{Wan et al.}

\RUNTITLE{Exploiting Exogenous Structure for RL}

\TITLE{Exploiting Exogenous Structure for
Sample-Efficient Reinforcement Learning}

\ARTICLEAUTHORS{%

\AUTHOR{Jia Wan}
\AFF{Laboratory for Information and Decision Systems,
Massachusetts Institute of Technology, \EMAIL{jiawan@mit.edu}}

\AUTHOR{Sean R. Sinclair}
\AFF{Department of Industrial Engineering and Management Sciences,
Northwestern University, \EMAIL{sean.sinclair@northwestern.edu}}

\AUTHOR{Devavrat Shah, Martin J. Wainwright}
\AFF{Laboratory for Information and Decision Systems,
Massachusetts Institute of Technology, \EMAIL{\{devavrat,mjwain\}@mit.edu}}

} 

\ABSTRACT{We study a structured class of Markov Decision
Processes, known as Exo-MDPs, in which the state space is partitioned into exogenous and endogenous components. Exogenous states evolve stochastically, independent of the agent's actions, while endogenous states evolve deterministically based on both state components and actions. Exo-MDPs capture many operations research settings, including inventory control, resource management, and ride-sharing.  Our first contribution is structural: we establish a representational equivalence between discrete MDPs, Exo-MDPs, and discrete linear mixture MDPs.  Our second contribution is statistical: we characterize the minimax regret of learning in Exo-MDPs when the effective dimension $r$ is small relative to the endogenous state and action spaces. When the exogenous states are unobserved, we prove matching upper and lower regret bounds of order $\tilde{\Theta}(H r \sqrt{K})$ over $K$ episodes of horizon $H$, where $r$ is the effective dimension of the Exo-MDP. When exogenous states are observed, the minimax regret improves to $\tilde{\Theta}(H\sqrt{ r K})$, revealing a $\tilde{\Theta}(\sqrt{r})$ statistical gap due to observation of the exogenous states.  These results show that Exo-MDPs decouple sample complexity from the action and endogenous state space. We validate these insights with experiments on inventory control and resource allocation.
}


\maketitle

\section{Introduction}
\label{sec:introduction}

Markov decision processes provide a canonical framework for sequential decision-making under uncertainty, and reinforcement learning (RL) provides data-driven approaches for estimating near-optimal policies.
The past few decades have witnessed tremendous empirical success, notably in ``data-rich'' areas such as competitive game-playing~\citep{silver2016mastering}, computational advertising~\citep{zhao2019deep}, robotics~\citep{kober2013reinforcement}, and human-guided training of large language models~\citep{ouyang2022training}.  
This success relies on the availability of massive datasets, either due to large amounts of pre-collected data or via access to simulators for generating synthetic trajectories.  In contrast, there are other application domains notorious for being ``data-poor'', including finance~\citep{rao2022foundations}, resource allocation~\citep{hadary_protean_2020}, inventory control~\citep{madeka2022deep,liu2024reinforcement,qin2023sailing}, supply chain management~\citep{rolf2023review}, and ridesharing systems~\citep{dai2021queueing}.  In these settings, data is scarce either because new samples are costly to obtain, or because high-fidelity simulators are unavailable.  Without additional structure, information-theoretic lower bounds dictate that learning near-optimal policies requires a sample size that scales with the size of the state-action space.  For many domains of interest in operations research, this scaling is prohibitively expensive.  
{For example, the state space in resource allocation grows exponentially with the number of resources, while in inventory control it grows exponentially with the lead time, i.e. the delay between placing and receiving an order~\citep{agrawal2022learning}.}
Therefore, identifying and exploiting domain-specific structure is essential for data-efficient reinforcement learning.


{Motivated by this challenge, we study the class of \emph{Exogenous Markov Decision Processes}, as studied in a line of recent and on-going work~\citep{dietterich_discovering_2018,efroni2022sample,powell2022reinforcement,sinclair2023hindsight,feng2021scalable,liang2026pure}.}
The Exo-MDP class is defined by a decomposition of the state space into \emph{exogenous} and \emph{endogenous} components.  Exogenous states evolve stochastically, but do \emph{not} depend on the agent's actions, while endogenous states evolve as a purely deterministic function of the past state and action.  Given this decomposition, all uncertainty is isolated to the exogenous process, enabling tractable analysis of regret.  Exo-MDPs capture a broad range of operations research problems, including inventory control, portfolio optimization, and resource allocation, where system dynamics are known and stochasticity arises primarily from external uncertainty.

{\paragraph{An example.} Let us consider a concrete example,\footnote{Here we provide a high-level description; see~\Cref{sec:exo_mdp_setup}, for more detailed discussion, and~\Cref{sec:experiments} for numerical results.} the \emph{newsvendor model} for inventory control~\citep{goldberg_survey_2021}.}  A vendor wishes to control their inventory, which is the primary endogenous state. The external demand, which is not controllable by actions, represents the exogenous state. The action or control variable corresponds to purchase orders: how much new inventory should be purchased and when? The inventory evolves according to known deterministic ``queuing'' dynamics: new inventory equals past inventory minus demand (if it can be satisfied) plus new inventory purchased, based on some lead time $L \geq 1$ steps ago. In order to ensure Markovian dynamics, the endogenous state must include past purchase orders over the previous $L$ steps, so the endogenous state grows exponentially with the lead time $L$.

In real-world applications of this model, observations are sparse and limited. Any given vendor has relatively few historical trajectories, and with at best partial observation of the external demand~\citep{hssaine2024data}. 
{At the same time, this problem is naturally cast as an Exo-MDP.  All stochasticity arises from the external demand, which is low-dimensional compared to the exponentially large endogenous state (inventory and outstanding orders).} 

For this and other models, Exo-MDPs hold promise for designing data-efficient algorithms, since they explicitly use the fact that all randomness is captured through the exogenous states~\citep{sinclair2023hindsight,mao2018variance}.  However, existing analyses crucially assume the exogenous variables are fully observed, an assumption that fails to hold in many applications.  For example, the inventory model discussed above typically involves lost sales, where the true exogenous demand is unobserved due to stockouts~\citep{madeka2022deep}. Similarly, ridesharing systems exhibit demand shortfalls when drivers are unavailable, resulting in users leaving the platform~\citep{dai2021queueing}.

With this motivation in place, we are led to the primary questions tackled in this work:
\emph{
\begin{enumerate} \item What is the statistical complexity
  of learning in Exo-MDPs?
\item How is this complexity influenced by whether or not the
  exogenous states are observed?
\item
How can we design algorithms with near-optimal performance?
\end{enumerate}
}
\noindent These questions are particularly salient when the exogenous state is of ``low-complexity'' compared to the endogenous state.  This property is common to many application areas of interest, including inventory control, cloud compute resource scheduling, and matching in ride-sharing platforms, among others.

\subsection{Contributions}
Let us briefly summarize the main contributions of this work.
\begin{table}[t!]
\centering
\begin{center}
\resizebox{\columnwidth}{!}{
\begin{tabular}{||c|c|c|c||}
\hline
\textbf{Setting} & \textbf{Algorithm / Result} & \textbf{Section} & \textbf{Regret Bound} \\ [0.5ex]
\hline
\multirow{2}{*}{Unobserved Exogenous States}
& Upper bound (\HFUCRL) & \cref{sec:unobs_upper_bound} & $\tilde{O}\!\left(H r \sqrt{K}\right)$ \\
& Lower bound & \cref{sec:unobs_lower_bound} & $\Omega\!\left(H r \sqrt{K}\right)$ \\
\hline
\multirow{2}{*}{Observed Exogenous States}
& Upper bound (\PlugIn) & \cref{sec:obs_upper_bound} & $\tilde{O}\!\left(H\sqrt{rK}\right)$ \\
& Lower bound & \cref{sec:obs_lower_bound} & $\Omega\!\left(H\sqrt{rK}\right)$ \\
\hline
General MDPs (baseline)
& Lower bound~\citep{domingues2021episodic} & --- & $\Omega\!\left(H \sqrt{|\StateSpace||\ActionSpace|K}\right)$ \\
\hline
\end{tabular}
}
\end{center}
\caption{ Comparison of regret bounds for Exo-MDPs across the two observability regimes. For both settings, we establish nearly matching upper and lower bounds.
Observing the exogenous state improves the optimal regret by a factor of $\tilde \Theta(\sqrt{r})$ compared to the unobserved setting, quantifying the precise statistical benefit of exogenous observability.
As discussed in the introduction, our model focuses on applications where $r \ll \min\{|\StateSpace|,|\ActionSpace|\}$, in which case the minimax lower bound for general discrete MDPs, $\Omega(H \sqrt{|\StateSpace||\ActionSpace|K})$, is substantially larger.}
\label{tab:main_results}
\end{table}

\paragraph{Structural equivalence between model classes.}  Our first contribution establishes a structural connection between three classes of Markov decision processes: Exo-MDPs, classical discrete MDPs, and discrete linear mixture MDPs. At least superficially, the Exo-MDP class appears to be small due to the structural conditions imposed.  However, this initial perspective turns out to be misleading. {As shown in~\Cref{thm:equivalence}, any discrete MDP (with finitely supported rewards) can be represented as an Exo-MDP.}  Conversely, any Exo-MDP with exogenous state size $d$ can be expressed as a discrete linear mixture MDP of dimension $d$, where the transition and reward functions are linear in the exogenous state distribution.  This representational equivalence formally unifies three important model classes in reinforcement learning and establishes that the structural assumptions of Exo-MDPs do not limit generality.  In addition, this connection motivates the introduction of an \emph{effective dimension} $r \le d$, which depends on the linear structure induced by the exogenous distribution in the transition and reward models.  This measure quantifies the intrinsic rank of the resulting mixture representation and can be computed a priori without any samples.

\paragraph{Unobserved exogenous states: fundamental limits and an optimal algorithm.}
Building on the structural equivalence above, we next study the learning problem when the exogenous states are \emph{unobserved}. Note that any Exo-MDP with effective dimension $r$ is an instance of a linear mixture MDP with dimension $r$; yet the inverse mapping from linear mixture MDP to an Exo-MDP does not necessarily preserve the same dimension. A fundamental question is therefore whether or not the presence of latent exogenous structure reduces the sample complexity of learning an Exo-MDP compared to learning general linear mixture MDPs.  We answer this question by characterizing the minimax-optimal scaling of the regret.  For learning an Exo-MDP with horizon $H$ over $K$ episodes, in which the exogenous states have effective dimension $r$, the minimax regret scales as $\tilde{\Omega}(H r \sqrt{K})$, which coincides with the regret lower bound for linear mixture MDPs.  This result shows that even with structured exogenous dependence, learning remains as difficult as in a general linear mixture MDP. We establish this result through a new minimax lower bound for Exo-MDPs in~\Cref{thm:lower_bound}, along with a complementary upper bound obtained by leveraging the linear mixture representation, and a careful analysis of the \HFUCRL algorithm~\citep{zhou2022computationally}, as described in~\Cref{thm:linear_mixture_mdp_low_rank}.

\paragraph{Observed exogenous states: minimax regret characterization.}
We next turn to the regime in which the exogenous states are fully observed. In this setting, the learner can directly estimate the exogenous distribution and use it to form a plug-in model of the endogenous dynamics.  In this context, our main result is to provide a characterization of the minimax-optimal regret that is sharp up to logarithmic factors.  In particular, given the availability of exogenous observations, we prove that the minimax-optimal regret scales as $\tilde{\Theta}(H \sqrt{r K})$.  This result gives a crisp summary of the utility of exogenous observations: their availability reduces the achievable regret by a factor of $\sqrt{r}$ compared to the unobserved setting.  We establish this minimax-optimal scaling by first proving a new minimax lower bound on the regret (\cref{thm:lower_bound_obs}). We then complement this result with a matching upper bound, obtained by analyzing a plug-in estimation method that directly conditions on the observed exogenous states (\Cref{thm:pto_regret}).  Additionally, while this work focuses on the setting of independent and identically distributed (i.i.d.) exogenous states, we discuss how we can naturally capture more general dynamics, including when the exogenous state depends on the action (see \cref{sec:extension}). See \cref{tab:main_results} for a summary of our results.

\paragraph{Empirical validation.}

Finally, we validate our theoretical findings in two canonical application domains, online inventory control with lost sales and positive lead times, and airline revenue management~\citep{littlewood1972,goldberg_survey_2021}.  {In both settings, our algorithms explicitly exploit the structure of the underlying Exo-MDP to achieve strong empirical performance and substantially outperform standard heuristics~\citep{agrawal2022learning}.}  On the other hand, plug-in algorithms that have extra observation on exogenous variables consistently achieve lower regret, highlighting the performance gain from observability.  These experiments demonstrate both the robustness and practical relevance of exploiting exogenous structure for sample-efficient reinforcement learning.

\paragraph{Paper organization.}  The remainder of this paper is organized as follows. \Cref{sec:related} reviews related work, and \cref{sec:preliminary} describes the formal setup of Exo-MDP.
\Cref{sec:structure_and_dimension} establishes the structural equivalence between Exo-MDPs, discrete MDPs, and linear mixture MDPs and defines the ``effective dimension''.  \Cref{sec:no_observation} studies the unobserved exogenous-state regime, presenting matching minimax lower bounds and an optimal algorithm (up to logarithmic factors) based on the linear mixture reduction.  \Cref{sec:full_observation} analyzes the full observation regime, providing nearly matching lower and upper bounds as well as extending the Exo-MDP framework to more general exogenous dynamics, including Markov and action-dependent processes. \Cref{sec:experiments} reports our empirical results.

\paragraph{Notation.} For a positive integer $n$, we denote $[n]\mydefn \{1,2,\dots, n\}$.  For a finite set $\StateSpace$, let $|\StateSpace|$ denote its cardinality. We use calligraphic letters to denote sets, e.g., $\StateSpace,\ActionSpace$; capital letters denote random variables, e.g., $S,\Action,\Reward$; lower case letters denote specific realization of random variables, e.g., $\state, \action, \reward$; and for a distribution over a discrete set of elements, we use bold lower case letters to denote the probability vector corresponding to the multinomial distribution,  e.g., $\pvecxi$. We use lowercase letters with superscripts, e.g., $\exostate^j\in\ExoStateSpace$ to denote elements of a set $\ExoStateSpace$ indexed by $j$.  For a vector $x$, we use $[x]_j$ to denote its $j$-th entry. We use $\tilde{O}(\cdot)$ to denote rates omitting absolute constants and polylogarithmic factors. Fixing an episode $k$, $h \in [H]$ denotes the $h$-th stage of the MDP.  Lastly, we let $(x)^+ = \max\{x, 0\}$.  See~\cref{tab:notation} in the appendix for a table of notation.


\section{Related work}
\label{sec:related}
In this section, we discuss the lines of research most closely related to the study of exogenous MDPs for operations research problems. For more background, we refer the reader to standard books on dynamic programming and reinforcement learning ~\citep{Bertsekas_dyn1,puterman2014markov,sutton_reinforcement_2018,powell2022reinforcement,agarwal_reinforcement_nodate}, and sources for multi-armed bandits~\citep{bubeck_regret_2012,slivkins_introduction_2019}.


\paragraph{MDPs with exogenous states.}  
Exogenous MDPs, as a sub-class of structured MDPs, were described by~\citet{powell2022reinforcement}, and have been further studied in an evolving line of work (e.g., \cite{dietterich_discovering_2018,efroni2022sample,powell2022reinforcement,alvo2023neural,sinclair2023hindsight,feng2021scalable,chen2024qgym}). Specifically, \citet{dietterich_discovering_2018} and~\citet{efroni2022sample} have studied the case when the rewards or transitions factorize so that the exogenous process can be filtered out. While doing so simplifies algorithm development, it can lead to sub-optimality, since policies agnostic to the exogenous states need not be optimal. Other work studies the use of hindsight optimization, showing that the regret for hindsight optimization policies can be bounded by the hindsight bias, a problem-dependent term~\citep{sinclair2023hindsight,feng2021scalable}. The overarching assumption in this literature is that the exogenous states are fully observed, or equivalently, that one can obtain an unbiased signal~\citep{madeka2022deep}; this assumption of full observations is not realistic for many problems of interest.  We extend the literature by presenting algorithms tailored for Exo-MDPs with unobserved exogenous states as well as refining the results on the fully observed case.

\smallskip
\paragraph{Linear mixture MDPs and other structured MDPs.} 
As shown in~\Cref{thm:equivalence}, any Exo-MDP can be cast as a {discrete} linear mixture MDP (and vice versa), and our analysis is the first to establish the close connection of Exo-MDPs to linear mixture MDPs~\citep{ayoub2020model,zhou2021nearly,chae25a}.
Related work by~\citet{zhou2022computationally} provides an $\Omega(Hd\sqrt{K})$ lower bound on the regret for time-homogeneous linear mixture MDPs, whereas \citet{zhou2021nearly} proved an $\Omega(H^{3/2}d\sqrt{K})$ lower bound for time-inhomogeneous linear mixture MDPs.  In contrast, we provide an entirely new construction tailored to Exo-MDPs, and use it to sharpen the dependence to the \emph{effective dimension} $r$ as opposed to the (typically larger) ambient dimension $d$. Note that any Exo-MDP with $|\ExoStateSpace| = d$ is a special case of a linear mixture MDP with dimension $d$. However, important for our focus, the constructions in past work~\citep{zhou2022computationally,zhou2021nearly} are \emph{not} Exo-MDPs, so these results do not translate to a lower bound for Exo-MDPs with the same dimension. The novel lower bound that we derive reveals an interesting fact: the class of Exo-MDPs with exogenous state size $|\ExoStateSpace| = d$, despite being a subset of linear mixture MDPs of dimension $d$, is as difficult as the class of linear mixture MDPs.

More generally in the literature, RL algorithms exploiting structural properties either make parametric or non-parametric assumptions on the underlying MDP.  For example, a common non-parametric assumption is Lipschitz or smoothness conditions on the $Q$-function (e.g.,~\citet{shah2018q, sinclair2023adaptive}). Such models, while being more flexible, lead to regret bounds that (due to the curse of dimensionality) scale exponentially in dimension $d$.  Parametric assumptions sacrifice flexibility, but with the statistical and computational benefits of polynomial dependence on the dimension. Roughly speaking, such models are based on a feature representation under which the underlying MDP is well-approximated by a parametric (and often linear) model. Recent years have seen tremendous activity on RL with linear function approximation. These works can be categorized depending on their assumptions on the underlying MDP, including MDPs with low Bellman rank~\citep{pmlr-v70-jiang17c,dann2018oracle}, linear MDPs~\citep{pmlr-v97-yang19b,jin2020provably,hu2024fast}, low inherent Bellman error~\citep{pmlr-v119-zanette20a}, or linear mixture MDPs~\citep{ayoub2020model,zhou2021nearly}.

\smallskip

\paragraph{Exo-MDPs in practice.} 
We present simulation results on airline revenue management and inventory control with lost sales, censored demand, and positive lead times.  \citet{agrawal2022learning} design an online learning algorithm tailored specifically to the inventory control setting to learn the optimal base-stock policy, a well-known heuristic policy that is optimal under restrictive settings.  Our algorithms have the benefit of providing regret guarantees to the {\em true optimal} policy. Our empirical results in \cref{sec:experiments} show that our algorithms surpass the sub-optimality of this heuristic class and instead converge to the true optimal policy. Various researchers~\citep{madeka2022deep,alvo2023neural} have studied specializations of Exo-MDPs to these inventory settings, along with associated regret analysis.  Their analysis is predicated on observing an unbiased signal from which the true demand can be recovered; in sharp contrast, our algorithms apply even when the demand is fully unobserved. In related work, \citet{fan2024don} have analyzed sample complexity for inventory control problems for when the demand is fully observed, and \citet{qin2023sailing} for when censored demand is observed. We lastly note that deep reinforcement learning algorithms have been applied in other applications (without exploiting their Exo-MDP structure) including ride-sharing systems~\citep{feng2021scalable}, stochastic queuing networks~\citep{dai2021queueing}, and jitter buffers~\citep{fang2019reinforcement}. Applications of our method can potentially improve sample efficiency in these applications by exploiting the underlying exogenous structure.


\section{Formal Setup}
\label{sec:preliminary}

In this section, we provide basic background on Markov decision processes and the Exo-MDP class, along with a discussion of the performance metrics of interest.  We consider time-homogeneous episodic tabular (discrete) Markov decision processes (MDPs) with finite state and action spaces.  In brief, a Markov decision process (MDP) can be represented by a tuple \mbox{$\MDP =(\StateSpace, \ActionSpace,\Horizon, \state_1, \transition, \Reward)$,} where $\StateSpace$ is the set of states, $\ActionSpace$ is the set of actions, horizon $\Horizon \geq 1$ is the number of stages in each episode, $\state_1 \in \StateSpace$ is a fixed initial state, $\transition(\cdot \mid \state_h, \action_h)$ gives the probability distribution over $\StateSpace$ for the next state $\state_{h+1}$ based on the state-action pair $\state_h, \action_h$ at stage \mbox{$h =1, 2, \ldots, H$} in an episode.  Let $\Delta([0,1])$ denote the set of probability distributions on $[0,1]$, the reward distribution is specified by a mapping $\Reward:\StateSpace\times \ActionSpace\rightarrow \Delta([0,1])$, specifying the distribution of the stochastic reward over the interval $[0,1]$. {We distinguish the reward distribution $\Reward$ from the mean reward and write $r(s,a):=\E[R\mid S=s,A=a]$.} Without loss of generality, throughout this paper we assume a fixed starting state $\state_1$; see~\Cref{sec:fixed_starting_state} for a discussion on the generality this assumption. In a time-homogeneous MDP, the {transition probability} $\transition$ and reward functions $\Reward$ remain fixed across stages $h$, whereas they may change in the time-inhomogeneous setting. We focus on the time-homogeneous setting in the main text and explain extensions to the time-inhomogeneous setting in~\cref{app:preliminary,app:non_homogeneous}.


\subsection{Exo-MDP: Markov Decision Processes with Exogenous States}
\label{sec:exo_mdp_setup}

\begin{figure}[!t]
\centering 

\scalebox{.6}{
\begin{minipage}[c]{0.3\textwidth}
\begin{center}
\begin{tikzpicture} [->, 
  >=stealth, 
  node distance=3cm,
  every state/.style ={thick,fill=gray!10},
  initial text=$ $]
  \node[state](1){$S_h$}; 
  \node[state,right of=1](2){$S_{h+1}$}; 
  \node[state,below of=1,fill=blue!10](3){$A_h$}; 
  \node[state,right of=3,fill=green!10](4){$R_h$}; 
  \draw
  (1) edge[]node{}(2) 
  (1) edge[]node{}(4)
  (3) edge[]node{}(2) 
  (3) edge[]node{}(4);
\end{tikzpicture}    
\end{center}
\end{minipage}

\begin{minipage}{.05\textwidth} 
\hspace{.05cm}
\end{minipage}

\begin{minipage}[c]{0.3\textwidth}
    \begin{center}
    \begin{tikzpicture} [->, 
    >=stealth, 
    node distance=3cm,
    every state/.style ={thick,fill=gray!10},
    initial text=$ $]
    \node[state,fill=red!10](0){$\ExoState_h$};
    \node[state, right of = 0,fill=red!10](5){$\ExoState_{h+1}$};
    \node[state, below of = 0](1){$S_h$};
    \node[state,right of=1](2){$S_{h+1}$};
    \node[state,below of=1,fill=blue!10](3){$A_h$};
    \node[state,right of=3,fill=green!10](4){$R_h$};
    \draw
    (1) edge[]node{}(2) 
    (1) edge[]node{}(4)
    (3) edge[]node{}(2) 
    (3) edge[]node{}(4)
    (0) edge[above]node{}(2) 
    (0) edge[above]node{}(4) ;
    \end{tikzpicture}
    \end{center}
\end{minipage}

\begin{minipage}{.05\textwidth} 
\hspace{.05cm}
\end{minipage}

\begin{minipage}[c]{0.4\textwidth}
    \begin{center}
    \begin{tikzpicture}[
        >=stealth,
        node distance=3cm,
        on grid,
        auto
    ]
    \node[
        circle,
        draw,
        fill=yellow!10,
        minimum size=4cm,
        align=center,
        font=\large
    ] (B) {
        Exo-MDPs\\[4pt]
        {\rotatebox{90}{$=$}}\\[4pt]
        Discrete Linear Mixture MDPs\\[4pt]
        {\rotatebox{90}{$=$}}\\[4pt]
        Discrete MDPs
    };
    \end{tikzpicture}
    \end{center}
\end{minipage}

}

\caption{Directed graphical models representing a generic MDP (left), and an Exo-MDP (middle). In a generic MDP, the state space is fully endogenous, the current state $S_h$ and action $\Action_h$ impact the next state $S_{h+1}$ and reward $\Reward_{h}$. In an Exo-MDP, the state is partitioned into endogenous component $S_h$ and exogenous component $\ExoState_h$.  The $\ExoState_h$ is drawn i.i.d from distribution $\PXi$ independent of $(S_h, \Action_h)$. The known deterministic functions $\SysFun$, $\RewFun$ are such that $S_{h+1} = \SysFun(S_h, \Action_h, \ExoState_h)$ and $\Reward_h = \RewFun(S_h, \Action_h, \ExoState_h)$.  The right panel gives the structural equivalence relations between the class of Exo-MDPs, discrete MDPs and discrete linear mixture MDPs.}
\label{fig:graphical_model_ExoMDP}
\end{figure}
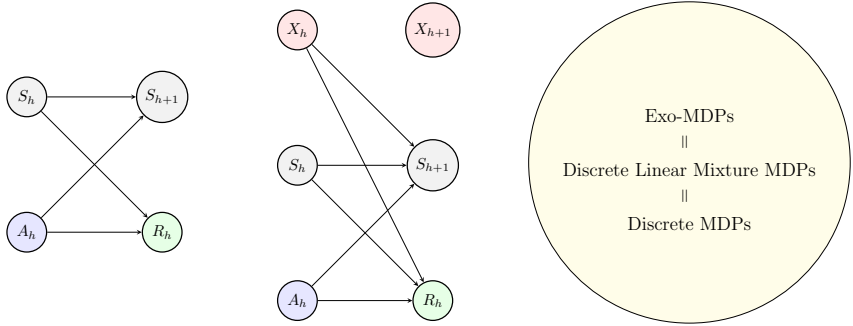

We now turn to the class of Exogenous Markov Decision Processes (Exo-MDPs)~\citep{efroni2022sample,sinclair2023hindsight}.  They are defined by a partition of the state space into two parts: the \emph{endogenous states} $\StateSpace$, and the \emph{exogenous states} $\ExoStateSpace$.  {Both the endogenous and exogenous states affect the dynamics of the system, but the agent's actions only influence the dynamics of the endogenous states, not the exogenous states. See~\Cref{fig:graphical_model_ExoMDP} (left and middle) for an illustration of the distinctions between a standard MDP and an Exo-MDP.}

%

More formally, an Exo-MDP is represented by a tuple \mbox{$\MDP[\PXi,\SysFun,\RewFun] = (\StateSpace, \ExoStateSpace, \ActionSpace,\Horizon, \state_1, \PXi, \SysFun, \RewFun)$,} where $\StateSpace$ is the (endogenous) state space, $\ExoStateSpace$ is the exogenous state space, $\ActionSpace$ is the action space, $\Horizon$ is the horizon length, $\state_1$ is the fixed starting state, and $\SysFun$ and $\RewFun$ are the corresponding transition and reward functions. At stage $h$ the system is in an endogenous state $S_h \in \StateSpace$; an action $\Action_h \in \ActionSpace$ is chosen, and an exogenous state $\ExoState_h \in \ExoStateSpace$ is realized. The exogenous process $(\ExoState_h)_{h=1}^\Horizon$ evolves in a time-homogeneous way that is independent of $(S_h,\Action_h)$, with each $\ExoState_h$ an i.i.d. sample from an unknown distribution $\PXi$.  We fix some particular indexing $\ExoStateSpace = \{\exostate^j\}_{j=1}^d$ of the exogenous state space, and let $\pvecxi$ denote the probability vector corresponding to $\PXi$, i.e.  $\pvecxi=(\PXi(\ExoState=\exostate^1), \dots, \PXi(\ExoState=\exostate^{|\ExoStateSpace|})) \in [0,1]^{|\ExoStateSpace|}$.
Let $d = |\ExoStateSpace|$ denote the cardinality of the exogenous state space. Throughout this paper we assume the exogenous state space is discrete with finite cardinality. This fits many  application scenarios in operations research, e.g. the demand in inventory control, the type of jobs submitted in a cloud computing system, among other examples~\citep{fan2024don,sinclair2023hindsight}.

The state transition dynamics are specified in terms of a deterministic function $\SysFun: \StateSpace \times \ActionSpace \times \ExoStateSpace \rightarrow \StateSpace$, whereas the rewards are specified in terms of a deterministic function $\RewFun: \StateSpace \times \ActionSpace \times \ExoStateSpace \rightarrow [0,1]$.  Given a $H$-length realization $\{ \ExoState_h \}_{h=1}^H$ of exogenous variables, the states and rewards are
generated as
\begin{subequations}
  \begin{align}
    \label{eqn:sysfun}
S_{h+1} & = \SysFun(S_h, \Action_h, \ExoState_h) \qquad \mbox{for $h =
  1, \ldots, H-1$, and} \\
\label{eqn:rewfun}  
\Reward_h &= \RewFun(S_h, \Action_h, \ExoState_h) \qquad \mbox{for $h
  = 1, \ldots, H$.}
\end{align}
\end{subequations}
For the applications of interest in this paper, it is reasonable to assume that $\SysFun$ and $\RewFun$ are known and deterministic (e.g., as in inventory control, discussed in the next paragraph). Thus, an Exo-MDP can be viewed as a special case of an MDP with state space $\StateSpace$, and action space $\ActionSpace$ whose transition and reward dynamics are completely determined by the distribution of the exogenous states through known functions $\SysFun$ and $\RewFun$. Lastly, we note that our main theoretical results focus on the time-homogeneous case; we discuss time-inhomogeneous Exo-MDPs where the dynamics of the exogenous state $\ExoState_h\sim \PXih$ can differ across each stage $h$ in~\cref{app:preliminary,app:non_homogeneous}.

\paragraph{Example: Exo-MDPs for inventory control.}
Recall the inventory control (or newsvendor) model briefly discussed in the introduction; we now show how to formalize it as an instance of an Exo-MDP.  Consider the inventory control problem where the goal is to manage inventory (i.e., products in the store) in an episodic setting over some horizon $H$.  For any stage $h = 1, \ldots, H$, let $\Inventory_h$ be the on-hand (i.e. in-store) inventory; let $\ExoState_h$ be the external demand for the product (i.e. customer walking into store to purchase some product); and let $\Order_h$ be the additional number of products ordered.  Assuming that the ordered products arrive instantly, i.e. lead time $L = 0$, the inventory level transitions are given by
\begin{align*}
\Inventory_{h+1} = \SysFun(\Inventory_h, \Order_h, \ExoState_h) =
(\Inventory_h + \Order_h - \ExoState_h)^+.
\end{align*}
Since the external demand is not impacted by decisions about inventory, we assume that $\ExoState_h$ is drawn i.i.d. from an unknown distribution $\PXi$.
{For this application it is natural to state the objective in terms of costs. We define the one-period cost function as
\begin{align*}
\CostFun(\Inventory_h,\Order_h,\ExoState_h)
=
\HoldingCost(\Inventory_h+\Order_h-\ExoState_h)^+
+
\LostSales(\ExoState_h-\Inventory_h-\Order_h)^+,
\end{align*}
where the first term is the holding cost for remaining inventory and the second is the penalty for lost sales. We assume that this cost is normalized to lie in $[0,1]$, and define the corresponding Exo-MDP reward by $\RewFun(\Inventory_h,\Order_h,\ExoState_h) = 1-\CostFun(\Inventory_h,\Order_h,\ExoState_h).$  Thus maximizing cumulative reward is equivalent to minimizing cumulative inventory cost, and all of our regret guarantees apply directly to this cost formulation.}

This can be formulated as an Exo-MDP with $d$ denoting the size of the support\footnote{The assumption of finite support for the demand is common in the literature (e.g.,~\cite{besbes2013implications,fan2024don}).}  for the external demand $\ExoState_h$; the state and action correspond to inventory $\Inventory_h$ and order $\Order_h$ respectively.
The true exogenous state $\ExoState_h$ is unobserved, while only the realized sale $\min\{\Inventory_h + \Order_h, \ExoState_h\}$ is observed.
In \cref{sec:experiments}, we extend the model to incorporate a positive lead time $L \geq 0$ and present an experimental study demonstrating the application of our algorithms to this more general setting.


\subsection{Observations and Performance Objective}
We now turn to the observation models and performance objectives of interest.

\paragraph{Observation models and online learning.}

We consider two observation regimes for the exogenous state: (i) the \emph{no observation regime}, in which $\ExoState_h$ is never observed, and (ii) the \emph{full observation regime}, in which the learner observes $\ExoState_h$ at every stage \emph{after} taking their action.  The majority of the existing literature on Exo-MDPs considers the full observation setting~\citep{sinclair2023adaptive}.  In contrast, a central goal of this work is to develop methods for the setting where  the exogenous state is \emph{not} observed, and quantify the impact of observing the exogenous states on regret.  We analyze both regimes and quantify the exact statistical learning gap between them in terms of regret.

At the beginning, the learning agent is given knowledge of the endogenous and exogenous state spaces ($\StateSpace$ and $\ExoStateSpace$ respectively), the action space $\ActionSpace$, the horizon $H$, along with the deterministic functions $\SysFun, \RewFun$, but does \emph{not} know the probability vector $\pvecxi$ that characterizes the distribution of $(\ExoState_h)_{h \in [H]}$.  {A policy $\policy = \{\policy_h: h \in [H]\}$ consists of a potentially stochastic decision rule for each of the $h \in [H]$ stages with $\policy_h : \StateSpace \rightarrow \distrover(\ActionSpace)$, where $\distrover(\ActionSpace)$ denotes a distribution over the action space.}

The agent interacts with the environment for $K$ episodes. At the beginning of each episode $k\in [K]$, the agent fixes a policy $\policy^k =\{\policy_h^k: h \in [H]\}$.  For any given episode $k = 1, \ldots, K$, the interaction takes place over $H$ stages: each stage $h \in [\Horizon]$, the agent observes $S_{h,k}$ and picks action $\Action_{h,k} \sim \policy_h^k(\cdot \mid S_{h,k})$.  The underlying system dynamics sample the exogenous state $\ExoState_{h,k}$ from $\PXi$, and the agent receives the realized stochastic reward $\Reward_{h,k} = \Reward(S_{h,k},\Action_{h,k}) = \RewFun(S_{h,k},\Action_{h,k}, \ExoState_{h,k})$ with conditional mean $r(S_{h,k},\Action_{h,k})$. The system transitions to $S_{h+1,k} = \SysFun(S_{h,k}, \Action_{h,k}, \ExoState_{h,k})$.
In the full observation regime, the agent additionally observes the exogenous state $\ExoState_{h,k}$ at the end of each stage, whereas the exogenous state is unobserved in the no observation regime.  This continues until the final transition to state $S_{H+1,k}$, at which point the agent chooses policy $\pi^{k+1}$ for the next episode based on all the observations thus far.

For a given policy $\policy$, its value function $\ValFun^\policy_h: \StateSpace\times [\Horizon]\rightarrow \Reals$ at any stage $h \in [H]$ is given by
\begin{align*}
{\ValFun_h^{\policy}(\state,\MDP) \coloneqq \EE_{\ExoState_{\geq
    h},\policy}\left[\sum_{\tau \geq h}\RewFun (S_{\tau},
  \Action_{\tau}, \ExoState_{\tau}) \mid S_h=s, A_\tau \sim \pi_\tau(S_\tau) \right],}
\end{align*}
where $\ExoState_{\geq h}$ denotes a vector $(\ExoState_h, \ldots, \ExoState_H)$.  Let $\pi^* \in \argmax_{\pi} V_h^\pi(s, \MDP)$ be an \emph{optimal policy}, and let $V_h^*$ denote the \emph{optimal value function} with components $V_h^*(s, \MDP) = V_h^{\pi^*}(s, \MDP)$ for $h \in [H]$.


\paragraph{Performance metrics.} Our goal is to learn a near-optimal policy $\pi = \{\pi_h: h \in [H]\}$ with minimal number of samples.  We characterize the quality of a policy sequence $\{\pi^k\}_{k=1}^K$ via its \emph{cumulative regret}
\begin{align}
  \Regret(K) = \sum_{k=1}^K \Big \{ \ValFunOpt_1(s_1) -
  V_1^{\pi^k}(s_1) \Big \}.
\end{align}
We note that bounds on regret also imply bounds on the \emph{value function estimation error}, given by $\ValFunOpt_1(s_1) - \ValFun_1^\policy(s_1)$, where $\policy$ is the final policy after $K$ episodes.  Standard results (e.g.,~\citep{jin2018q}) guarantee that any algorithm with regret $\epsilon$ leads to a policy with value function estimation error that is $O\big(\frac{\epsilon}{K}\big)$. See~\Cref{lem:online_batch} in \cref{app:sec:regret_val_conversion} for more discussion.  In this paper, we focus on procedures with low regret.

  \subsection{Modeling Assumptions}
  \label{ssec:discuss-model}

We conclude the section with some discussion of our modeling assumptions.

\paragraph{Known $\SysFun$ and $\RewFun$.} We assume that the endogenous dynamics $\SysFun$ and reward function $\RewFun$ are known and deterministic. Thus, randomness enters only via the exogenous state.  Known dynamics and rewards may not be suitable in general RL settings, but are realistic for various applications in operations research. For instance, the problems of inventory control, pricing, scheduling, and resource allocation can be modeled with known system dynamics and the only source of uncertainty is the underlying stochasticity.  This assumption also facilitates simulator design, widely used in the queuing and inventory control literature~\citep{madeka2022deep,alvo2023neural,che2024differentiable}. It also avoids pathological cases that are often seen in black-box RL. In particular, many known hardness results for RL stem from planting signals in an exponentially large latent tree.  These adversarial constructions are excluded by knowledge of $\SysFun$ and $\RewFun$, since the complexity of the underlying learning task reduces to estimating the exogenous distribution $\PXi$.

\paragraph{Discrete Exogenous State Space.}  This paper focuses on discrete exogenous state spaces, which (among other consequences) allows us to establish the equivalence between Exo-MDPs and discrete linear mixture MDPs. While some works allow for continuous exogenous states~\citep{sinclair2023hindsight}, our assumption is well-aligned with the inventory control literature which often assumes discrete demand distributions~\citep{besbes2013implications,cheung2023nonstationary}, or scheduling with job types which are naturally discrete~\citep{balseiro2020dual,vera_bayesian_2021}. 
{
A possible extension of our framework is to continuous exogenous variables under suitable regularity conditions on the distribution $\PXi$ and the system functions $\SysFun$ and $\RewFun$, for example Lipschitz dependence on the exogenous state. In such settings, one could discretize the exogenous state space and apply the algorithms developed here~\citep{sinclair2023adaptive}.
}

\paragraph{Computational Complexity.} The focus of this work is on the statistical complexity of learning in Exo-MDPs, which we show depends only on the effective dimension $r$ that is at most the size of the exogenous state space $d$ and independent of the sizes of the endogenous state and action spaces. However, it is worth noting that the computational complexity of our algorithms still scales with these dimensions, since we solve the Bellman equations over the full state space.  In many OR applications, such as those with long lead times or multiple resources, this can become an implementation bottleneck. Addressing this limitation by incorporating techniques from hindsight learning, information relaxation, or stochastic approximation is an important direction for future work~\citep{sinclair2023hindsight,mercier2007performance,balseiro2019approximations}.

\paragraph{Full and No Observations.} 
We study two cases of observability on the exogenous states: full observation and no observation of the exogenous state.  While practical models can fall somewhere in between, analyzing these extremes provides insight into the role of censored information in learning as we are able to quantify a $\tilde{\Theta}(\sqrt{r})$ learning loss due to censored observations. Quantifying how the regret interpolates between these two extremes in settings with additional side information (whether that be lost-sales in inventory control~\citep{besbes2013implications} or one-sided feedback in pricing models~\citep{zhao2019stochastic}) is an important next step.  See \citet{zhang2025reinforcement} for some preliminary work on this topic.


\section{Structural Equivalence and the Effective Dimension of Exo-MDPs}
\label{sec:structure_and_dimension}

In this section, we develop the structural foundations that underlie our regret guarantees.  A central feature of Exo-MDPs is that randomness comes from the exogenous state.  We show how this induces a low-dimensional structure in the transition and reward models; in particular, any Exo-MDP admits a linear mixture representation whose parametrization depends only on the exogenous dynamics.  This perspective not only reveals an explicit mapping between Exo-MDPs and discrete linear mixture MDPs, but also motivates the notion of an \emph{effective dimension} $r$ that captures how many independent directions of exogenous uncertainty influence the Exo-MDP.  


\subsection{Structural Equivalence Between MDP Models}
\label{sec:equivalence_classes}

We start by establishing a structural equivalence between the class of Exo-MDPs and discrete MDPs, along with a connection to the class of discrete linear mixture MDPs. See the right panel of~\Cref{fig:graphical_model_ExoMDP} for an illustration. We emphasize that this structural relationship between Exo-MDPs and other common subclasses of MDPs is novel in the literature.  {Throughout this subsection we restrict attention to discrete MDPs whose reward distributions have finite support. We let $\RewardSpace \subset [0,1]$ denote the finite set containing the support of $\Reward(s,a)$ for every $(s,a)$.\footnote{{This is needed for the exact equivalence between a discrete Exo-MDP and a discrete MDP. If the reward distributions have continuous support the equivalence is valid but requires a continuous exogenous state space.  However, one can approximate general reward distributions by finitely supported distributions, preserving the set of optimal policies.}}}

{We begin by defining the class of discrete linear mixture MDPs~\citep{ayoub2020model,zhang2021model,zhou2021nearly,chae25a}. At a high level, they are characterized by a certain type of linear structure in the probability transitions and reward function.  More precisely:
\begin{definition}
\label{def:linear_mixture_mdp}
An MDP $\MDP=(\StateSpace, \ActionSpace, \Horizon, \state_1, \transition, \Reward)$ is called a \emph{linear mixture MDP} if there exists a vector $\theta \in \RR^d$ along with known feature maps $(s,a) \mapsto \phi_r(s,a) \in \RR^d$ and $(s', s, a) \mapsto \phi_p(s' \mid s, a) \in \RR^d$ such that the dynamics and expected rewards can be written as
\begin{align}
  \transition(s' \mid s,a) = \phi_p(s' \mid s,a)^\top \theta, \quad
  \mbox{and} \quad r(s,a) =
  \phi_r(s,a)^\top \theta.
\end{align}
\end{definition}}
Note that any discrete linear mixture MDP is also a discrete MDP with state space $\StateSpace$ and action space $\ActionSpace$.

We let $\ClassExoMDP(d,\Horizon)$ denote the class of Exo-MDPs with exogenous state size of at most $d$ and horizon $\Horizon$, with the endogenous state and action spaces being unrestricted, and we let $\ClassDiscreteMDP(\StateSpace,\ActionSpace,\Horizon)$ denote the class of discrete MDPs with state space $\StateSpace$, action space $\ActionSpace$, and horizon $\Horizon$.  Finally, we let $\ClassLinearMixture(d,\Horizon)$ denote the class of discrete linear mixture MDPs with dimension at most $d$. With this notation, we have the following structural equivalence:

\begin{theorem}
\label{thm:equivalence}
{
Under the finite reward-support assumption above, the classes of Exo-MDPs, discrete MDPs, and discrete linear mixture MDPs are equivalent. More specifically:
}
\begin{itemize}
\item We have the inclusion
  $\ClassDiscreteMDP(\StateSpace,\ActionSpace,\Horizon)\subset
  \ClassExoMDP(|\StateSpace|^{|\StateSpace||\ActionSpace|}|\RewardSpace|^{|\StateSpace||\ActionSpace|},\Horizon)$
  with $\RewardSpace$ being the set of all possible reward values.
\item We have the inclusion $\ClassExoMDP(d,\Horizon)\subset
  \ClassLinearMixture(d,\Horizon)$.
\end{itemize}
\end{theorem}
Since any instance of a {discrete} linear mixture MDP by definition is an instance of a discrete MDP, the two inclusions given in~\Cref{thm:equivalence} show that all three classes are equivalent, that is, include the same set of MDP instances. In order to prove the theorem, we construct an explicit mapping from any discrete MDP with state space $\StateSpace$ and action space $\ActionSpace$ to an Exo-MDP with exogenous state size $|\StateSpace|^{|\StateSpace| |\ActionSpace|} \; |\RewardSpace|^{|\StateSpace| |\ActionSpace|}$, as well as a mapping from any Exo-MDP with exogenous state space $\ExoStateSpace$ to a {discrete} linear mixture MDP with dimension $d = |\ExoStateSpace|$.

\paragraph{(a) Any discrete MDP is an instance of an Exo-MDP:}
Exo-MDPs by definition are a subclass of MDPs where the transition and reward dynamics are characterized by the restricted forms of \cref{eqn:sysfun} and \cref{eqn:rewfun}. However, it turns out that Exo-MDPs can represent any discrete MDP with the addition of an exogenous state space. Intuitively, we can \emph{lift} the randomness from the transition and reward dynamics as a $2|\StateSpace||\ActionSpace|$-dimensional exogenous state with cardinality $|\StateSpace|^{|\StateSpace||\ActionSpace|}|\RewardSpace|^{|\StateSpace||\ActionSpace|}$. We present an explicit one-to-one mapping between discrete MDPs and discrete Exo-MDPs, concluding that the two classes contain the exact same set of problems.\footnote{We note that while \citet{sinclair2023hindsight} also studies the relationship between discrete MDPs and Exo-MDPs, their reduction requires a continuous exogenous state $\ExoStateSpace=[0,1]$ via the inverse CDF trick and is not one-to-one, therefore, does not lead to a structural equivalence relation between the two classes.}

\begin{lemma}\label{lem:exo_mdp_generality}
Let $\RewardSpace$ denote the range of the reward function $\Reward$.
For any discrete MDP $\MDP =(\StateSpace, \ActionSpace, \Horizon, \state_1, \transition, \Reward)$, there exists an exogenous state space
$\ExoStateSpace\subseteq \StateSpace^{|\StateSpace||\ActionSpace|}\times \RewardSpace^{|\StateSpace||\ActionSpace|}$
following distribution $\PXi$, and transition and reward functions $\SysFun$ and $\RewFun$ such that $\MDP$ is equivalent to an Exo-MDP
$\MDP'[\PXi,
  \SysFun,\RewFun]=(\StateSpace, \ExoStateSpace, \ActionSpace,
\Horizon, \state_1, \transition, \Reward)$. In particular,
$\ClassDiscreteMDP(\StateSpace,\ActionSpace,\Horizon)\subset
\ClassExoMDP(|\StateSpace|^{|\StateSpace||\ActionSpace|}|\RewardSpace|^{|\StateSpace||\ActionSpace|},\Horizon)$.
\end{lemma}

\begin{rproof}
The proof involves recasting the MDP $\MDP$ as an Exo-MDP via a lifting argument. Let $\ExoState$ be a random vector of dimension $2 |\StateSpace| |\ActionSpace|$ with components
\begin{align*}
  \ExoState_{s,a}^{\SysFun} \in \StateSpace \quad \mbox{and} \quad
  \ExoState_{s,a}^{\RewFun} \in \Ima(R(s,a)) \quad \mbox{for each state-action
    pair $(s, a)$.}
\end{align*}
With this definition,  the transition and reward functions can be written as
\begin{align*}
\SysFun(\state,\action,\exostate) \mydefn \exostate_{s,a}^{\SysFun},
\quad \mbox{and} \quad \RewFun(\state,\action,\exostate) \mydefn
\exostate_{s,a}^{\RewFun} \qquad \mbox{for each state-action pair
  $(s,a)$.}
\end{align*}
Let the random variable $\ExoState_{s,a}^{\SysFun}$ be distributed as $\transition(\cdot\mid \state,\action)$, that is, $\SysFun(\state,\action,\exostate)$ takes the distribution of the next state conditional on current state-action pair $(s,a)$ by reading off the coordinate of the random vector $\ExoState_{s\in\StateSpace,a\in\ActionSpace}^{\SysFun}$. Similarly, let the random variable $\ExoState_{s,a}^{\RewFun}$ be distributed as $\Reward(s,a)$, that is, $\ExoState_{s,a}^{\RewFun}$ takes the distribution of the random reward given current state-action pair $(s,a)$. Then the MDP $\MDP$ is equivalent to an Exo-MDP $\MDP'=(\StateSpace, \ExoStateSpace, \ActionSpace, \Horizon, \state_1, \transition, \Reward)$ with deterministic transition and reward functions $\SysFun,\RewFun$, where $\ExoStateSpace\subseteq \StateSpace^{|\StateSpace||\ActionSpace|} \times \RewardSpace^{|\StateSpace||\ActionSpace|}$ with $\RewardSpace$ being the set of all possible rewards.
\end{rproof}

\paragraph{(b) Any Exo-MDP is a linear mixture MDP:} We next show that Exo-MDPs have a linear representation in terms of $\SysFun$ and $\RewFun$, so that they can be recast as linear mixture MDPs (\cref{def:linear_mixture_mdp}).  Note that we can write \begin{align*}
\transition(s' \mid s,a) &= \sum_{\exostate \in \ExoStateSpace}
\indicator_{s' = \SysFun(s,a,\exostate)} \PXi(\exostate) =
\sum_{i=1}^d \indicator_{s' = \SysFun(s,a,\exostate^i)} \pvecxii =
\phi_p(s' \mid s,a)^\top \pvecxi \\
{r}(s,a) &= \sum_{\exostate \in \ExoStateSpace}
\RewFun(s,a,\exostate)\PXi(\exostate) = \sum_{i=1}^d
\RewFun(s,a,\exostate^i) \pvecxii = \phi_r(s,a)^\top \pvecxi.
\end{align*}
From this representation, we see that any Exo-MDP is a linear mixture MDP with the probability mass function $\pvecxi$ serving as the parameter vector.  We summarize as the following lemma:
\begin{lemma}
\label{lem:linear_mixture_reduction}
Any Exo-MDP with exogenous state space indexed as $\ExoStateSpace =
\{\exostate^j\}_{j=1}^d$ can be written as a linear mixture MDP with {coefficient vector $\theta = \pvecxi=(\PXi(\ExoState=\exostate^1),\dots,\PXi(\ExoState=\exostate^d))$}, and feature vectors
\begin{align}
\label{EqnKeyFeature}
\phi_p(s' \mid s,a) = \bigl[\indicator_{s'=\SysFun(s,a,\exostate^j)}
  \bigr]_{j=1}^d \in \mathbb{R}^d \quad \mbox{and} \quad \phi_r(s,a) =
\bigl[\RewFun(s,a,\exostate^j)\bigr]_{j=1}^d \in \mathbb{R}^d.
\end{align}
\mbox{for each state-action pair $(s,a)$.}
\end{lemma}

\begin{remark}
Importantly we note that although any Exo-MDP with exogenous state size $|\ExoStateSpace| = d$ is an instance of a linear mixture MDP with dimension $d$; the inverse mapping from linear mixture MDP with dimension $d$ to Exo-MDP does not necessarily preserve the same dimension since the lifting argument presented earlier can lead to an Exo-MDP with exogenous state space larger than $d$. Therefore, taking the dimensionality of the problem classes in mind, we have a strict inclusion of $\ClassExoMDP(d,\Horizon)\subsetneq\ClassLinearMixture(d,\Horizon)$. This means that lower bounds for linear mixture MDPs with dimension $d$ do not automatically imply lower bounds for Exo-MDPs with exogenous state size $d$.
\end{remark}


\subsection{The Effective Dimension}
\label{sec:effective_dimension}

A key consequence of the structural equivalence established above is that the transition and reward models of an Exo-MDP depend on the exogenous process via an embedded linear structure that may have dimension substantially lower than the cardinality $d$ of the exogenous state space. Recalling the feature vectors~\eqref{EqnKeyFeature} from~\Cref{lem:linear_mixture_reduction}, we define the \emph{transition information matrix}
\begin{align}
\Fmattrans \in \mathbb{R}^{|\StateSpace|^2|\ActionSpace| \times d}
\quad \mbox{with entries} \quad \qquad
\Fmattrans_{(s',s,a),\cdot} \defn \phi_p(s' \mid s,a)^\top,
\end{align}
and the \emph{reward information matrix}
\begin{align}
\Fmatrew \in \mathbb{R}^{|\StateSpace||\ActionSpace| \times d}, \qquad
\Fmatrew_{(s,a),\cdot} \defn \phi_r(s,a)^\top.
\end{align}
{We combine these into the \emph{joint information matrix}
\begin{align}
\label{eq:joint_information_matrix}
\Fmat \defn
\begin{pmatrix}
\Fmattrans \\
\Fmatrew
\end{pmatrix}
\in \mathbb{R}^{\left(|\StateSpace|^2|\ActionSpace|+|\StateSpace||\ActionSpace|\right)\times d}.
\end{align}
The key complexity parameter in our analysis is the \emph{effective dimension}  $r \defn \rank(\Fmat).$ } Thus, the effective dimension measures the dimension of the joint linear subspace through which the exogenous distribution affects both the transition dynamics and expected rewards.

Note that the effective dimension $r$ can be computed \emph{a priori} based on knowledge of the exogenous state space $\ExoStateSpace$, endogenous state space $\StateSpace$, the action space $\ActionSpace$, and the system functions $\SysFun,\RewFun$. Since the joint information matrix $\Fmat$ has $d$ columns, a conservative upper bound on $r$ is given by
\begin{align*}
r \leq \min\left\{d,\; |\StateSpace|^2|\ActionSpace|+|\StateSpace||\ActionSpace|\right\}.
\end{align*}
{Moreover, by construction we have $\max\left\{\rank(\Fmattrans),\rank(\Fmatrew)\right\} \leq r \leq \rank(\Fmattrans)+\rank(\Fmatrew).$ }However, there are many settings in which $r$ is much lower than this worst-case upper bound.
We next briefly outline several different assumptions on the joint information matrix which result in the effective dimension $r$ being smaller than the size of the exogenous state space~$d$.

\begin{proposition}
\label{prop:effective_low_rank_range}
Consider an arbitrary Exo-MDP $\MDP[\PXi,\SysFun,\RewFun]$.  Suppose
there exists a \emph{coarsening map} $\psi: \ExoStateSpace
\to \ExoStateSpace$ such that, for each state-action pair $(s,a)$ we
have
\begin{align*}
\SysFun(s,a,\exostate)=\SysFun(s, a, \exostate') \qquad \text{and}
\qquad \RewFun(s, a, \exostate)=\RewFun(s, a, \exostate') \quad
\text{for all pairs $\exostate, \exostate'$ with $\psi(\exostate) =
  \psi(\exostate')$.}
\end{align*}
Then the effective dimension is bounded as $r \le |\range(\psi)|$.
\end{proposition}
\begin{rproof}
Consider the row of $\Fmattrans$ indexed by an arbitrary triple
$(s',s,a)$.  If $\psi(\exostate)=\psi(\exostate')$, then we have the
equivalence \mbox{$\Fmattrans[(s',s,a),\exostate] =
  \indicator_{s'=\SysFun(s,a,\exostate)} =
  \indicator_{s'=\SysFun(s,a,\exostate')} =
  \Fmattrans[(s',s,a),\exostate']$.}  Likewise, for every reward row
$(s,a)$, we have \mbox{$\Fmatrew[(s,a),\exostate] =
  \RewFun(s,a,\exostate) = \RewFun(s,a,\exostate') =
  \Fmatrew[(s,a),\exostate']$.}  {Consequently, the columns corresponding to $\exostate$ and $\exostate'$ are identical in both $\Fmattrans$ and $\Fmatrew$, and therefore also in the joint information matrix $\Fmat$.  Thus $\Fmat$ has at most $|\range(\psi)|$ distinct columns. Hence $r = \rank(\Fmat) \leq |\range(\psi)|,$ as claimed.}
\end{rproof}

This proposition is useful, because it allows us to identify tabular MDPs for which the effective dimension $r$ is far smaller than what the naive embedding result of~\Cref{thm:equivalence} would suggest. Applying~\Cref{prop:effective_low_rank_range} hinges upon identifying a suitable coarsening map. Examples include Exo-MDPs in which exogenous states possess latent types that fully determine both transition and reward behavior. As one simple example, suppose that the exogenous state space factors as a finite product $\ExoStateSpace = \ExoStateSpace_1 \times \cdots \times \ExoStateSpace_m$, so that any exogenous state can be written as $\exostate = (\exostate_1,\ldots,\exostate_m)$. Moreover, suppose that there is a nonempty index set $J \subseteq [m]$ such that, for all state-action pairs $(s,a)$, we have $\SysFun(s,a,\exostate)=\SysFun(s,a,\exostate')$ and $\RewFun(s,a,\exostate)=\RewFun(s,a,\exostate')$ for any pair $\exostate,\exostate'$ with $\exostate_j=\exostate'_j$ for all $j\in J$. Given this set-up, it follows that the $J$-projection map
\begin{align*}
\psi_J(\exostate_1,\ldots,\exostate_m) = (\exostate_j)_{j\in J}
\end{align*}
is a coarsening map. Noting that $\range(\psi_J)$ has cardinality $\prod_{j\in J}|\ExoStateSpace_j|$, an application of~\Cref{prop:effective_low_rank_range} yields that the effective dimension $r$ is upper bounded as $r \leq \prod_{j\in J} |\ExoStateSpace_j|.$


\section{Learning with Unobserved Exogenous States}
\label{sec:no_observation}
In this section, we study the learning problem when the exogenous state is completely unobserved, and derive two main results.  First, we establish a minimax lower bound that characterizes the fundamental difficulty of learning in this setting.  Second, we leverage the linear mixture structure from \cref{sec:equivalence_classes} to adapt an existing algorithm from the linear-mixture MDP literature~\citep{zhou2022computationally}, and show that it achieves a matching regret guarantee.  Our minimax lower bound is novel, in that the argument given by \citet{zhou2022computationally} establishes a lower bound of $\Omega(Hd\sqrt{K})$ for \emph{time-homogeneous} linear mixture MDPs of dimension $d$. However, this result does not apply to Exo-MDPs. This is because any Exo-MDP with exogenous state size $|\ExoStateSpace| = d$ can be embedded into a $d$-dimensional linear mixture MDP (i.e., $\ClassExoMDP(d,\Horizon)\subsetneq\ClassLinearMixture(d,\Horizon)$), but not the other way around: lower bounds for linear mixture MDPs do not automatically imply lower bounds for Exo-MDPs.  A natural question, therefore, is whether Exo-MDPs with exogenous state dimension $d$ are statistically as difficult as general linear mixture MDPs with dimension $d$. We answer this affirmatively. In \cref{sec:unobs_lower_bound}, we construct a new family of Exo-MDP instances that force any algorithm to incur regret at least $\Omega(H r\sqrt{K})$ in the time-homogeneous case.  We then show that an adaptation of the \HFUCRL algorithm achieves a regret upper bound $\tilde{O}(H r \sqrt{K})$, matching these lower bounds up to logarithmic factors~\citep{zhou2022computationally}. Together, these results establish that Exo-MDPs are statistically as hard as linear mixture MDPs of the same dimension, and that the adapted algorithm is (nearly) minimax optimal.  In~\Cref{app:non_homogeneous}, we describe the extension of these results to the non-homogeneous setting. 

\subsection{Lower Bound on Regret}
\label{sec:unobs_lower_bound}

In this section, we present a regret lower bound of $\Omega(Hr\sqrt{K})$ for time-homogeneous Exo-MDPs.  The lower bound is on the expected regret, calculated over both the distribution $\PXi$ and the chosen policy.  See~\Cref{app:non_homogeneous} for the extension to the time-inhomogeneous setting, where we establish a lower bound scaling as $\Omega(H^{3/2}r\sqrt{K})$.

\begin{theorem}
\label{thm:lower_bound}
Consider an effective dimension $r \geq 2$ and number of episodes $K \geq \frac{2}{5}r^2$.  Then for any learning algorithm, there exists a time-homogeneous Exo-MDP $\MDP$ with effective dimension $r$ such that its expected regret over $K$ episodes on the Exo-MDP $\MDP$ is lower bounded as
\begin{align}
\label{eq:lb.1}
\E[\Regret(K)] & \geq c \: H \: r \: \sqrt{K}
\end{align}
for a universal constant $c > 0$.
\end{theorem}

Here we describe the hard instance that underlies our lower bound; see~\Cref{app:lower_bound_proof} for the full argument.  Our lower bound construction builds upon the hardness of learning a single-horizon $(H = 1)$ Exo-MDP, which we call an {Exo-Bandit}. Specifically, we construct a ``hard'' {Exo-Bandit} instance, cast as a certain type of linear bandit with actions ranging over the Boolean hypercube, for which any procedure must have regret lower bounded as $\Omega(r\sqrt{K})$. Using this hard Exo-Bandit instance as a building block, we then construct a hard instance $\widetilde{\MDP}$ of an Exo-MDP.  At stage $h = 1$, the Exo-MDP $\widetilde{\MDP}$ inherits the reward structure as the hard Exo-Bandit. For subsequent stages $h= 2, 3, \dots, H$, the specific forms of $\SysFun$ and $\RewFun$ force the reward from the first stage to repeat $H$ times regardless of the actions or exogenous states, without revealing any additional information on $\PXi$.  This directly leads to a lower bound of $\Omega(Hr\sqrt{K})$. We outline the hard instance of the Exo-MDP $\widetilde{\MDP}[\PXi(\Tilde{Z}), \SysFun,\RewFun] = (\StateSpace, \ExoStateSpace, \ActionSpace, \Horizon, \state_1, \transition, \Reward)$ below.   As verified in the full proof (\cref{app:lower_bound_proof}), the effective dimension of this Exo-MDP instance is $r+1$.

The endogenous state space of $\widetilde{\MDP}$ is given by $\Ss = s_1\cup \{(h, r)\mid h\in \{2,3,\dots,H\}, r\in \{-1,1\}\}$. That is, the endogenous state space $\Ss$ consists of $s_1$, a single starting state, and each of the next $H-1$ states are indexed by the stage $h \in [H]$ as well as a single number $r\in \{-1,1\}$. The exogenous state space is given by $\ExoStateSpace = [d] = \{1,2,\dots,d\}$, where $d = 2r$. The action set $\ActionSpace$ sits on a subset of the $d$-dimensional hypercube, where
\begin{align*}
\ActionSpace = \{([Z]_1,-[Z]_1,[Z]_2,-[Z]_2,\dots, [Z]_{r}, -[Z]_{r})\mid Z\in\{-1,1\}^{r}\} \subset \{-1,1\}^d.
\end{align*}
Each action $a\in \ActionSpace$ is completely characterized by a vector $Z\in\{-1,1\}^{r}$ where
\begin{align*}
a(Z)=([Z]_1,-[Z]_1,[Z]_2,-[Z]_2,\dots, [Z]_{r}, -[Z]_{r}).
\end{align*}
The (unknown) distribution $\PXi$ for the exogenous state $\ExoState$, parameterized by $\Tilde{Z}\in \{-1,1\}^{r}$ and constant $c=\frac{1}{10}\sqrt{\frac{2}{5K}}$, is given by
\begin{align*}
\pvecxi(\Tilde{Z}) = (\PXi(1),\dots, \PXi(d)) =
  \Bigl(\tfrac{1}{d}+c [\Tilde{Z}]_1,\tfrac{1}{d}-c
       [\Tilde{Z}]_1,\dots, \tfrac{1}{d}+c
       [\Tilde{Z}]_{r},\tfrac{1}{d}-c [\Tilde{Z}]_{r}\Bigr).
\end{align*}
In other words, $\pvecxi$ is almost a uniform distribution except each coordinate is perturbed from $\frac{1}{d}$ by a small constant $c$ or $-c$ depending on the value of $\tilde{Z}$. Intuitively, the hardness comes from correctly guessing the coordinates of these perturbations by choosing an action $a(Z)$ that matches $\tilde{Z}$ closely.

The known state transition function is given by
$$s_{h+1} = \SysFun(s_h,a_h,\exostate_h) = \begin{cases}
(h+1,R) & \quad \text{if } s_h = (h,R),h = 2,3,\dots,H-1\\
(2, R=[a_1]_{\exostate_1}) & \quad \text{if } h=1, s_h = s_1.
\end{cases}$$
The action $a_h$ has no effect on the state transition, except, in the first stage, action $a_1$ assigns value $R=[a_1]_{\exostate_1}$ to the second coordinate of the state, which is then retained and shared across all stages afterwards.
The known reward function is given by
$$\Reward_h = \RewFun(s_h, a_h, \exostate_h) = \begin{cases}
[a_1]_{\exostate_1}&\quad\text{if } h=1, s_h = s_1\\
R &\quad \text{if } s_h = (h,R).
\end{cases}$$
At stage $h=1$, taking action $a_1$ incurs reward $[a_1]_{\exostate_1}$, where $\exostate_1\sim \PXi$. For all $H-1$ stages afterwards, the same reward at the first stage is repeated, leading to a total reward of $H\cdot [a_1]_{\exostate_1}$. Note that the optimal action is to choose $a(Z)$ such that $Z$ exactly matches the unknown vector $\tilde{Z}$.



\subsection{Sample Efficient Algorithm}
\label{sec:unobs_upper_bound}

In this subsection, we continue to study the setting in which the exogenous states are unobserved and complement our minimax lower bound with a matching upper bound.  {Leveraging the linear mixture representation developed in \cref{sec:effective_dimension}, we adapt the \HFUCRL algorithm (\cref{alg:hf-ucrl-exo}) to the Exo-MDP structure~\citep{zhou2022computationally}.}  Since the feature vectors associated with the exogenous process lie in an $r$-dimensional subspace, we run a \emph{rank-reduced} version of the algorithm that restricts exploration to this effective subspace.
This yields a regret guarantee of $\tilde{O}(H r \sqrt{K})$, matching our lower bound up to logarithmic factors.  Thus, we show the effective dimension $r$ fully characterizes the statistical complexity of learning in Exo-MDPs when the exogenous state is unobserved.

Recall that the feature vectors~\eqref{EqnKeyFeature} arose as part of establishing the connection between Exo-MDPs and linear mixture MDPs in~\Cref{lem:linear_mixture_reduction}. This connection is a key enabler since it allows us to leverage algorithms developed for linear mixture MDPs~\citep{ayoub2020model,zhou2021nearly}.  While other connections are possible, here we adopt the \HFUCRL algorithm~\citep{zhou2022computationally} to our setting.
In \cref{thm:linear_mixture_mdp_low_rank_non_homogeneous} we adapt the \UCRL algorithm from \citet{zhou2021nearly} to show an $\tilde{O}(H^{3/2} r \sqrt{K})$ algorithm for non-homogeneous Exo-MDPs.

\paragraph{Details of the \HFUCRL algorithm.}  We apply the \HFUCRL algorithm to Exo-MDPs through the linear mixture representation introduced in \cref{sec:equivalence_classes}~\footnote{{Linear-mixture MDP algorithms are commonly stated under the simplifying assumption that the reward function is known; see, e.g., \citet{ayoub2020model,chae25a}. We use the standard extension to unknown linear rewards by incorporating the observed reward into the value-targeted regression; see, e.g., \citet{wagenmaker22a} for further discussion.}}.  At a high level, the algorithm alternates between two steps each episode. First, it uses weighted ridge regression to update an estimate of the unknown linear parameter associated with the exogenous distribution. This produces an ellipsoidal confidence set based on a self-normalized Bernstein inequality.  Second, using this confidence set, the algorithm performs an optimistic dynamic programming step. It solves the Bellman equations with an upper-confidence bonus to obtain optimistic $Q$-values and selects the greedy policy.  Executing this procedure over episodes yields the regret guarantees stated in \cref{thm:linear_mixture_mdp_low_rank}.  See   \cref{app:linear_mixture_algorithms} for a full algorithm description.

Our main result is to show that an algorithm that exploits the SVD of the information matrix can achieve regret that scales with the rank $r$, as opposed to the ambient exogenous dimension $d$.

\begin{theorem}
\label{thm:linear_mixture_mdp_low_rank}
For any $H$-horizon Exo-MDP with effective dimension $r$, applying a rank-reduced \HFUCRL algorithm over $K$ episodes yields a sequence of policies $\{\pi^k\}_{k=1}^K$ with regret at most
\begin{align}
\label{EqnRegretBound}
\Regret(K) & = \tilde{O}\left( Hr \sqrt{K} + Hr^2 \right).
\end{align}
\end{theorem}
Note that when $K \gtrsim r^2$, we can restate the regret bound~\eqref{EqnRegretBound} more succinctly as $\Regret(K) =   \tilde{O}(H r \sqrt{K})$.  Thus, up to poly-logarithmic factors, it grows linearly in the effective dimension $r$.  When no rank reduction occurs (i.e., $r = d$), then we obtain a regret bound that scales linearly with the cardinality $d$ of the exogenous state space at $\tilde{O}( H d \sqrt{K})$.

{
\begin{rproof}
Let $\Fmattrans$ and $\Fmatrew$ denote the transition and reward
information matrices defined in~\cref{sec:effective_dimension}, and
consider the joint information matrix
\[
\Fmat \defn
\begin{pmatrix}
\Fmattrans\\
\Fmatrew
\end{pmatrix},
\]
with $r = \rank(\Fmat)$.  Consider any rank-$r$ factorization $\Fmat = AB^\top$.  For each transition row $(s',s,a)$ and reward row $(s,a)$, let $A^{(p)}_{s',s,a}\in\RR^{r}$ and $A^{(r)}_{s,a}\in\RR^{r}$ denote the corresponding rows of $A$, viewed as column vectors. Define the common unknown parameter
\[
\theta \defn B^\top\pvecxi \in\RR^{r},
\]
and the reduced transition and reward features $\widetilde\phi_p(s'\mid s,a) = A^{(p)}_{s',s,a}$ and $\widetilde\phi_r(s,a) = A^{(r)}_{s,a}.$  By the definition of the information matrices and the factorization $\Fmat=AB^\top$, we then have
\begin{align*}
\transition(s'\mid s,a)
=
e_{s'\mid s,a}^\top\Fmattrans\pvecxi
=
\widetilde\phi_p(s'\mid s,a)^\top\theta,
\quad
r(s,a)
=
e_{s,a}^\top\Fmatrew\pvecxi
=
\widetilde\phi_r(s,a)^\top\theta.
\end{align*}

We apply the rank-reduced \HFUCRL algorithm to this
$r$-dimensional representation.   Applying their regret guarantee therefore gives
$\Regret(K) = \widetilde O\left(H r\sqrt{K}+H r^2\right).$
\end{rproof}
}

\noindent Note that the full information matrices depend on $\StateSpace, \ActionSpace, \ExoStateSpace, \SysFun, \RewFun$, all of which are known a priori to the agent, therefore require no samples to compute. The singular decompositions can be computed in time polynomial in $|\StateSpace|,|\ActionSpace|$, and $d$.


\begin{remark}
The rank reduction procedure we have described above generally holds for linear mixture MDPs with discrete state and action spaces. For any generic linear mixture MDP, we can similarly compute the full information matrices $\Fmattrans$ and $\Fmatrew$ by vertically stacking the known feature vectors, and perform the same SVD decomposition to project the features to their rowspace. Essentially, for a feature space with low-rank, then, at the cost of computing the rank of the full information matrices as well as its SVD decomposition, one can reduce the regret by a factor $\frac{d}{r}$, corresponding to the ratio of rank reduction.
\end{remark}

\section{Learning with Observed Exogenous States}\label{sec:full_observation}
 So far we have focused on the no-observation regime, where the algorithm does not observe samples of the exogenous states. In some applications, it may be possible to observe past trajectories of the exogenous states. For example, in inventory control under the demand backlog model, it could be possible to observe not only the realized order, but also the actual order after the fact~\citep{madeka2022deep}.

 It thus becomes interesting to quantify the reduction in regret due to extra observability of the exogenous state trajectory. Accordingly, we now turn to the full observation regime when the exogenous states are observed and evolve i.i.d. Intuitively, this extra observation helps with policy learning because trajectories of the exogenous state effectively give us a simulator for the transition and reward, which reduces the problem closer to a generative model sampling scheme rather than a general MDP. We establish a minimax lower bound of $\Omega(H\sqrt{rK})$ and present a nearly matching upper bound via a simple plug-in algorithm achieving regret upper bound of $\tilde{O}(H\sqrt{rK})$.  Comparing this with the $\tilde{\Theta}(Hr\sqrt{K})$ rate from \cref{sec:no_observation} shows that observing the exogenous states yields a \emph{$\tilde{\Theta}(\sqrt{r})$-factor improvement} in sample complexity.  We then extend the analysis beyond the i.i.d.\ setting to more general exogenous dynamics, including cases where the exogenous state depends on the action.  Using reductions to linear mixture MDPs, we obtain regret guarantees that remain independent of the size of the endogenous state space.

\subsection{Lower Bound on Regret}
\label{sec:obs_lower_bound}

In this section, we present a lower bound on regret of $\Omega(H\sqrt{rK})$ for time-homogeneous Exo-MDPs.  The lower bound is on the expected regret, calculated over both the distribution $\PXi$ and the chosen policy.  See \cref{app:proofs_observed_lower} for the proof.

\begin{restatable}{theorem}{ObsLowerBound}
\label{thm:lower_bound_obs}
Consider an effective dimension $r \geq 1$, number of episodes $K \geq r$ and horizon $H > 3$.  Then for any algorithm operating in the full-observation Exo-MDP setting, there exists a time-homogeneous Exo-MDP $\MDP$ with effective dimension $r$ such that the expected regret over $K$ episodes obeys
\begin{align}
\E[\Regret(K)] \ge\ \gamma H\sqrt{rK}, \quad \text{where $\gamma$ is a universal constant}.
\end{align}
\end{restatable}

While the full construction of the lower-bound instance is detailed in \cref{app:proofs_observed_lower}, we give some intuition here as to where the $\sqrt{r}$ discrepancy between \cref{thm:lower_bound_obs} and \cref{thm:lower_bound} arises.
With full observations of the exogenous state, each step reveals the realized exogenous state, so that the learner can estimate the underlying $r$-dimensional projection of the exogenous distribution directly from the i.i.d. samples.  Standard concentration then yields estimation error scaling as $\Omega(\sqrt{rK})$, which propagates linearly into the regret through our Exo-MDP construction.  In contrast, when the exogenous states are {\em unobserved}, the hard instance hides the latent vector behind an aggregated scalar signal, forcing any learning algorithm to explore across the $r$ coordinates.  This indirect observation structure introduces an additional $\sqrt{r}$ statistical penalty, leading to the larger regret lower bound in \cref{thm:lower_bound}.

\subsection{Sample Efficient Algorithm}\label{sec:obs_upper_bound}

{In the full-observation regime, the agent observes the exogenous state $\ExoState_h$ after choosing action $\Action_h$ at each stage $h\in[H]$. Recall that all randomness in an Exo-MDP is generated by the exogenous process, while the deterministic system and reward functions $\SysFun$ and $\RewFun$ are known. Consequently, estimating the distribution of the exogenous state $\pvecxi$ is sufficient to construct an estimate of the entire MDP.  To make this dependence explicit, for any probability vector $p\in\Delta(\ExoStateSpace)$, we use $\MDP_p$ to denote the Exo-MDP induced by the known functions $\SysFun,\RewFun$ when the exogenous states are drawn i.i.d. according to $p$.

This observation motivates a simple plug-in algorithm. At the beginning of episode $k\geq2$, let $\D_k:=\{\ExoState_{h,k'}:h\in[H],\,k'<k\}$ denote the $D_k=H(k-1)$ exogenous states observed so far, and let $\widehat p_k$ be their empirical distribution:
\[
\widehat p_k(x) = \frac{1}{D_k} \sum_{k'<k}\sum_{h=1}^H \indicator_{\ExoState_{h,k'}=x}, \qquad x\in\ExoStateSpace.
\]

The \PlugIn algorithm constructs the Exo-MDP obtained by replacing the unknown distribution $\pvecxi$ with $\widehat p_k$, computes a deterministic Markov policy that is optimal for this estimated model,
\[
\pi_k\in\argmax_\pi V_1^\pi(s_1,\MDP_{\widehat p_k}),
\]
using a fixed deterministic rule to break ties, and executes $\pi_k$ during episode $k$. The policy in the first episode may be chosen arbitrarily. The full procedure is summarized in \cref{alg:plugin-exomdp}.

\begin{algorithm}[t]
\caption{\PlugIn for Fully Observed Exo-MDPs}
\label{alg:plugin-exomdp}
{
\begin{algorithmic}[1]
\Require Horizon $H$, number of episodes $K$, starting state $s_1$
\State Initialize $\D_1\gets\emptyset$
\For{$k=1,2,\dots,K$}
    \If{$k=1$}
        \State Choose any policy $\pi_1$
    \Else
        \State Set $D_k\gets H(k-1)$ and compute the empirical distribution
        \[
        \widehat p_k(x)
        \gets
        \frac{1}{D_k}
        \sum_{k'<k}\sum_{h=1}^H
        \indicator_{\ExoState_{h,k'}=x},
        \qquad x\in\ExoStateSpace
        \]
        \State Construct the plug-in model $\MDP_{\widehat p_k}$
        \State Compute
        $\displaystyle
        \pi_k\in\argmax_{\pi}
        V_1^\pi(s_1,\MDP_{\widehat p_k})
        $
    \EndIf
    \State Execute $\pi_k$ for one episode in the true model $\MDP_p$
    \State Observe $\{\ExoState_{h,k}\}_{h=1}^H$
    \State $\D_{k+1}\gets
    \D_k\cup\{\ExoState_{h,k}\}_{h=1}^H$
\EndFor
\end{algorithmic}
}
\end{algorithm}

Although $\widehat p_k$ is a $d$-dimensional empirical distribution, the statistical complexity of \PlugIn does not scale with the ambient dimension $d$. To see why, recall the joint information matrix (\cref{eq:joint_information_matrix})
\[
\Fmat = \begin{pmatrix}
\Fmattrans\\
\Fmatrew
\end{pmatrix},
\qquad
r := \rank(\Fmat).
\]
All transition and reward quantities depend on the exogenous distribution $\pvecxi$ only through its projection onto the row space of $\Fmat$.
We now establish a performance guarantee for our algorithm:

\begin{restatable}{theorem}{ObsUpperBound}
\label{thm:pto_regret}
For any $H$-horizon Exo-MDP with effective dimension $r$, applying \PlugIn for any $\delta \in (0,1)$ over $K$ episodes yields a
sequence of policies $\{\pi^k\}_{k=1}^K$ satisfying with probability at least $1 - \delta$,
\[
\Regret(K) \leq O\left( H\sqrt{rK\log(1+HK)+K\log(2K/\delta)}\right).
\]
\end{restatable}
\noindent In particular, ignoring logarithmic factors, $\Regret(K)\leq \tilde O(H\sqrt{rK})$. See \cref{app:proofs_observed_upper} for the proof.
}

\subsection{Extension to General Dynamics of the Exogenous States}
\label{sec:extension}

The idea of reducing Exo-MDPs to linear mixture MDPs applies to more general dynamics of $\PXi$. Under the assumption that the trajectory of the exogenous state $\ExoState_h$ is fully observed, that is, having access to full trajectories of $\{S_h,\ExoState_h,\Action_h\}$, we show that simple modifications of the feature representation allow us to extend the i.i.d. setting to general dynamics for the exogenous states. We here outline several examples including when $\PXi$ follows a Markov process, a $k$-step Markov process, and a Markov process affected by the agent’s actions, all with efficient regret bounds.

\paragraph{If the exogenous state follows a Markov process affected by the agent’s actions.} Augmenting the state as $(\state_{h+1}, \exostate_h)$, the transition probability can be written as $$\transition(\state_{h+1},\exostate_h \mid \state_h, \exostate_{h-1}, \action_h) = \phi_p^\top(\state_{h+1}, \state_h, \exostate_{h-1}, \action_h) \theta$$
where the feature and unknown coefficient are given by 
\begin{align*}
\phi_p(\state_{h+1}, \state_h, \exostate_{h-1}, \action_h) & = [\indicator_{\state_{h+1}=\SysFun(\state_h,\action_h,\exostate')} \cdot \indicator_{\action_h = \action} \cdot \indicator_{\exostate_h = \exostate'} \cdot \indicator_{\exostate_{h-1} = \exostate}]_{\exostate,\exostate'\in \ExoStateSpace, \action\in\ActionSpace},\\
\theta & = [\PXi(\exostate_{h+1}=\exostate' \mid \exostate_h=\exostate, \action_h = \action)]_{\exostate,\exostate'\in \ExoStateSpace, \action\in \ActionSpace}.
\end{align*}
Applying the \HFUCRL algorithm gives an upper bound $\tilde{O}(d^2 |{\mathcal{A}}|H\sqrt{K})$.

\paragraph{If the exogenous state is Markov.} We can write the transition probability as a $d^2$-dimensional linear mixture representation $$\transition(\state_{h+1}, \exostate_h \mid \state_h, \exostate_{h-1}, \action_h) = \phi_p^\top(\state_{h+1}, \state_h, \exostate_{h-1}, \action_h) \theta$$
where the feature vector and unknown coefficient are given by
\begin{align*}
\phi_p(\state_{h+1}, \state_h, \exostate_{h-1}, \action_h) & = [1_{\state_{h+1}=
\SysFun(\state_h,\action_h,\exostate')}1_{\exostate_{h-1}=\exostate} 1_{\exostate_h = \exostate'}]_{\exostate,\exostate'\in \ExoStateSpace} \\
\theta & = [\PXi(\exostate_{h}=\exostate' \mid \exostate_{h-1}=\exostate)]_{\exostate,\exostate'\in \ExoStateSpace}.
\end{align*}
Again, applying  the \HFUCRL algorithm gives an upper bound $\tilde{O}(d^2 H\sqrt{K})$.

Generally, if the dynamics of the exogenous state transition is a $L$-step Markov process, the transition probability is the inner product of a $d^L$-dimensional feature vector, with the $L$-step Markov probability as the coefficient. Applying the same reduction gives an upper bound $\tilde{O}(d^{L+1} H\sqrt{K})$.

\begin{remark}
The regret bounds above depend on the ambient dimension of the feature representation.
However, applying the same rank-reduction ideas from \cref{sec:effective_dimension} yields regret bounds in terms of the corresponding \emph{effective dimension} $r$, which can be much smaller than the nominal dimension.
\end{remark}

\section{An Experimental Study: Inventory Control}
\label{sec:experiments}

In this section we revisit the inventory control application introduced earlier and use it to study how learning performance depends on key problem parameters. In particular, we evaluate the performance of \HFUCRL (introduced in \cref{sec:unobs_upper_bound}) relative to \PlugIn (\cref{sec:obs_upper_bound}) as the planning horizon $H$ and the dimension $d$ of the exogenous distribution vary. We also examine the robustness of both methods to violations of the baseline modeling assumptions, specifically allowing the exogenous process to be time-inhomogeneous and weakly correlated across timesteps. {See \cref{app:arm_simulations} for additional simulations on an airline revenue management problem~\citep{littlewood1972}.}

To summarize our results, we find that \HFUCRL achieves competitive performance compared with \PlugIn, but also outperforms other black-box reinforcement learning and state-of-the-art algorithms tailored to the inventory control problem.
See~\Cref{app:inventory_control,app:experiment_details} for further details on the experimental setup and simulation details, and \url{https://github.com/jw3479/Exogenous_MDPs} for the implementation.


\subsection{Inventory Control with Lead Time}
\label{app:inventory_description}

{In our experiments, we consider the online inventory control problem from~\Cref{sec:preliminary} with the addition of a {\em lead time $L$}; see the survey~\citep{goldberg_survey_2021} for more details. In particular, a retailer is faced with the task of making ordering decisions online over a fixed horizon of $H$ stages. At the beginning of each stage $h$, the inventory manager observes the current inventory level $\Inventory_h$ as well as the $L$ previous unfulfilled orders in the pipeline, denoted by $\Order_{h-L}, \ldots, \Order_{h-1}$. Here the integer $L \geq 1$ is the {\em lead time}, or the number of stages between placing an order and receiving it. We assume for ease of notation that the system starts with no on-hand inventory and no outstanding orders, i.e., $\Inventory_1=\Order_{1-L}= \dots = \Order_0=0$, and omit the dependence on the initial state.

With this set-up, the system evolves as follows. At the beginning of stage $h$, the inventory manager observes the current inventory level $\Inventory_h$ and previous orders $\Order_{h-L}, \ldots, \Order_{h-1}$, and places an order $\Order_h$ that arrives at stage $h+L$. The order $\Order_{h-L}$ placed $L$ stages earlier then arrives. Next, an unobserved demand $\ExoState_h \geq 0$ is drawn from the unknown demand distribution $\PXi$, independently across stages. The number of products sold is $\min\{\Inventory_h+\Order_{h-L},\ExoState_h\}$.
Thus, the decision maker observes realized sales but not the underlying demand $\ExoState_h$.

The one-period inventory cost is
\[
\CostFun(\Inventory_h,\Order_{h-L},\Demand_h) = \HoldingCost(\Inventory_h+\Order_{h-L}-\Demand_h)^+ + \LostSales(\Demand_h-\Inventory_h-\Order_{h-L})^+,
\]
where $\HoldingCost$ and $\LostSales$ are known holding and lost-sales cost parameters. We normalize this cost to lie in $[0,1]$ and define the corresponding Exo-MDP reward as $\Reward_h = \RewFun(S_h,\Action_h,\ExoState_h) = 1-\CostFun(\Inventory_h,\Order_{h-L},\ExoState_h).$ Thus maximizing cumulative reward is equivalent to minimizing cumulative inventory cost. In the numerical results, we report costs in their original unnormalized scale.  Lastly, the on-hand inventory is updated as $\Inventory_{h+1} = (\Inventory_h+\Order_{h-L}-\ExoState_h)^+.$  The objective is to design an ordering policy $\policy$ that minimizes the total expected inventory cost
\[
\TotalCost_1^\policy
=
\EE_\policy\left[
\sum_{h=1}^H
\CostFun(\Inventory_h,\Order_{h-L},\ExoState_h)
\right].
\]
}

\paragraph{Exo-MDP Formulation.} 
{
To formulate this problem as an Exo-MDP, let $d$ denote the size of the support of the independent demand distribution $\PXi$. We assume for convenience that the support is $\{0, 1, \ldots, d-1\}$.  The state space is $\StateSpace = \{0,1,\dots,d-1\}^{L+1}$, where each state
\[
S_h=(\Inventory_h,\Order_{h-L},\ldots,\Order_{h-1})
\]
consists of the current inventory level $\Inventory_h$, along with the previous $L$ unfulfilled orders $\Order_{h-L},\dots,\Order_{h-1}$ in the pipeline. The exogenous state space is given by $\ExoStateSpace=\{0,1,\dots,d-1\}$, where $\ExoState_h=\Demand_h$ denotes the demand at stage $h$. The action space is given by $\ActionSpace=
\{0,1,\dots,d-1\}$, where action $\Action_h=\Order_h$ denotes the order placed at stage $h$.

We can write the deterministic transition and cost dynamics as
\begin{align}
S_{h+1} & = \SysFun(S_h, \Action_h, \ExoState_h) =
((\Inventory_h+\Order_{h-L} - \Demand_h)^+, \Order_{h-L+1}, \dots,
\Order_{h-1}, \Order_{h}), \\ \AllCost_h & = \CostFun(S_h, \Action_h,
\ExoState_h) = \HoldingCost(\Inventory_h + \Order_{h-L} -
\ExoState_h)^+ + \LostSales(\ExoState_h - (\Inventory_h +
\Order_{h-L}))^+.
\end{align}
The corresponding Exo-MDP reward is then $\RewFun(S_h,\Action_h,\ExoState_h) = 1 -\CostFun(S_h,\Action_h,\ExoState_h).$ We note that the assumption of discrete demand is commonly made in the literature~\citep{besbes2013implications,fan2024don}.
}

\begin{table}[!tb]
{
\caption{Total observed cost at the final episode $K = 500$ for policy $\pi_K$ with a 95\% confidence interval. In parentheses we show the relative performance to the total expected cost of the optimal policy, $(\widehat{C}^{\pi_K} - C_1^*) / C_1^*$, where $\widehat{C}^{\pi_K}$ denotes the empirical average final-episode cost. Some empirical final-episode averages may fall below the optimal expected value due to sampling error.}
\label{tab:performance}

\setlength\tabcolsep{0pt}
\smallskip
\footnotesize
\begin{tabular*}{\columnwidth}{@{\extracolsep{\fill}}lcccc}
\toprule
Algorithm & Scenario I & Scenario II & Scenario III & Scenario IV \\
\midrule
Optimal Policy ($\TotalCost_1^*$)
& $110.9$
& $7.6$
& $10.1$
& $76.2$ \\
Optimal Base-Stock Policy ($\TotalCost_1^{\BaseStock^*}$)
& $131.0\ (18\%)$
& $7.7\ (0\%)$
& $10.1\ (0\%)$
& $76.4\ (0\%)$ \\
\midrule
\PlugIn
& $113.6 \pm 3.5\ (2\%)$
& $7.3 \pm 0.6\ (0\%)$
& $10.6 \pm 1.2\ (5\%)$
& $76.7 \pm 1.9\ (1\%)$ \\
\PlugInNH
& $108.8 \pm 3.6\ (0\%)$
& $8.5 \pm 0.7\ (12\%)$
& $9.6 \pm 0.9\ (0\%)$
& $77.2 \pm 2.0\ (1\%)$ \\
\midrule
\HFUCRL
& $109.0 \pm 3.6\ (0\%)$
& $7.8 \pm 0.7\ (2\%)$
& $9.6 \pm 0.9\ (0\%)$
& $74.7 \pm 2.0\ (0\%)$ \\
\UCRL
& $112.0 \pm 3.7\ (1\%)$
& $7.8 \pm 0.6\ (2\%)$
& $10.0 \pm 0.9\ (0\%)$
& $77.4 \pm 2.0\ (2\%)$ \\
\midrule
\OnlineBaseStock
& $136.2 \pm 4.0\ (23\%)$
& $7.5 \pm 0.7\ (0\%)$
& $10.9 \pm 1.2\ (9\%)$
& $76.5 \pm 2.2\ (0\%)$ \\
\QLearning
& $175.5 \pm 4.8\ (58\%)$
& $7.8 \pm 0.7\ (2\%)$
& $10.4 \pm 1.1\ (3\%)$
& $110.5 \pm 4.2\ (45\%)$ \\
\Random
& $176.0 \pm 4.7\ (59\%)$
& $22.1 \pm 2.4\ (190\%)$
& $28.3 \pm 2.3\ (181\%)$
& $202.3 \pm 7.3\ (165\%)$ \\
\bottomrule
\end{tabular*}

\smallskip

\begin{tabular*}{\columnwidth}{@{\extracolsep{\fill}}lccc}
\toprule
Algorithm & Scenario V & Scenario VI & Scenario VII \\
\midrule
Optimal Policy ($\TotalCost_1^*$)
& $17.3$
& $121.2$
& $111.7$ \\
Optimal Base-Stock Policy ($\TotalCost_1^{\BaseStock^*}$)
& $30.0\ (73\%)$
& $131.6\ (9\%)$
& $133.9\ (20\%)$ \\
\midrule
\PlugIn
& $29.9 \pm 0.4\ (73\%)$
& $120.3 \pm 5.5\ (0\%)$
& $123.2 \pm 6.5\ (10\%)$ \\
\PlugInNH
& $15.1 \pm 4.1\ (0\%)$
& $115.0 \pm 5.5\ (0\%)$
& $131.8 \pm 6.8\ (18\%)$ \\
\midrule
\HFUCRL
& $29.7 \pm 0.4\ (71\%)$
& $128.8 \pm 5.6\ (6\%)$
& $127.3 \pm 5.6\ (14\%)$ \\
\UCRL
& $20.5 \pm 2.4\ (18\%)$
& $119.4 \pm 5.7\ (0\%)$
& $125.6 \pm 6.4\ (12\%)$ \\
\midrule
\OnlineBaseStock
& $39.6 \pm 2.6\ (129\%)$
& $133.7 \pm 6.1\ (10\%)$
& $136.1 \pm 8.9\ (22\%)$ \\
\QLearning
& $29.0 \pm 3.1\ (68\%)$
& $133.4 \pm 7.3\ (10\%)$
& $135.7 \pm 6.8\ (21\%)$ \\
\Random
& $127.3 \pm 11.8\ (635\%)$
& $133.6 \pm 7.1\ (10\%)$
& $143.6 \pm 6.9\ (29\%)$ \\
\bottomrule
\end{tabular*}
}
\end{table}


\subsection{Regret Guarantee for \HFUCRL and \PlugIn on Inventory Control}
\label{regret_inventory_control}

We focus on the performances of \HFUCRL (\cref{sec:unobs_upper_bound}) and the \PlugIn method (\cref{sec:obs_upper_bound}). For comparison purposes, we grant the \PlugIn method additional access to past demand trajectories for comparison, even though the algorithms only observe the sales of $\min\{\ExoState_h, \Inventory_{h} + \Order_{h-L}\}$ under the problem setting. As such, we report the performance of \PlugIn separately since the other algorithms are not granted observation of the exogenous states.

Direct applications of~\Cref{thm:linear_mixture_mdp_low_rank} and~\Cref{thm:pto_regret} yield regret guarantees of $\tilde{O}(H d\sqrt{K})$ and $\tilde{O}(H\sqrt{dK})$ respectively, where $d$ is the size of the support of the demand distribution.  We note that there is no additional rank reduction in this model ($r=d$). See \cref{app:cor:inventory_control} for further details establishing that the full information matrix is full rank.

\begin{corollary}
\label{cor:inventory_control}
Consider a single product stochastic inventory control problem with lost sales and lead time $L \geq 0$, where the demand {$\Demand_h\in \{0, \ldots, d-1\}$}. Then, in the time-homogeneous regime, for any $\delta\in (0,1)$, with probability at least $1-\delta$, \HFUCRL achieves $\Regret(K) = \tilde{O}\left({H d \sqrt{K}} \right)$ for the no observation regime; and the \PlugIn algorithm achieves $\Regret(K) = \tilde{O}\left(H \sqrt{dK}\right)$ for the full observation regime.
\end{corollary}

\subsection{Baseline Algorithms}

To evaluate the performance of our proposed algorithms, we compare against reinforcement learning methods that do not exploit problem structure and state-of-the-art heuristics tailored to the inventory control problem.

\paragraph{\QLearning.}  We benchmark against the standard $Q$-Learning algorithm~\cite{jin2018q}.  While \citet{gong2020provably} develop variants of $Q$-Learning tailored to inventory models with one-sided feedback, where the full demand $\ExoState_h$ is observed whenever the on-hand inventory $\Inventory_h \geq \ExoState_h$, their analysis does not extend to settings with positive lead times where censoring becomes state dependent. A direct application of Theorem~2 in \citet{jin2018q} yields a regret bound of order $O(\sqrt{H^3 d^{L+2}K})$. This dependence on $d^{L}$ reflects a severe curse of dimensionality as the lead time $L$ increases. In contrast, our guarantees in \cref{cor:inventory_control} exploit the exogenous structure of the inventory system to obtain regret bounds that are independent of the lead time.

\paragraph{\PlugIn.} Since the inventory control setting   does not admit rank reduction, we implement a simplified version of \PlugIn that maintains an empirical estimate of the exogenous (demand) distribution and computes an optimal policy for the corresponding plug-in Exo-MDP. Similarly, we additionally compare against \PlugInNH, which maintains a separate empirical estimate of the exogenous distribution at each time step and solves the resulting time-inhomogeneous plug-in problem. By granting full observation of the exogenous state, these algorithms serve as idealized benchmarks and help isolate the statistical cost of learning under censoring.

\paragraph{\HFUCRL, \UCRL.} We compare the time-homogeneous \HFUCRL algorithm, described in \cref{sec:unobs_upper_bound}, to its time-inhomogeneous variant \UCRL, presented in \cref{app:linear_mixture_algorithms}, in order to assess the impact of modeling non-stationarity.

\paragraph{Base-stock policies.} We also compare against a widely-used heuristic for the inventory control problem with positive lead time and lost sales, the so-called {\em base-stock  policies}~\citep{goldberg_survey_2021}.  Intuitively, these policies order a quantity that brings the sum of leftover inventory and outstanding order to some fixed value $\BaseStock$, also referred to as the {\em base-stock level}.  Formally, fixing the base-stock level $\BaseStock$, the action at stage $h$ is given by:
\begin{equation}
\pi_h^b(\Inventory_h, \Order_{h-L}, \ldots, \Order_{h-1}) = \Order_h =
\left(\BaseStock - \Inventory_h - \sum_{i=1}^L \Order_{h-i}\right)^+.
\end{equation}
For scenarios with positive lead times $L > 0$, base-stock policies do not recover the optimal policy.  However, prior work such as~\citet{huh2009asymptotic,zipkin2008old} shows that base-stock policies are optimal when either $L = 0$ or the lost-sales cost $p \rightarrow \infty$. See~\citet{goldberg_survey_2021} for more discussion.  In \cref{tab:performance} we use $\TotalCost_1^{\BaseStock^*}$ to denote the total cost of the best base-stock policy, i.e. the value which achieves $\min_{\BaseStock} C_1^{\pi^b}$ and use $\BaseStock^*$ to denote the optimal base stock level.

\paragraph{Online base-stock algorithm.}
{Since the demand distribution is unknown, we design a simple baseline heuristic algorithm for learning the optimal base-stock policy $\pi^{\BaseStock^\*}$ online. First, we note that the total expected cost is convex in the base-stock level (see \cref{fig:sub1}, appendix). Let $\TotalCost_1^\BaseStock$ denote the $H$-stage total expected cost of the base-stock policy $\pi^\BaseStock$ starting from an initial state with no inventory or outstanding orders, i.e.,
$s_1=(\Inventory_1,\Order_{1-L},\dots,\Order_0)=(0,0,\dots,0)$.
Theorem~8 of \citet{janakiraman2004lost} establishes that, for any demand distribution $\PXi$, the total cost $\TotalCost_1^\BaseStock$ is convex in the base-stock level $\BaseStock$.

Moreover, once the policy $\pi^\BaseStock$ is fixed, each $H$-stage evaluation of the policy in a given episode can be treated as a single sample for its expected $H$-stage total cost $\TotalCost_1^\BaseStock$.  These observations lead to our baseline, the \OnlineBaseStock policy, which applies Algorithm 1 of \citet{agarwal2011stochastic} and searches for the optimal base-stock level using existing work on establishing algorithms for stochastic online convex optimization with bandit feedback.  Moreover, this algorithm amounts to a finite horizon extension to that provided in \citet{agrawal2022learning}.  In \cref{app:online_base_stock} we provide further details, as well as full pseudocode for the algorithm in \cref{alg:convex_opt}.  However, we note that by applying Theorem 1 of \citet{agarwal2011stochastic} this yields an algorithm with a regret guarantee:
\begin{align*}
\sum_{k=1}^K \TotalCost_1^{\BaseStock^k} - \TotalCost_1^{\BaseStock^*}
= \tilde{O}(H \sqrt{K}),
\end{align*}
where we denote $\BaseStock^*$ as the best performing base-stock policy, $ \BaseStock^* = \argmin_{\BaseStock} \TotalCost_1^{\BaseStock}.$}

At first glance, this guarantee seems stronger (scaling independent of $d$).  However, this defines regret relative to the performance of the best-performing base-stock policy (which as discussed above, is not optimal when $L > 0$ or for finite values of the lost-sales cost $p$). Hence, this gap leads to $\Omega(K)$ regret when compared relative to the {\em true} optimal policy\footnote{However, it does allow slightly different assumptions on the demand distribution (notably it directly applies to scenarios where the demand does not have finite support).}.


\begin{figure}[!t]
\caption{In \cref{fig:sub2,fig:sub3} on the $x$-axis we show the episode $k \in [500]$ and on the $y$-axis the total cost $\TotalCost^{\pi^k}$ under the different algorithms.  \cref{fig:sub2} shows results on Scenario I, and \cref{fig:sub3} results for Scenario II.  The grey line corresponds to the performance (total cost) of the optimal base-stock policy, and the black line to the performance of the optimal policy.  We note that under Scenario II there is no optimality gap.}
\label{fig:inventory_control}
\centering

\begin{subfigure}[b]{0.49\textwidth}
\centering
\includegraphics[width=\textwidth]{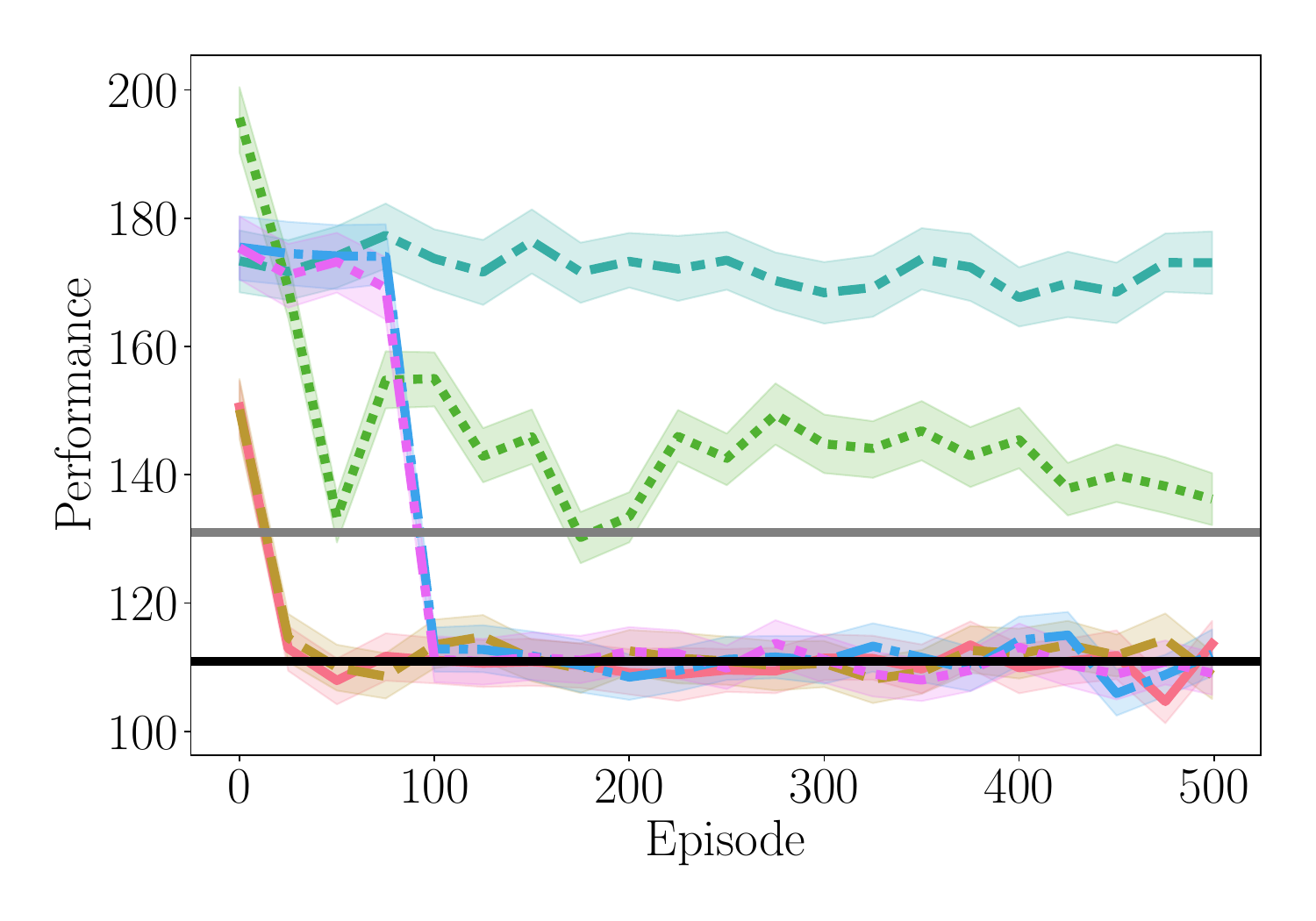}
\caption{$\TotalCost^{\pi^k}$ on {Scenario I}}
\label{fig:sub2}
\end{subfigure}
\hfill
\begin{subfigure}[b]{0.49\textwidth}
\centering
\includegraphics[width=\textwidth]{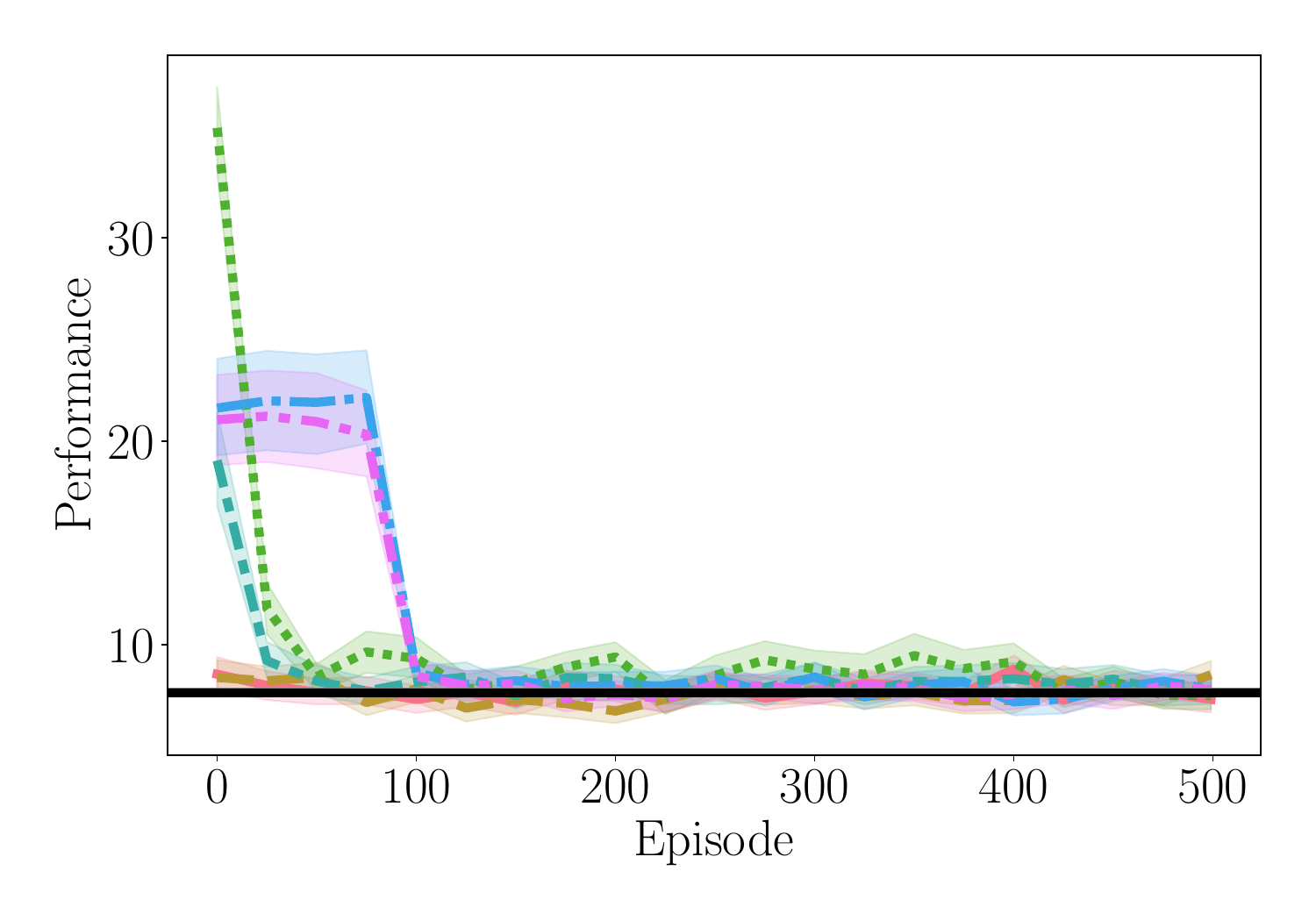}
\caption{$\TotalCost^{\pi^k}$ on {Scenario II}}
\label{fig:sub3}
\end{subfigure}
\begin{subfigure}[b]{\textwidth}
\vspace{.2cm} \centering
\includegraphics[width=0.95\textwidth]{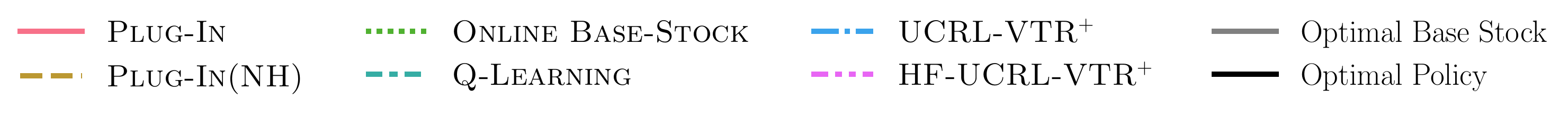}
\label{fig:sub4}
\end{subfigure}
\end{figure}

\subsection{Simulation Results}
In \cref{tab:performance} we compare the performance of our algorithms to the online base-stock algorithm and $Q$-Learning algorithm from \citet{jin2018q} at the final episode $K = 500$.

\paragraph{Description of Scenarios.}
{Scenario I is a time-homogeneous scenario where there is a large optimality gap between the optimal $(\TotalCost_1^*)$ and best performing base-stock policy $(\TotalCost_1^{\BaseStock^*})$. Scenarios II-IV are time-homogeneous scenarios where the best performing base-stock policy is also optimal: Scenario III reflects the effect of a larger dimension $d$ and Scenario IV reflects the effect of a longer horizon $H$. Scenario V is a time-inhomogeneous setting where the base-stock policy is not optimal, and, as we see in the simulation results, adapting time-inhomogeneous algorithms shows significant improvement. Scenarios VI and VII introduce Markovian demand with moderate and strong temporal dependence, respectively, allowing us to evaluate robustness to violations of the independence assumption.  See \cref{app:inventory_scenario_descriptions} for full parameter specifications.}

\paragraph{Empirical Convergence Rates.}
We start by showing the empirical convergence rates of different algorithms for Scenarios~I and~II in \cref{fig:inventory_control} (other scenarios are qualitatively similar and are therefore omitted). We observe that in time-homogeneous environments, \PlugIn, \HFUCRL, and \UCRL converge to the true optimal policy quickly, whereas the \OnlineBaseStock algorithm converges to the optimal base-stock policy. Consequently, in settings where the optimal base-stock policy is suboptimal, \OnlineBaseStock performs strictly worse than our proposed algorithms. We also observe \QLearning performs poorly in Scenario I due to the large state-action space induced by a positive lead time $L$ which slows exploration.  However, its performance improves in Scenario II where $L = 0$ and the state space is smaller.

\paragraph{Optimality of base-stock policy class.}
In Scenario I we observe a substantial optimality gap between the true optimal policy ($\TotalCost_1^*$) and best performing base-stock policy ($\TotalCost_1^{\BaseStock^*}$).  This bias directly propagates to the \OnlineBaseStock algorithm, which although it successfully learns the optimal base-stock level, it nonetheless suffers a gap relative to the true optimal policy.  This example highlights the importance of algorithms that provably converge to the {\em true optimal policy} rather than to a heuristic class, as is commonly considered in the inventory control literature.  {In contrast, each of the procedures $\{$\HFUCRL, \UCRL, \PlugIn\!$\}$ show convergence to the true-optimal policy in this setting.  We also note that \QLearning performs worse than \OnlineBaseStock and significantly worse than \HFUCRL and \UCRL.  This gap arises because \QLearning ignores the low-rank Exo-MDP structure of the problem, leading to excess exploration of the full endogenous state-action space (and theoretically, with regret that grows exponentially in the lead time $L$).}  By contrast, \HFUCRL and \UCRL explicitly exploit the exogenous structure, allowing them to rapidly learn the optimal policy and achieve performance comparable to \PlugIn without requiring full observation of the exogenous state. 
{In Scenarios II to IV} we observe no optimality gap between the best performing base-stock policy and the true optimal policy.  In these regimes, \OnlineBaseStock performs substantially better, as it converges to the true optimal policy rather than a biased heuristic. Nonetheless, \QLearning continues to converge more slowly, reflecting its failure to exploit the Exo-MDP structure.  Even in these favorable settings for base-stock policies, \OnlineBaseStock remains empirically inferior to \HFUCRL, despite \HFUCRL being a more general-purpose algorithm.

Together, these experiments demonstrate the robustness of \HFUCRL and \UCRL across different regimes: they converge to the true optimal policy even when the optimal base-stock policy is suboptimal, while achieving performance comparable to \OnlineBaseStock when a base-stock policy is optimal.

\paragraph{Effect of dimension $d$ and $H$.} Relative to Scenario II, Scenario III has a larger dimension $d$, whereas Scenario IV has a longer horizon $H$.  {From these two comparisons, we see that increasing either $d$ or $H$ has little impact on the relative performance of the algorithms, except for \QLearning, whose performance is worse in these scenarios.}

\paragraph{Effect of non-homogeneity.} In Scenario~V, we consider a time-inhomogeneous environment and observe that the time-inhomogeneous \UCRL algorithm outperforms the time-homogeneous \HFUCRL algorithm. This scenario illustrates the importance of explicitly modeling time inhomogeneity when it is present, as failing to do so can lead to suboptimal policy learning. In contrast, across the remaining time-homogeneous scenarios, \UCRL and \HFUCRL exhibit comparable performance, indicating that the cost of modeling time inhomogeneity in a homogeneous setting is relatively mild.

{\paragraph{Effect of non-independent exogenous process.}
Finally, Scenarios VI and VII evaluate robustness to violations of the independence assumption by introducing Markovian demand with different levels of temporal dependence. In Scenario VI, where the temporal parameter is moderate, we observe little degradation in the performance of our algorithms relative to the optimal policy. In contrast, Scenario VII increases the temporal dependence. Here, the performance of several algorithms deteriorates, with \PlugInNH, \HFUCRL, and \UCRL exhibiting gaps of approximately $10$--$20\%$ relative to the optimal policy. Nonetheless, these methods continue to outperform \OnlineBaseStock and \QLearning. These experiments suggest that the proposed methods are empirically robust to moderate temporal dependence, but their performance can degrade as departures from the independence assumption become larger.}


\section{Conclusion and Future Work}
\label{sec:conclusion}

In this paper, we studied the class of Exo-MDPs, which is defined by a partition of the state space into endogenous and exogenous components. This class is useful for many real-world applications of Markov decision processes (MDPs), in which randomness is induced by exogenous states that are outside the control of the decision maker. Our work highlights that Exo-MDPs, despite their structural assumptions, represent a rich class of MDPs equivalent to both the class of discrete MDPs and discrete linear mixture MDPs. {We fully characterize the minimax-optimal regret rate up to logarithmic factors for both the ``no observation'' setting, in which the exogenous states are completely unobserved, and the ``full observation'' setting, in which the exogenous states are observed, thereby characterizing the $\tilde{\Theta}(\sqrt{r})$ gap due to observing exogenous states. Importantly, all our algorithms achieve regret bounds that scale only with the effective dimension $r$ of the exogenous state, with no dependence on the dimensions of the endogenous state and action spaces.} One interesting open direction is to investigate intermediate observation regimes with sample complexities interpolating between the full- and no-observation regimes, such as the one-sided feedback structure in inventory control models~\citep{gong2024bandits} and the algorithms developed by~\citet{zhang2025reinforcement}. {One might also investigate Exo-MDPs under more general settings, for example when the exogenous state is continuous rather than discrete~\citep{liang2026pure}.}

\ACKNOWLEDGMENT{The authors would like to thank Nic Fishman and Yassir Jedra for insightful discussions about this work. Part of this work was done while Sean Sinclair was a Postdoctoral Associate at MIT. This work was partially supported by the MIT IBM project, the MIT EECS Alumni Fellowship, ONR grant N00014-21-1-2842, and NSF grant DMS-2311072.}


\bibliographystyle{informs2014} 
\bibliography{exo_mdp_ref}

\newpage

\AtBeginEnvironment{APPENDICES}{%
\renewcommand{\theHsection}{appendix.\Alph{section}}%
\renewcommand{\theHsubsection}{\theHsection.\arabic{subsection}}%
\renewcommand{\theHsubsubsection}{\theHsubsection.\arabic{subsubsection}}%
\renewcommand{\theHfigure}{\theHsection.\arabic{figure}}%
\renewcommand{\theHtable}{\theHsection.\arabic{table}}%
\renewcommand{\theHequation}{\theHsection.\arabic{equation}}%
\renewcommand{\theHtheorem}{\theHsection.\arabic{theorem}}%

\crefalias{section}{appendix}%
\crefalias{subsection}{appendix}%
\crefalias{subsubsection}{appendix}%
}

\begin{APPENDICES}


\begin{table}[H]
\caption{List of common notations.}
\label{tab:notation}
\centering
\begin{tabular}{ll}
\toprule \textbf{Symbol} & \textbf{Definition} \\ \midrule
\multicolumn{2}{c}{{\em Problem setting specification}}\\ \hline
$\StateSpace, \ActionSpace, H, \state_1, \transition, \Reward$ & State
and action space, horizon, starting state, transition, reward
\\ $\ExoStateSpace, d$ & Exogenous state space with $d =
|\ExoStateSpace|$ \\ $S_h, \Action_h, \ExoState_h$ & State, action,
exogenous state at stage $h$ \\ $\state_h, \action_h, \exostate_h$ &
Realization of state, action, exogenous state at stage $h$ \\ $\PXi,
\pvecxi$ & Marginal distribution of $\exostate$ and its vector
representation \\ $\SysFun, \RewFun$ & Known transition function and
reward function\\ $\pi$ & Stochastic policies $\pi_h : \StateSpace \rightarrow \distrover(\ActionSpace)$ \\ $K$ & Number of
episodes \\ $\pi^k$ & Policy chosen by learning algorithm at start of
episode $k$ \\ $V_h^\pi(s, \MDP)$ & Value for policy $\pi$ starting in
state $s$ at stage $h$ in MDP $\MDP$ \\ $\exostate_{\geq h}$ & The
vector $(\exostate_h, \ldots, \exostate_H)$ \\ $\Regret(K)$ &
Cumulative loss for policies $\{\pi^k\}_{k \in [K]}$ relative to the
optimal policy $\pi^*$ \\ $\theta$ & Common unknown parameter for the linear mixture MDP representation \\
$\phi_p,\phi_r$ & Known transition and reward feature mappings $\phi_p(s' \mid s,a)$ and $\phi_r(s,a)$ \\
$\Fmattrans,\Fmatrew$ & Transition and reward information matrices \\
$\Fmat$ & Joint information matrix $\Fmat = \begin{pmatrix}\Fmattrans \\ \Fmatrew\end{pmatrix}$ \\
$r$ & Effective dimension $r=\rank(\Fmat)$ of the Exo-MDP \\
\hline \multicolumn{2}{c}{{\em Inventory control
    specification}}\\ \hline $\Inventory_h$ & On-hand inventory at the
start of stage $h$ \\ $\Demand_h$ & Demand at stage $h$ \\ $\Order_h$
& Ordering decision at stage $h$ \\ $\HoldingCost, \LostSales$ &
Holding cost and lost sales penalty\\ $\BaseStock, \BaseStock^*$ &
Base-stock parameter $b$ and optimal base-stock value \\ \hline
\multicolumn{2}{c}{{\em Airline revenue management
    specification}}\\ \hline $K$, $M$ & Number of resources and number
of customer types \\ $A_j, r_j$ & Resource consumption and revenue for
customer type $j$ \\ $B_k$ & Starting capacity for resource $k$
\\ \bottomrule
\end{tabular}
\end{table}


\section{Background for Time-Inhomogeneous MDPs}
\label{app:preliminary}

Our discussion in the main text was limited to time-homogeneous
Exo-MDPs, where the exogenous state is drawn i.i.d.\ from a fixed
distribution $\PXi$ at each stage of the episode.  In this appendix,
we describe the \emph{time-inhomogeneous} or \emph{non-stationary}
extension, in which there may be a different exogenous distribution $\PXih$ for each stage $h \in [H]$.

\paragraph{Model.}
Recall that a (time-homogeneous) Exo-MDP is represented by a tuple \(
\MDP[\PXi,\SysFun,\RewFun] =
(\StateSpace, \ExoStateSpace, \ActionSpace,\Horizon, \state_1, \PXi,
\SysFun, \RewFun), \) where the exogenous state $\ExoState_h$ is drawn
i.i.d.\ from $\PXi$ independent of $(S_h,\Action_h)$, and the
endogenous dynamics and rewards are given by the known deterministic
functions $\SysFun$ and $\RewFun$:
\begin{subequations}
\begin{align}
S_{h+1} &= \SysFun(S_h, \Action_h, \ExoState_h),
&& \SysFun: \StateSpace \times \ActionSpace \times \ExoStateSpace \rightarrow \StateSpace,
\label{eq:nonhom-sysfun}
\\
\Reward_h &= \RewFun(S_h, \Action_h, \ExoState_h),
&& \RewFun: \StateSpace \times \ActionSpace \times \ExoStateSpace \rightarrow [0,1].
\label{eq:nonhom-rewfun}
\end{align}
\end{subequations}

In the time-inhomogeneous Exo-MDP model, we retain the same $\StateSpace$,
$\ActionSpace$, $\ExoStateSpace$, $\SysFun$, and $\RewFun$, but allow the
distribution of the exogenous state to depend on the stage.  Specifically, assume a sequence of exogenous distributions
\(
\{\PXih\}_{h=1}^H,\, \PXih \in \distrover(\ExoStateSpace),
\)
and the exogenous state at each stage is drawn as
\(
\ExoState_h \sim \PXih,\, h \in [H],
\)
independently across $h$ and across episodes, and independent of
$(S_h,\Action_h)$.  Fixing an indexing
$\ExoStateSpace = \{\exostate^j\}_{j=1}^d$, we let
$\pvecxih \in \Delta(\ExoStateSpace)$ denote the probability vector corresponding to
$\PXih$, with entries
\[
[\pvecxih]_j
= \PXih(\ExoState = \exostate^j), \qquad j \in [d], ~ h \in [H].
\]
When $\PXih \equiv \PXi$ for all $h \in [H]$, we recover the time-homogeneous
setting studied in the main body of the paper.

\paragraph{Policies and value functions.}
The definitions of policies, value functions, and optimal value functions are
the straightforward time-inhomogeneous extension of those in
\cref{sec:preliminary}: at each stage $h$, the policy may depend on $h$, and
the Bellman recursion uses the stage-dependent exogenous kernels
$\PXih$.  We omit the restatement for brevity.

\paragraph{Induced time-inhomogeneous MDP.}
Given the exogenous distribution sequence $\{\PXih\}_{h=1}^H$ and the known
functions $\SysFun$ and $\RewFun$, the time-inhomogeneous Exo-MDP induces a
time-inhomogeneous tabular MDP over the endogenous state space
$\StateSpace$ and action space $\ActionSpace$.  For each stage
$h \in [H]$, we define the stage-dependent transition kernel
$\transition_h$ and (expected) reward function $r_h$ as
\begin{subequations}
\begin{align}
\transition_h(\state' \mid \state, \action)
&\coloneqq
\sum_{\exostate \in \ExoStateSpace}
\PXih(\exostate)
\, \indicator_{\SysFun(\state,\action,\exostate) = \state'},
\label{eq:nonhom-P-h}
\\
r_h(\state,\action)
&\coloneqq
\sum_{\exostate \in \ExoStateSpace}
\PXih(\exostate)
\, \RewFun(\state,\action,\exostate).
\label{eq:nonhom-r-h}
\end{align}
\end{subequations}
Equivalently, we can view the time-inhomogeneous Exo-MDP as a time-inhomogeneous
MDP
\begin{align*}
(\StateSpace,\ActionSpace,\Horizon,\state_1,\{\transition_h\}_{h=1}^H,\{\Reward_h\}_{h=1}^H),
\end{align*}
where $\Reward_h(\state,\action)$ is the bounded stochastic reward
at stage $h$ with mean $r_h(\state,\action)$ as in
equation~\eqref{eq:nonhom-r-h}.


\paragraph{Extending results from~\Cref{sec:equivalence_classes}.}

All of the structural results in \Cref{sec:equivalence_classes} extend
directly to the time-inhomogeneous setting.  The proofs of the three
equivalence directions do not rely on time-homogeneity.  The only
change is that the mixture coefficients become stage-dependent
($\pvecxih$), while the deterministic maps $(\SysFun,\RewFun)$ remain
the same.

\paragraph{Effective Dimension.} In the time-inhomogeneous Exo-MDP, the transition and reward expectations
at stage $h$ take the linear form
\begin{align*}
\transition_h(s' \mid s,a) = \phi_p(s' \mid s,a)^\top \pvecxih, \qquad
r_h(s,a) = \phi_r(s,a)^\top \pvecxih,
\end{align*}
where the feature vectors $\phi_p,\phi_r\in\mathbb{R}^d$ are defined
by the deterministic and stage-independent maps $\SysFun$ and
$\RewFun$.

Since the feature representation does not change with $h$, the
effective dimension $r$ defined in the homogeneous setting from these
features remains unchanged in the time-inhomogeneous model.  The only
modification in the linear-mixture reduction is that the unknown
coefficient vectors become stage-dependent:
\begin{align*}
\theta_h = \pvecxih,\qquad h\in[H].
\end{align*}
Thus, the reduction yields a linear mixture MDP with fixed feature
maps but a sequence of stage-dependent parameter vectors
$\{\theta_h\}_{h=1}^H$. 


\section{Learning in Time-Inhomogeneous Exo-MDPs}
\label{app:non_homogeneous}

We next present the regret lower and upper bounds for learning in
time-inhomogeneous Exo-MDPs when the exogenous states are unobserved.
The results mirror those stated in the main text with added details
for handling time-inhomogeneous dynamics. We focus on the setting in which the exogenous states $\ExoState_h$ are not observed. In this case, the learner must infer each
$\pvecxih$ indirectly through the endogenous transitions.  We show
that this introduces an additional $\sqrt{H}$ factor in the regret.
Formally, we establish the following lower bound.

\begin{theorem}
\label{thm:lower_bound_non_homogeneous}
For a given horizon length $H \geq 2$, consider an effective dimension $r \geq
H$ and number of episodes {$K \geq Hr^2$}.  Then for any
learning algorithm, there exists a time-inhomogeneous Exo-MDP $\MDP$
with effective dimension $r$ such that its expected regret over $K$
episodes is lower bounded as $\E[\Regret(K)] \geq c \; H^{3/2} \; r \;
\sqrt{K}$ for a universal constant $c > 0$.
\end{theorem}

\paragraph{Lower bound construction for time-inhomogeneous
  Exo-MDPs.} \label{sec:non_stationary_lower}
Here we give a proof sketch that highlights the intuition behind the
construction for the time-inhomogeneous Exo-MDP, see
\cref{sec:unstationary_lower_bound_H} for full details of the hard
instance and the complete proof.

The construction of the time-inhomogeneous Exo-MDP proceeds in two
parts, each lasting $\frac{H}{2}$ stages. The first $\frac{H}{2}$
stages entail no reward, but require the agent to correctly guess the
exogenous state distribution which is drawn uniformly from a
collection of $\frac{H}{2}$ different unknown options. The second
$\frac{H}{2}$ stages use the same trick as the instance for
time-homogeneous Exo-MDPs to repeat this reward $\frac{H}{2}$
times. The added factor of $\sqrt{H}$ is due to the additional
hardness of guessing $\frac{H}{2}$ exogenous state distributions
instead of a single distribution.

Next, analogous to the homogeneous case in
\cref{sec:unobs_upper_bound}, we adapt the time-inhomogeneous version
of the \UCRL algorithm of \citet{zhou2021nearly} (a time-varying
extension of the \HFUCRL algorithm).  This yields the following upper
bound.

\begin{theorem}
\label{thm:linear_mixture_mdp_low_rank_non_homogeneous}
For any $H$-horizon Exo-MDP with effective dimension $r$, applying a
rank-reduced \UCRL algorithm over $K$ episodes yields a sequence of
policies $\{\pi^k\}_{k=1}^K$ with regret at most
\begin{align}
\Regret(K) & = \tilde{O}\left( \sqrt{r^2 H^2 + rH^3} \sqrt{KH} + r^2
H^3 + r^3 H^2 \right).
\end{align}
\end{theorem}
For sufficiently large $K$, the leading term reduces to
$\tilde{O}(H^{3/2} r \sqrt{K})$, matching the lower bound in
\cref{thm:lower_bound_non_homogeneous} up to polylogarithmic factors.
See \cref{app:linear_mixture_algorithms} for a description of the
\UCRL algorithm, and \cref{app:proofs_unobserved_upper} for the proof.

Hence, relative to the homogeneous case, the regret in the
time-inhomogeneous setting deteriorates by exactly an additional
factor of $\sqrt{H}$, reflecting the need to estimate $H$ distinct
stage-dependent mixture vectors rather than a single exogenous
distribution.


\section{Linear Mixture MDP Algorithms}
\label{app:linear_mixture_algorithms}

{This appendix presents the rank-reduced adaptations of HF-UCRL-VTR+ and UCRL-VTR+ to Exo-MDPs under the no-observation regime. \Cref{app:HF_UCRL_VTR_algo} addresses the time-homogeneous case, while \cref{app:UCRL_VTR_algo} presents addresses the time-inhomogeneous case.}

\subsection{Complete Algorithm of \HFUCRL~\citep{zhou2022computationally} (\cref{sec:unobs_upper_bound})}
\label{app:HF_UCRL_VTR_algo}

{The rank-reduced \HFUCRL algorithm can be applied to an Exo-MDP by viewing it as an $r$-dimensional linear mixture MDP, where $r=\rank(\Fmat)$ is the effective dimension of the joint information matrix. In this representation, the reduced transition and reward features are $\tilde\phi_p(s'\mid s,a),\tilde\phi_r(s,a)\in\RR^r$, and both depend on the same unknown parameter $\theta\in\RR^r$.} At each episode, the algorithm builds an ellipsoidal confidence set for $\theta$ using variance-aware, value-targeted ridge regression. It then performs optimistic dynamic programming over this confidence set and follows the greedy policy. Repeating this over episodes gives the regret bound in \cref{thm:linear_mixture_mdp_low_rank}. The subroutine \textsc{HOME} refers to Algorithm 3 of~\citet{zhou2022computationally}, which sets the variance-aware regression weights using a high-order moment estimator.

\begin{algorithm}[t]
\caption{Rank-Reduced HF-UCRL-VTR+ for Time-Homogeneous Exo-MDPs}
\label{alg:hf-ucrl-exo}
\begin{algorithmic}[1]
\Require Horizon $H$, number of episodes $K$, effective dimension $r$, reduced features $\tilde\phi_p(s'\mid s,a),\tilde\phi_r(s,a)\in\RR^r$, regularization parameter $\lambda>0$, norm bound $B$, confidence radii $\{\beta_k\}_{k\ge1}$, and variance-aware weights produced by \textsc{HOME}
\State Initialize $\Sigma_1\gets\lambda I_r$, $b_1\gets0$, and $\widehat\theta_1\gets0$.
\For{$k=1,2,\dots,K$}
\State Set $V_{k,H+1}(s)\gets0$ for all $s\in\StateSpace$.
\Statex \hspace{-1.2em}\textbf{Planning: optimistic dynamic programming}
\For{$h=H,H-1,\dots,1$}
\ForAll{$(s,a)\in\StateSpace\times\ActionSpace$}
\State $\psi_{k,h}(s,a)\gets\tilde\phi_r(s,a)+\sum_{s'\in\StateSpace}\tilde\phi_p(s'\mid s,a)V_{k,h+1}(s')$.
\State $\mathrm{bonus}_{k,h}(s,a)\gets\beta_k\|\psi_{k,h}(s,a)\|_{\Sigma_k^{-1}}$, where $\|x\|_{\Sigma_k^{-1}}=\sqrt{x^\top\Sigma_k^{-1}x}$.
\State $\widehat Q_{k,h}(s,a)\gets\big[\langle\widehat\theta_k,\psi_{k,h}(s,a)\rangle+\mathrm{bonus}_{k,h}(s,a)\big]_{[0,H-h+1]}$.
\EndFor
\State $\pi_{k,h}(s)\in\arg\max_{a\in\ActionSpace}\widehat Q_{k,h}(s,a)$ for all $s$.
\State $V_{k,h}(s)\gets\widehat Q_{k,h}(s,\pi_{k,h}(s))$ for all $s$.
\EndFor
\Statex \hspace{-1.2em}\textbf{Execution and value-targeted regression}
\State $\Sigma_{k+1}\gets\Sigma_k$ and $b_{k+1}\gets b_k$.
\State Reset to the initial state $S_{1,k}=s_1$.
\For{$h=1,2,\dots,H$}
\State Take action $A_{h,k}\gets\pi_{k,h}(S_{h,k})$ and observe $S_{h+1,k}$ and reward $R_{h,k}$.
\State $\psi_{k,h}\gets\tilde\phi_r(S_{h,k},A_{h,k})+\sum_{s'\in\StateSpace}\tilde\phi_p(s'\mid S_{h,k},A_{h,k})V_{k,h+1}(s')$.
\State {$y_{k,h}\gets R_{h,k}+V_{k,h+1}(S_{h+1,k})$.}
\State Query \textsc{HOME} to obtain a variance-aware weight $w_{k,h}$ using the history of features and value targets. {Set $w_{k,h}$ to $H^{-2}$ times the weight assigned to the normalized
feature--target pair $(\psi_{k,h}/H,y_{k,h}/H)$.}
\State $\Sigma_{k+1}\gets\Sigma_{k+1}+w_{k,h}\psi_{k,h}\psi_{k,h}^\top$ and $b_{k+1}\gets b_{k+1}+w_{k,h}y_{k,h}\psi_{k,h}$.
\EndFor
\State $\widehat\theta_{k+1}\gets\Sigma_{k+1}^{-1}b_{k+1}$.
\EndFor
\end{algorithmic}
\end{algorithm}

{\noindent\textbf{Adjustments to HOME.}
Algorithm~2 includes the observed
reward in the regression target, whose conditional mean is therefore
$\psi_{k,h}^{\top}\theta$ under the common reduced parameterization.
We follow the regression and HOME procedures in Algorithms~2--3 of
\citet{zhou2022computationally} with continuation values normalized
by $H$, using $(\psi_{k,h}/H,y_{k,h}/H)$ for the regression that
computes $\widehat\theta_k$ and leaving the auxiliary moment targets
otherwise unchanged.
The weight $w_{k,h}$ is $H^{-2}$ times the weight assigned to this
normalized regression pair in their Eq.~(5.1), computed from
pre-transition statistics.\footnote{{Normalize the reduced features
so that $\|\widetilde\phi_r(s,a)\|_2\le1$ and
$\|\sum_{s'}\widetilde\phi_p(s'\mid s,a)v(s')\|_2\le1$
for every $v:\mathcal S\to[0,1]$, and let $B$ bound
$\|\theta\|_2$. Set
$M=\lceil\log_2(3KH)\rceil$, $\alpha=H^{-1/2}$, $\gamma=1$,
and $\lambda=r/B^2$. Use their Eq.~(5.3) in dimension $r$ with
failure parameter $\delta/(2M+1)$, multiplying its first square-root
term by $\sqrt3$.}}
The factor $H^{-2}$ converts the normalized regression weight to the
units used in Algorithm~2. Separately, bounded rewards imply that the
additional reward-containing second-moment terms are of order
$H^{-1}$ after normalization. The chosen modifications of the confidence radius accommodate
these terms while preserving the continuation-moment analysis and
the regret order stated in Theorem~3.
}

\subsection{Complete Algorithm of \UCRL~\citep{zhou2021nearly} (\cref{app:non_homogeneous})}
\label{app:UCRL_VTR_algo}

{\citet{zhou2021nearly} analyze time-inhomogeneous linear mixture MDPs in which the transition kernel at each stage admits a linear representation with a stage-dependent parameter. In our rank-reduced Exo-MDP representation, the joint information matrix $\Fmat$ has rank $r$, and the stage-dependent exogenous distribution $\pvecxih$ induces the reduced parameter $\theta_h=B^\top\pvecxih\in\RR^r$. The transition and reward models can therefore be written using the reduced features $\tilde\phi_p(s'\mid s,a),\tilde\phi_r(s,a)\in\RR^r$ as
\[
\transition_h(s'\mid s,a)=\tilde\phi_p(s'\mid s,a)^\top\theta_h,\qquad r_h(s,a)=\tilde\phi_r(s,a)^\top\theta_h.
\]
Thus, after rank reduction, the relevant linear-mixture dimension is the effective dimension $r=\rank(\Fmat)$ rather than the ambient exogenous dimension $d$.

\noindent\textbf{Details of the \UCRL algorithm.} At a high level, the \UCRL algorithm is an optimistic model-based method that repeatedly performs the following sequence of estimations. For each episode and stage, it first uses weighted ridge regression to learn the underlying stage-dependent parameter $\theta_h$. In an Exo-MDP, this parameter is the reduced projection $B^\top\pvecxih$ of the stage-$h$ exogenous distribution. It then constructs an ellipsoid confidence set in the parameter space centered around the estimated parameter according to a Bernstein-type self-normalized concentration inequality which, with high probability, contains the true parameter. Using the confidence set, the algorithm then constructs an optimistic estimate of the action-value function by solving the Bellman equations. Solving for the estimated action-value function gives the estimated policy.
}

{The pseudocode adapts their Algorithm~2 to unknown mean rewards by using the combined feature $\psi_{k,h}=\tilde\phi_r+\phi_{V_{k,h+1}}$ and target $R_h^k+V_{k,h+1}(S_{h+1}^k)$ in the regression. 
We normalize the reduced features so that $\|\tilde\phi_r(s,a)\|_2\le1$, $\left\|\sum_{s'}\tilde\phi_p(s'\mid s,a)v(s')\right\|_2\le1$.  In the confidence radii of Eqs.~(14)--(15) of \citet{zhou2021nearly}, replace the dimension by $r$.
}

\begin{algorithm}[!t]
\caption{\UCRL for Episodic Linear Mixture MDPs}
\label{alg:ucrl_vtr}
\begin{algorithmic}[1]
\Require Regularization parameter $\lambda$, upper bound $B$ on $\|\theta_h^\ast\|_2$

\State \textbf{Initialization:}
\For{$h \in [H]$}
\State $\widehat{\Sigma}_{1,h} \gets \lambda I$, \quad
$\tilde{\Sigma}_{1,h} \gets \lambda I$
\State $\widehat{b}_{1,h} \gets 0$, \quad
$\tilde{b}_{1,h} \gets 0$
\State $\widehat{\theta}_{1,h} \gets 0$, \quad
$\tilde{\theta}_{1,h} \gets 0$
\EndFor

\For{$k = 1,2,\dots,K$}
\State $V_{k,H+1}(\cdot) \gets 0$
\For{$h = H,H-1,\dots,1$}
\State {$Q_{k,h}(\cdot,\cdot) \gets \Bigl[ \langle \widehat{\theta}_{k,h},\psi_{k,h}(\cdot,\cdot)\rangle +\widehat{\beta}_k \bigl\|\widehat{\Sigma}_{k,h}^{-1/2}\psi_{k,h}(\cdot,\cdot)\bigr\|_2
\Bigr]_{[0,H-h+1]}$}
\State $\pi_h^k(\cdot) \gets
\arg\max_{a \in \ActionSpace} Q_{k,h}(\cdot,a)$
\State $V_{k,h}(\cdot) \gets
\max_{a \in \ActionSpace} Q_{k,h}(\cdot,a)$
\EndFor

\For{$h = 1,2,\dots,H$}
\State Take action $\Action_h^k \gets \pi_h^k(S_h^k)$, observe
$S_{h+1}^k \sim \transition_h(\cdot \mid S_h^k,\Action_h^k)$
\State {$\psi_{k,h}\gets\psi_{k,h}(S_h^k,\Action_h^k)$ and $y_{k,h}\gets R_h^k+V_{k,h+1}(S_{h+1}^k)$.}
\State $\overline{\sigma}_{k,h} \gets
\sqrt{\max\Bigl\{H^2/{r},\,
[\overline{\mathbb{V}}_{k,h}V_{k,h+1}](S_h^k,\Action_h^k)
+ E_{k,h}\Bigr\}}$



\State {$\widehat{\Sigma}_{k+1,h} \gets \widehat{\Sigma}_{k,h} + \overline{\sigma}_{k,h}^{-2}\psi_{k,h}\psi_{k,h}^\top$.}
\State {$\widehat{b}_{k+1,h} \gets \widehat{b}_{k,h} + \overline{\sigma}_{k,h}^{-2}\psi_{k,h}y_{k,h}$.}

\State $\tilde{\Sigma}_{k+1,h} \gets
\tilde{\Sigma}_{k,h}
+ \phi_{V_{k,h+1}^2}(S_h^k,\Action_h^k)
\phi_{V_{k,h+1}^2}(S_h^k,\Action_h^k)^\top$

\State $\tilde{b}_{k+1,h} \gets
\tilde{b}_{k,h}
+ \phi_{V_{k,h+1}^2}(S_h^k,\Action_h^k)
V_{k,h+1}^2(S_{h+1}^k)$

\State $\widehat{\theta}_{k+1,h} \gets
\widehat{\Sigma}_{k+1,h}^{-1}\widehat{b}_{k+1,h}$

\State $\tilde{\theta}_{k+1,h} \gets
\tilde{\Sigma}_{k+1,h}^{-1}\tilde{b}_{k+1,h}$
\EndFor
\EndFor
\end{algorithmic}
\end{algorithm}

\noindent
The quantities used in \cref{alg:ucrl_vtr} are defined as:
\begin{align*}
\phi_V(s,a)
&= \sum_{s' \in \StateSpace} {\tilde{\phi}_p}(s' \mid s,a)\,V(s'),\\
{\psi_{k,h}(s,a)}&={\tilde\phi_r(s,a)+\phi_{V_{k,h+1}}(s,a),}\\
[\transition_h V](s,a)
&= \mathbb{E}_{S' \sim \transition_h(\cdot \mid s,a)} V(S'),\\
[\mathbb{V}_h V](s,a)
&= [\transition_h V^2](s,a)
- \bigl([\transition_h V](s,a)\bigr)^2,\\[4pt]
\widehat{\beta}_k
&= 8 \sqrt{{r} \log(1 + k/\lambda)\,\log(4k^2 H/\delta)}
+ 4 \sqrt{{r}}\,\log(4k^2 H/\delta)
+ \sqrt{\lambda}\,B,\\[4pt]
{\widehat q_{k,h}}&= {\bigl[\langle \phi_{V_{k,h+1}^2}(S_h^k,\Action_h^k),
\tilde{\theta}_{k,h} \rangle\bigr]_{[0,H^2]}
- \bigl[\langle
\psi_{k,h},
\widehat{\theta}_{k,h}\rangle\bigr]_{[0,H]}^2}\\
E_{k,h}
&= \min\Bigl\{H^2,\,
2H\,\breve{\beta}_k
\bigl\|\widehat{\Sigma}_{k,h}^{-1/2}
{\psi_{k,h}}
\bigr\|_2
\Bigr\}  + \min\Bigl\{H^2,\,
\tilde{\beta}_k
\bigl\|\widehat{\Sigma}_{k,h}^{-1/2}
\phi_{V_{k,h+1}^2}(S_h^k,\Action_h^k)\bigr\|_2
\Bigr\},\\[4pt]
\breve{\beta}_k
&= 8 {r} \sqrt{\log(1 + k/\lambda)\,\log(4k^2 H/\delta)}
+ 4 \sqrt{{r}}\,\log(4k^2 H/\delta)
+ \sqrt{\lambda}\,B,\\[4pt]
\tilde{\beta}_k
&= 8 \sqrt{{r} H^4 \log \bigl(1 + k H^4 / ({r} \lambda)\bigr)\,
\log(4k^2 H/\delta)}
+ 4 H^2 \log(4k^2 H/\delta)
+ \sqrt{\lambda}\,B,
\end{align*}
where $[\cdot]_{[a,b]}$ denotes clipping to the interval $[a,b]$. {The quantity $\widehat q_{k,h}$ is obtained from Eq.~(13) of \citet{zhou2021nearly} by replacing $\phi_{V_{k,h+1}}$ with $\psi_{k,h}$ in the squared-mean term, which we use to construct the regression scale.}

{The scale based on $\widehat q_{k,h}$ is enlarged to cover the noise from $R_h^k$. This changes the cumulative variance bound only by constant factors and an $O(KH^2)$ term. By Eq.~(66) of \citet{zhou2021nearly}, this added term contributes $\tilde{O}(r\sqrt{H\cdot KH^2})=\tilde{O}(rH^{3/2}\sqrt{K})$ to regret. This is bounded by the leading term $\tilde{O}(\sqrt{r^2H^2+rH^3}\sqrt{KH})$ in our time-inhomogeneous upper bound (\cref{thm:linear_mixture_mdp_low_rank_non_homogeneous}).}

\clearpage

\section{Remarks on the Exo-MDP setup.}

\subsection{Remarks on Fixed Single Starting State}
\label{sec:fixed_starting_state}

Note that we defined the MDP and Exo-MDP models with a fixed initial state. This simplifies the technical arguments in this paper but is not a restrictive assumption since we can capture MDPs and Exo-MDPs with random initial states by the following modification. Suppose the initial state is instead a random variable following distribution $s_1\sim \initialdistr$.  Then we can add a {\em dummy} initial state $\state_0$ with $\Reward(\state_0,\action)=0$, $\forall \action\in\ActionSpace$, $\transition(\state_1\mid \state_0, \action)=\initialdistr(\state_1)$, $\forall \action\in\ActionSpace$ and $\transition(\state_0\mid \state, \action)=0$, $\forall (\state, \action) \in\StateSpace \times \ActionSpace$. In the case of MDPs, $\mu$ can be any distribution over $\StateSpace$. 
In the case of Exo-MDPs, $\mu$ can be any distribution captured by $\PXi$. Specifically, the probability vector $\pvec_{\initialdistr} = (\initialdistr(\state_1=\state^1), \dots,\initialdistr(\state_1=\state^{|\StateSpace|})) \in [0,1]^{|\StateSpace|}$ corresponding to the multinomial distribution $\initialdistr$, must be expressed as $\pvec_{\initialdistr} = \Mmat \pvecxi$ where $\Mmat\in \{0,1\}^{|\StateSpace|\times |\ExoStateSpace|}$ is such that $\Mmat^\top \indvec_{|\StateSpace|} = \indvec_{|\ExoStateSpace|}$, where $\indvec_{|\StateSpace|}$ and $\indvec_{|\ExoStateSpace|}$ are the vectors of all $1$'s of dimensions $|\StateSpace|$ and $|\ExoStateSpace|$. Note that this implicitly imposes that the support of $\initialdistr$ is at most $|\ExoStateSpace| = d$.

\subsection{Regret to Value Function Estimation Error Conversion}\label{app:sec:regret_val_conversion}

The two different performance metrics, \emph{regret} and \emph{value
function estimation error}, are closely
connected~\cite{jin2018q}. Specifically, for any sequence of policies
$\{\policy^k\}^K_{k = 1}$, one can construct a policy $\pi$ such that
its value function estimation error is upper bounded by $\frac{1}{K}$
times the regret of the sequence of policies $\{\policy^k\}^K_{k =
1}$.

\begin{lemma}
\label{lem:online_batch}
Given an online algorithm generating a sequence of policies
$\{\pi^k\}_{k=1}^K$, let policy $\policyhat$ be a uniform draw from
$\{\pi^k\}_{k=1}^K$. Suppose the algorithm achieves cumulative online
regret $\Regret(K)\leq \rho(\StateSpace,\ActionSpace,\Horizon)
K^{1-\alpha}$, where $\rho$ is some fixed function of
$\StateSpace,\ActionSpace,\Horizon$.  Then $\policyhat$ achieves value
function estimation error:
$$\E[\ValFunOpt_1(\state_1)-\ValFun_1^{\policyhat}(\state_1)] \leq
\rho(\StateSpace,\ActionSpace,\Horizon) K^{-\alpha}.$$
\end{lemma}

\begin{rproof}
By definition of $\policyhat$ we have:
\begin{align*}
\E[V_1^*(s_1) - V_1^{\policyhat}(s_1)] & = \frac{1}{K} \sum_{k=1}^K
V_1^*(s_1) - \frac{1}{K} \sum_{k=1}^K V_1^{\pi^k}(s_1) =
\frac{1}{K} \left[ \sum_{k=1}^K (V_1^*(s_1) -
V_1^{\pi^k}(s_1))\right] \\ & = \frac{1}{K} \Regret(K) \leq
\frac{1}{K} \rho(\StateSpace,\ActionSpace,\Horizon) K^{1-\alpha}.
\end{align*}
\end{rproof}

\section{Proofs for Unobserved Exogenous States}
\label{app:proofs_unobserved}

\subsection{Lower Bounds}
\label{app:unobs_lower}

\subsubsection{Proof of \cref{thm:lower_bound}}
\label{app:lower_bound_proof}

\noindent We present our lower bound constructions in two steps. We
start by constructing an Exo-Bandit (an Exo-MDP with $H=1$) achieving
a lower bound of $\Omega(r\sqrt{K})$ on the expected regret. We then
use the Exo-Bandit as a building block for an Exo-MDP with horizon $H$
to get a lower bound of $\Omega(Hr\sqrt{K})$.

\paragraph{Step 1: the ``Exo-Bandit'' problem with lower bound $\Omega(r\sqrt{K})$.}\label{sec:linear_bandit_lower_bound}
Our lower bound construction builds upon the hardness of learning a
single-horizon Exo-MDP. We first show a specific construction with
$H=1$ which reduces to learning a linear bandit on a hypercube action
set. We call this construction the ``Exo-Bandit'' problem
$\bandit[\Tilde{Z}^*]$ parameterized by a vector
$\Tilde{Z}^*\in\{-1,1\}^{r}$.

\paragraph{Setup of the ``Exo-Bandit" problem
$\bandit[\Tilde{Z}^*]$.}
Let $r$ be given and set $d := 2r$. We consider the
exogenous state space as $\ExoStateSpace = [d] = \{1,2,\dots,d\}$.  We
define the linear bandit problem $\bandit[\Tilde{Z}^*]=
(\ActionSpace,\Reward)$ parameterized by a vector
$\Tilde{Z}^*\in\{-1,1\}^{r}$ with action space $\ActionSpace$ and
reward dynamics $\Reward$ as follows.  Note that since the horizon $H
= 1$, there is no notion of state and we omit it in the following
definition.

The action set $\ActionSpace$ sits on a subset of the $d$-dimensional
hypercube,
\[
\ActionSpace = \{([Z]_1,-[Z]_1,[Z]_2,-[Z]_2,\dots,
[Z]_{r}, -[Z]_{r})\mid [Z]_i\in\{-1,1\}\} \subset
\{-1,1\}^{d}.
\]
Note that each action $a\in \ActionSpace$ is completely
characterized by a vector $Z\in\{-1,1\}^{r}$, where
\[
a(Z)=([Z]_1,-[Z]_1,[Z]_2,-[Z]_2,\dots, [Z]_{r},
-[Z]_{r}).
\]
The (unknown) distribution $\PXi$ for the exogenous state
$\exostate$, parameterized by $\Tilde{Z}\in \{-1,1\}^{r}$, is given by
\begin{align*}
\pvecxi(\Tilde{Z}) = (\PXi(1),\dots, \PXi(d)) = \left(\frac{1}{d}+c
[\Tilde{Z}]_1, \frac{1}{d}-c [\Tilde{Z}]_1,\dots, \frac{1}{d}
+c [\Tilde{Z}]_{r},\frac{1}{d}-c
[\Tilde{Z}]_{r}\right),
\end{align*}
where constant $c=\frac{1}{10}\sqrt{\frac{2}{5K}}$. In other words, $\pvecxi$ is almost a uniform distribution over $[d]$ except each coordinate is perturbed from $\frac{1}{d}$ by a small constant $c$ or $-c$ depending on the value of $[\Tilde{Z}]_i$.
The reward function $\Reward$ is given by $\Reward(a) = \RewFun(a,
\exostate) = [a]_{\exostate}$ for $\exostate= 1,2,\dots,d$, that is,
$\RewFun(a, \exostate)$ takes value of the $\exostate$-th coordinate
of the action vector $a$. Equivalently, $R(a) = [a]_j$ with
probability $[\pvecxi]_j$.

\paragraph{Regret lower bound for the Exo-Bandit problem.}  We now
show that for any algorithm $\algo$, there exists some hard instance
$\bandit[\Tilde{Z}^*]$ of the ``Exo-Bandit" problem such that the
expected regret over $K$ time steps is lower bounded by
$\Omega(r\sqrt{K})$.

\begin{lemma}
\label{lem:horizon_1_lower}
Assume $K \geq \frac{2}{5}r^2$, and let $c =
\frac{1}{10}\sqrt{\frac{2}{5K}}$.  For any bandit algorithm $\algo$,
there exists a linear bandit $\bandit[\Tilde{Z}^*]$ parameterized by
$\Tilde{Z}^* \in \{-1,1\}^{r}$ with corresponding exogenous
state distribution
\[
\pvecxi(\Tilde{Z}^*)=\Big(\tfrac{1}{d}+c
[\Tilde{Z}^*]_1, \tfrac{1}{d}-c [\Tilde{Z}^*]_1,\dots, \tfrac{1}{d}+c
[\Tilde{Z}^*]_{r}, \tfrac{1}{d}-c
[\Tilde{Z}^*]_{r}\Big)
\]
such that the expected regret of $\algo$
over $K$ time steps on the bandit is lower bounded by
\begin{align*}
\E_{\ExoState \sim \pvecxi(\Tilde{Z}^*)}[\Regret(K)]\geq \gamma
r \sqrt{K}\quad \text{where $\gamma$ is a universal constant}.
\end{align*}
\end{lemma}

\begin{rproof}
We first show that the Exo-Bandit is a linear bandit where the reward
follows a shifted Bernoulli distribution, and the probability
parameter is a linear function of $a$ and $\PXi$.  Specifically, for
any time step, the reward for taking action $a = a(Z) \in \{-1,
1\}^{d}$ is given by $\RewFun(a,\exostate) = [a]_\exostate$. So
$R(a)\in\{-1,1\}$, and the probability of getting reward $1$ is given
by
\begin{align*}
p_1 = \Pr[R(a(Z)=1] &= \sum_{i=1}^{r} \frac{1}{d} + c
\sum_{i=1}^{r} \indicator_{[Z]_i = \sign([\Tilde{Z}]_i)} - c
\sum_{i=1}^{r} \indicator_{[Z]_i = -\sign([\Tilde{Z}]_i)}\\ &=
\left(\frac{1}{2} - \frac{c d}{2}\right) + 2c \langle Z,\Tilde{Z}
\rangle = \delta + 2c \langle Z,\Tilde{Z} \rangle
\end{align*}
where $\delta \coloneqq \frac{1}{2} - \frac{c d}{2}$.

We set $c=\frac{1}{10}\sqrt{\frac{2}{5K}}$, and suppose $K\geq
\frac{2}{5}r^2$.  Since the reward $\Reward(a) \in \{-1, 1\}$, the
reward distribution given action $a$ follows a shifted Bernoulli
distribution
\begin{align*}
\Reward(a)\stackrel{\textbf{d}}{=} -1+ 2 \bern(p_1)\quad \text{where
$\stackrel{\textbf{d}}{=}$ denotes equal in distribution.}
\end{align*}

The problem now reduces to showing a lower bound of regret on linear
bandits with hypercube action set and Bernoulli reward with linear
mean payoff. To prove the regret lower bound for this linear bandit,
we largely follow the arguments in Lemma C.8 of~\cite{zhou2021nearly},
with adaptations to our Exo-Bandit setting.

Let $a_k=a(Z_k)\in\ActionSpace$ for $k\in [K]$ denote an action chosen
at the $k$-th episode. Then for any $\pvecxi(\Tilde{Z})$ and
$\tilde{Z}\in\{-1,1\}^{r}$, the expected regret $\EE_{\ExoState\sim
\pvecxi(\Tilde{Z})}[\Regret(K)]$ by taking action sequence
$a_1=a(Z_1),\dots, a_K=a(Z_K)$ in Exo-Bandit $\bandit[\tilde{Z}]$ is
given by
\begin{align*}
\EE_{\ExoState\sim \pvecxi(\Tilde{Z})}[\Regret(K)] &= \sum_{k=1}^K
\EE_{\ExoState\sim \pvecxi(\Tilde{Z})} \left[ 2c
\left(\max_{Z\in\{-1,1\}^{r}}\langle \tilde{Z}, Z \rangle -
\langle \tilde{Z_k}, Z \rangle \right)\right]\\ &= 2c
\sum_{k=1}^K \sum_{j=1}^{r}\EE_{\ExoState\sim
\pvecxi(\Tilde{Z})}\left[\indicator_{\sign[\tilde{Z}]_j \neq
\sign[Z_k]_j}\right]\\ &= 2c \sum_{j=1}^{r}\sum_{k=1}^K
\EE_{\ExoState\sim
\pvecxi(\Tilde{Z})}\left[\indicator_{\sign[\tilde{Z}]_j \neq
\sign[Z_k]_j}\right]
\end{align*}

Let $N_j(\tilde{Z}) \coloneqq \sum_{k=1}^K
\indicator_{\sign[\tilde{Z}]_j \neq \sign[Z_k]_j}$, which denotes the
number of mistakes made on the $j$-th coordinate of $\tilde{Z}$ over
$K$ episodes. Let $\tilde{Z}^j\in \{-1,1\}^{r}$ denote the vector
which flips the sign of the $j$-th coordinate of vector
$\tilde{Z}$. Note that $N_j(\tilde{Z}) + N_j(\tilde{Z}^j) = K$ for any
$\tilde{Z}\in \{-1,1\}^{r}$. Let $\Pp_{\tilde{Z}}$ be the
distribution over $\Reward_1,\dots, \Reward_K$ induced by executing
$a_1,\dots, a_K$ on bandit $\bandit[\tilde{Z}]$.  Therefore,
\begin{align*}
& 2\sum_{\Tilde{Z}\in \{-1,1\}^{r}}\EE_{\ExoState\sim
\pvecxi(\Tilde{Z})}[\Regret(K)] \\
= & 2c \sum_{\Tilde{Z}\in
\{-1,1\}^{r}} \sum_{j=1}^{r}\left( \EE_{\ExoState\sim
\pvecxi(\Tilde{Z})} [N_j(\tilde{Z})] + \EE_{\ExoState\sim
\pvecxi(\Tilde{Z}^j)} [N_j(\tilde{Z}^j)] \right) \\
= & 2c \sum_{\Tilde{Z}\in \{-1,1\}^{r}} \sum_{j=1}^{r}\left( K +
\EE_{\ExoState\sim \pvecxi(\Tilde{Z})} [N_j(\tilde{Z})] -
\EE_{\ExoState\sim \pvecxi(\Tilde{Z}^j)} [N_j(\tilde{Z}^j)]
\right)\\
\geq & 2c\sum_{\Tilde{Z}\in \{-1,1\}^{r}} \sum_{j=1}^{r}\left( K -
\sqrt{1/2}K \sqrt{\mathrm{KL}(\Pp_{\tilde{Z}}, \Pp_{\tilde{Z}^j})}\right),
\end{align*}
where the last inequality is by Pinsker inequality. We proceed to
decompose $\mathrm{KL}(\Pp_{\tilde{Z}}, \Pp_{\tilde{Z}^j})$ by the chain rule
of relative entropy and upper bound it as follows.
\begin{align*}
\mathrm{KL}(\Pp_{\tilde{Z}}, \Pp_{\tilde{Z}^j}) &= \sum_{k=1}^K
\EE_{\ExoState\sim
\pvecxi(\Tilde{Z})}[\mathrm{KL}(\Pp_{\tilde{Z}}(\Reward_k\mid
\Reward_{1:k-1}),\Pp_{\tilde{Z}^j}(\Reward_k\mid \Reward_{1:k-1}))]
\\
& = \sum_{k=1}^K \EE_{\ExoState\sim
\pvecxi(\Tilde{Z})}[\mathrm{KL}(\bern(\delta + 2c\langle Z_k, \tilde{Z}
\rangle),\bern(\delta + 2c\langle Z_k, \tilde{Z}^j
\rangle))] \\
& \leq \sum_{k=1}^K \EE_{\ExoState\sim
\pvecxi(\Tilde{Z})}\left[\frac{20(2c\langle Z_k, \tilde{Z}-
\tilde{Z}^j \rangle)^2}{\delta+2c\langle Z_k, \tilde{Z}
\rangle}\right] \\
& \leq K \frac{20 (2c)^2}{\frac{\delta}{2}} =
\frac{160 K c^2}{\delta}.
\end{align*}
The first inequality is due to the fact that $\delta+2c\langle Z_k,
\tilde{Z} \rangle\leq \delta + cd \leq \frac{3}{5}$, since for any two
Bernoulli distributions $\bern(a)$ and $\bern(b)$, $\mathrm{KL}(\bern(a),
\bern(b))\leq \frac{20(a-b)^2}{a}$ when $a,b\in (0,\frac{3}{5})$.  The
second inequality uses the following facts.  For the numerator,
because $\tilde{Z}$ and $\tilde{Z}^j$ differ only in the $j$-th
coordinate, $|\langle Z_k, \tilde{Z}- \tilde{Z}^j \rangle|\leq 1$; for
the denominator, $2c \langle Z_k, \tilde{Z} \rangle \geq - cd\geq
-\frac{\delta}{2}$. These inequalities are satisfied by setting
$c=\frac{1}{10}\sqrt{\frac{2}{5K}}$ and assuming $K\geq
\frac{2}{5}r^2$ (so $cd \leq \frac{1}{5}$).
Then $cd\leq \frac{1}{5}$ and $\delta =
\frac{1}{2}-\frac{cd}{2}\geq\frac{2}{5}$, so $c\leq
\frac{1}{10}\sqrt{\frac{\delta}{K}}$ and $cd \leq \frac{\delta}{2}$.

Plugging this back in to the inequality above yields
\begin{align*}
2\sum_{\Tilde{Z}\in \{-1,1\}^{r}} \EE_{\ExoState\sim
\pvecxi(\Tilde{Z})}[\Regret(K)] & \geq 2c
\sum_{\tilde{Z}}\frac{r}{1}\left( K - \frac{1}{\sqrt{2}} K
\sqrt{\frac{160 K c^2}{\delta}}\right)\\
& = 2cr\sum_{\tilde{Z}}\left( K - \frac{1}{\sqrt{2}} K \sqrt{\frac{160
K c^2}{\delta}}\right) \\
& \geq \frac{1}{10}\sqrt{\frac{2}{5K}} r \sum_{\tilde{Z}}\left(K -
\frac{1}{\sqrt{2}} K \sqrt{\frac{160 K }{\delta}}
\frac{1}{10}\sqrt{\frac{\delta}{K}} \right) \\
& = \sum_{\tilde{Z}}\frac{1}{10}\sqrt{\frac{2K}{5}} r\left(1 -
\frac{\sqrt{80}}{10}\right)\\
& \geq \sum_{\tilde{Z}} \frac{1}{200} r \sqrt{K},
\end{align*}
where the first inequality is due to $c\leq
\frac{1}{10}\sqrt{\frac{\delta}{K}}$.  Therefore, selecting
$\tilde{Z}^*$ maximizing the left hand side gives
\begin{align*}
\EE_{\ExoState\sim \pvecxi(\Tilde{Z}^*)}[\Regret(K)] \geq
\frac{1}{400}r \sqrt{K}.
\end{align*}
Since this holds for any sequence of actions $a(Z_1),\dots, a(Z_K)$,
it follows for an arbitrary algorithm $\algo$.
\end{rproof}


\paragraph{Step 2: an Exo-MDP instance with lower bound $\Omega(Hr\sqrt{K})$.}\label{sec:stationary_lower_bound_H}
We can now use the lower bound for the Exo-Bandit problem above to
show that the hard instance $\widetilde{\MDP}$ in
\cref{sec:unobs_lower_bound} achieves a lower bound of $\Omega(H
r\sqrt{K})$.

Specifically, at stage $h=1$, the dynamics of $\widehat{\MDP}$ is the
same as the Exo-Bandit problem in
\cref{sec:linear_bandit_lower_bound}, with the same action space
$\ActionSpace$ parameterized by $Z\in \{-1,1\}^{r}$ and exogenous
state distribution $\pvecxi(\Tilde{Z})$ parameterized by $\Tilde{Z}\in
\{-1,1\}^{r}$. For stages $h=2,3,\dots, H$, the transition and
reward functions under the Exo-MDP framework are designed such that
the reward incurred at the first stage is repeated $H$ times.

\paragraph{Proof of regret lower bound in \cref{thm:lower_bound}.}
At stage $h=1$, the hard instance $\MDP$ acts exactly the same as an
Exo-Bandit as in \cref{sec:linear_bandit_lower_bound}, incurring
reward $r_1$. In stages $h=2,3,\dots,H$, the specific form of
$\SysFun$ and $\RewFun$ forces the reward from the first stage to
repeat $H$ times regardless of the actions. In other words, for each
episode on the Exo-MDP $\MDP$, the reward over the whole episode, $H
r_1$, is entirely determined by the action at the first stage. Because
both $\SysFun$ and $\RewFun$ are independent of $\exostate_h, a_h$ for
$h=2,3,\dots,H$, the only function of the last $H-1$ stages is to
repeat the reward without revealing any additional information on
$\PXi$.

By~\cref{lem:horizon_1_lower}, for any bandit algorithm, there exists
parameter $\tilde{Z}^*$ such that the expected regret is lower bounded
by $\frac{1}{400}r\sqrt{K}$. Therefore, for any policy $\algo$ on the
hard Exo-MDP $\MDP$, there exists parameter $\tilde{Z'}^*$ such that
the expected regret is lower bounded by $\frac{1}{400}H r\sqrt{K}$.

\paragraph{Verification of the effective dimension.} Lastly we show that the effective dimension of the Exo-MDP $\MDP$ is $r+1$.  Relabeling $r$ to be $r-1$ and updating the universal constant $\gamma$ then finalizes the proof.

Recall the transition and reward information matrices from
\Cref{sec:effective_dimension}:
\[
\Fmattrans_{(s',s,a),\cdot} = \phi_p(s' \mid s,a),\qquad
\Fmatrew_{(s,a),\cdot} = \phi_r(s,a).
\]
We handle the transitions and the rewards separately, showing that each of them have rank equal to $r+1$.

\emph{Transition matrix.}  We start off by considering the first-stage
transition from $s=(s_1,1)$ to $s'=(2,1)$, where we recall that the
first component corresponds to the stage $h = 2$, and the second
component contains the observed reward.  The reward in our constructed
Exo-MDP is $R = [a_1]_{x_1}$.  For an action $a(Z)$, each pair
$(2i-1,2i)$ of coordinates of the vector $\phi_p(s'\mid s,a(Z))$ is
$(1,0)$ if $Z_i=+1$ and $(0,1)$ if $Z_i=-1$. For example, if $Z =
\{-1, 1, \ldots, 1\} \in \{-1,1\}^r$ then $\phi_p(s' \mid s, a(Z)) =
    [0,1,1,0,\ldots, 1,0]$.

Define vectors $u_0, u_1, \ldots, u_r \in \RR^{d}$ that capture the
structure of these first-stage transition features.  The vector $u_0$
corresponds to the ``all positive'' configuration of $Z = \{1, \ldots,
1\}$ and consists of repeated $(1,0)$ pairs:
\[
u_0 = [1,0,\ldots,1,0].
\]
For each coordinate $i \in \{1, \ldots, r\}$, let $u_i$ be the vector
whose $(2i-1, 2i)$ pair is $(-1,1)$ and whose other pairs are zero,
i.e.:
\[
[u_i]_{(2i-1, 2i)} = (-1, 1) \qquad [u_i]_{(2j-1, 2j)} = (0,0) \,\, \forall j \neq i.
\]
These vectors encode how flipping the sign of a single coordinate of
the $Z$ changes the corresponding component of the transition feature
$\phi_p(s' \mid s_1, a(Z))$.

With this notation we can encode the structure of the feature vector
explicitly. When $Z_i = 1$, the $(2i-1, 2i)$ entry of $\phi_p(s' \mid
s,a(Z))$ is $(1,0)$.  When $Z_i = -1$ it becomes $(0,1) = (1,0) + (-1,
1)$.  Thus, each coordinate $Z_i$ toggles whether the transition
feature uses $(0,1)$ or the flipped pair $(1,0)$.  This yields:
\[
\phi_p(s' \mid s,a(Z)) = u_0 + \sum_{i:Z_i = -1} u_i.
\]

Similarly, transitioning from $s = (s_1, 1)$ to $s' = (2, -1)$ can be written as:
\[
\phi_p(s' \mid s, a(Z)) = u_0 + \sum_{i: Z_i = 1} u_i
\]
by the same argument.
Hence, all of the first-stage transition features $\phi_p(s' \mid
s_1,a(Z))$ lie in $\mathrm{span}\{u_0, \ldots, u_r\}$.

For the later stages $h\ge2$, transitions do not depend on the
exogenous state or the action: from $(h,b)$ the next state is
deterministically $(h+1,b)$ for all $\exostate$. Thus the
corresponding feature vectors are either $\mathbf{1}_d$ or $0$. Note
that $\mathbf{1}_d = 2u_0 + \sum_{i=1}^r u_i$, so these rows lie in
the same span and do not increase the dimension.  Therefore
\[
\rank(\Fmattrans) = r+1.
\]

\emph{Reward matrix.}
{
We now turn to the reward information matrix. For the initial state $s_1$ and action $a(Z)$, recall that the reward is given by $\RewFun(s_1,a(Z),\exostate)=[a(Z)]_{\exostate}$. Since $a(Z)=(Z_1,-Z_1,\ldots,Z_r,-Z_r)$, the corresponding reward feature vector is
\[
\phi_r(s_1,a(Z))=\sum_{i=1}^r Z_i v_i,
\]
where $v_i$ is $(1,-1)$ on the $(2i-1,2i)$ pair and zero elsewhere. Hence all first-stage reward feature vectors lie in $\mathrm{span}\{v_1,\ldots,v_r\}$. For later stages $h\ge2$, the reward no longer depends on the exogenous state or action. From any state $(h,b)$ with $b\in\{-1,1\}$, the reward is deterministically equal to $b$, so the reward feature vector is $b\mathbf 1_d$. Therefore all reward rows lie in $\mathrm{span}\{v_1,\ldots,v_r,\mathbf 1_d\}$.

The transition-rank calculation above already shows that $\mathrm{rowspan}(\Fmattrans)=\mathrm{span}\{v_1,\ldots,v_r,\mathbf 1_d\}$ and $\rank(\Fmattrans)=r+1$. Thus adding the reward block does not increase the rank, and the joint effective dimension is
\[
r_{\mathrm{eff}}
=
\rank\!\begin{pmatrix}
\Fmattrans\\
\Fmatrew
\end{pmatrix}
=
r+1.
\]
Relabeling $r$ to be $r-1$ and updating the universal constant $\gamma$ finalizes the proof.
}

\begin{remark}\label{rmk:dim_padding}
In the construction above we fixed the size of the exogenous state space to be $d = 2r$ for notational convenience.  This is without loss of
generality.  For any larger ambient dimension $d' \ge 2r$, we can
embed the same hard instance into an Exo-MDP with exogenous state
space $\ExoStateSpace' = [d']$ as follows.

Let $d_{\mathrm{eff}} := 2r$ and view the original construction as
living on the first $d_{\mathrm{eff}}$ symbols of the exogenous state space $\{1,\dots,
d_{\mathrm{eff}}\} \subset [d']$.  For each parameter
$\Tilde{Z}\in\{-1,1\}^r$, define the extended exogenous distribution
$p_\xi'(\Tilde{Z}) \in \Delta([d'])$ by
\[
p_\xi'(\Tilde{Z})(j)
=
\begin{cases}
p_\xi(\Tilde{Z})(j), & j \in \{1,\dots,d_{\mathrm{eff}}\},\\[2pt]
0, & j \in \{d_{\mathrm{eff}}+1,\dots,d'\},
\end{cases}
\]
where $p_\xi(\Tilde{Z})$ is the $d_{\mathrm{eff}}$-dimensional vector
from the core construction (which already sums to $1$).  For the
extra symbols $j>d_{\mathrm{eff}}$, define the transition and reward
functions $\SysFun,\RewFun$ arbitrarily, as long as they do not depend
on $\Tilde{Z}$ or the chosen action.

By construction, the trajectory distribution and all rewards when the
exogenous state lies in $\{1,\dots,d_{\mathrm{eff}}\}$ are exactly the
same as in the original $d_{\mathrm{eff}}$-dimensional instance, and
the symbols $j>d_{\mathrm{eff}}$ occur with probability zero.  Hence
the regret of any algorithm on the padded instance coincides with its
regret on the original instance, and the $\Omega(Hr\sqrt{K})$ lower
bound carries over verbatim.  Moreover, the information matrix for the
padded instance is obtained from the original information matrix by
adding zero columns, so its rank remains $r$.  Thus the lower bound
in terms of $\rank(F)=r$ continues to hold for any ambient dimension
$d' \ge 2r$.
\end{remark}

\subsubsection{Proof of \cref{thm:lower_bound_non_homogeneous}}\label{sec:unstationary_lower_bound_H}

To start we give the hard instance as a function of $r$ for time-inhomogeneous Exo-MDPs $\MDP_{ns}$.

\paragraph{Construction of the hard instance.}
We follow the same basic construction as the hard Exo-MDP instance
in~\cref{sec:stationary_lower_bound_H}, but now allow the latent
parameter $Z$ for the exogenous state distribution to differ across
each stage in $h=1,2,\dots,\frac{H}{2}$. For convenience, we again set $d = 2r$ and assume that $r > H$.

Formally, define the hard \emph{time-inhomogeneous} Exo-MDP
$\MDP_{ns}[\PXi^h(\Tilde{Z}), \SysFun,\RewFun] = (\StateSpace
\times \ExoStateSpace, \ActionSpace, \Horizon+1, S_1,
\transition_h, \Reward_h).$\footnote{Note that we use $H+1$ as the
horizon, and use $H$ instead in the lower bound for notational
convenience.}
The state space is given by $\Ss = s_0\cup \{(h, \ind_\exostate,r)\mid
h = 0,1,\dots,H, \ind_\exostate\in\{1,2,\dots,\frac{H}{2}\}, r\in
\{-1,1,\emptyset\}\}$. That is, the state space $\Ss$ consists of the
set of tuples consisting of $h \in [H]$, an index $\ind_\exostate$ as
well as a binary code for reward $r\in \{-1,1\}$; it also includes a
dummy starting state $s_0$.

The exogenous state space is given by $\ExoStateSpace = [d] =
\{1,2,\dots,d\}$.
The action set is $\A=\{([Z]_1,-[Z]_1,[Z]_2,-[Z]_2,\dots, [Z]_{r},
-[Z]_{r})\mid [Z]_i \in \{-1,1\}\}$.

For stages $h =1,\dots, \frac{H}{2}$, the exogenous distribution is
\[
\pvecxi^h = (\PXih(1),\dots,
\PXih(d)) = \Bigl(\tfrac{1}{d}+c [\Tilde{Z}]_1^h,\tfrac{1}{d}-c
[\Tilde{Z}]_1^h,\dots, \tfrac{1}{d}+c
[\Tilde{Z}]_{r}^h,\tfrac{1}{d}-c
[\Tilde{Z}]_{r}^h\Bigr),\quad \Tilde{Z}^h \in \{-1,1\}^{r},
\]
where constant $c = \frac{1}{10}\sqrt{\frac{2}{5K}}$.

The distributions $\PXih$ for $h = \frac{H}{2}+1,\dots,H$ can be any
arbitrary distribution over $\ExoStateSpace$, since $\exostate_h$ has
no effect on transition or reward during those stages. Note that the
dynamics of the Exo-MDP is completely determined by vectors
$\Tilde{Z}^1,\dots,\Tilde{Z}^{H/2}$.

For stage $h = 0$, $\PXi^0$ is uniform over $\{1, 2, \dots,
\frac{H}{2}\}$. Since $d > \frac{H}{2}$ this makes a valid
distribution. Specifically,
\begin{align*}
\PXi^0(\exostate) = \begin{cases} \frac{2}{H} & \text{if } \exostate =
1,2,\dots, \frac{H}{2} \\ 0 & \text{if } \exostate = \frac{H}{2},
\dots, d.\\
\end{cases}.
\end{align*}

The known state transition function is given by
\begin{align*}
s_{h + 1} = \SysFun(s_h, a_h, \exostate_h) =
\begin{cases}
(0,\ind_\exostate, \emptyset) & \quad \text{if } s_h=s_0,
\ind_\exostate\sim \PXi^0 \\ (h+1,\ind_\exostate, \emptyset) &
\quad \text{if } s_h = (h, \ind_\exostate, \emptyset), h+1\neq
\ind_\exostate \\
(h + 1,\ind_\exostate, r=[a_h]_{\exostate_h}) & \quad \text{if } s_h =
(h, \ind_\exostate, \emptyset), h+1 = \ind_\exostate, a_h,
\exostate_h \sim \PXih \\
(h + 1, \ind_\exostate, r) & \quad \text{if } s_h = (h,
\ind_\exostate, r), r\in\{-1,1\}. \\
\end{cases}
\end{align*}

The known reward function is given by
\begin{align*}
r_h = \RewFun(s_h, a_h, \exostate_h) = \begin{cases} 0&\quad\text{if }
h \leq \frac{H}{2}\\ r &\quad \text{if } h> \frac{H}{2}, s_h = (h,
\ind_\exostate, r).
\end{cases}
\end{align*}

\paragraph{Proof of regret lower bound in \cref{thm:lower_bound_non_homogeneous}.}  In each episode,
the Exo-MDP proceeds in two stages, each lasting $\frac{H}{2}$
stages. The first $\frac{H}{2}$ stages entail no reward, but decides
the reward $r$ that the agent will receive in the later $\frac{H}{2}$
stages. The second $\frac{H}{2}$ stages repeats this reward
$\frac{H}{2}$ times. Specifically, at stage $h=0$, the Exo-MDP draws a
sample of $\ind_\exostate$ from a uniform distribution in
$[\frac{H}{2}]$, which indexes the reward function that the agent
receives. For stages $h=1,2,\dots,\frac{H}{2}$, the exogenous state
follows distribution $\exostate_h\sim \PXi^h$, and when the stage
reaches the corresponding index $\ind_\exostate$, the third component
of the state $r$ is updated as $[a_h]_{\exostate_h}$, that is, the
action vector $a_h$ indexed at $\exostate_h \in [d]$. Given
$\ind_\exostate\leq \frac{H}{2}$, after $\frac{H}{2}$ stages, $r$ will
take on some value in $\{-1,1\}$. This will be the reward the agent
receives for stages $h = \frac{H}{2}+1,\dots, H$. Note that the
agent's action has no effect other than at the stage $h =
\ind_\exostate$, when the reward $r$ takes the value corresponding to
the $[a_h]_{\exostate_h}$.

By construction, the only components throughout an episode of length
$H$ that determine the final reward for the Exo-MDP is
$\ind_\exostate$, which is sampled at stage $h=0$, as well as the
action $a\in \A$ taken at stage $h= \ind_\exostate$. Therefore the
optimal policy $\pi^*$ maps $\ind_\exostate$ to the action $a_h$ taken
at $h=\ind_\exostate$. Moreover, the reward $r$ is maximized when $a_h
= \Tilde{Z}^{\ind_\exostate}$ (which maximizes the probability that $r
= 1$). Therefore, the optimal policy is to take action
$\Tilde{Z}^{\ind_\exostate}$ at $h= \ind_\exostate$:
\begin{align*}
a^*_{\ind_\exostate} = \pi_{\ind_\exostate}^*(\ind_\exostate) =
\Tilde{Z}^{\ind_\exostate}.
\end{align*}
Given at $h=0$, $\ind_\exostate$ is uniformly drawn from
$\{1,2,\dots,\frac{H}{2}\}$, in $K$ episodes, any specific bandit
$\bandit_{\ind_\exostate}$ is played $\frac{2K}{H}$ times in
expectation.
Therefore, by \cref{lem:horizon_1_lower}, the regret incurred on any specific bandit $\bandit_{\ind_\exostate}$ is given by
\begin{align*}
\EE_{\bandit_{\ind_\exostate}}\Bigl[\Regret\Bigl(\frac{2K}{H}\Bigr)\Bigr] \geq
\frac{H}{2}\cdot \frac{d\sqrt{\frac{2K}{H}}}{400}
= \frac{d\sqrt{2HK}}{800}
= \frac{r\sqrt{2HK}}{400},
\end{align*}
where we used $r=d/2$ in the last equality.
The total regret over $K$ episodes with starting state ranging across
$\ind_\exostate = 1,2,\dots,\frac{H}{2}$ is therefore
\begin{align*}
\sum_{\ind_\exostate = 1}^{\frac{H}{2}}
\EE_{\bandit_{\ind_\exostate}}\Bigl[\Regret\Bigl(\frac{2K}{H}\Bigr)\Bigr]
&= \frac{H}{2}
\cdot \frac{d \sqrt{2HK}}{800}
= \frac{\sqrt{2}}{1600} H^{3/2} d \sqrt{K} \\
&= \frac{H}{2}\cdot \frac{r\sqrt{2HK}}{400}
= \frac{\sqrt{2}}{800} H^{3/2} r \sqrt{K}.
\end{align*}
In particular, this yields a lower bound of order $\Omega(H^{3/2} r\sqrt{K})$.

\paragraph{Verification of the effective dimension.}
We now compute the effective dimension of our time-inhomogeneous hard instance $\MDP_{ns}$ and show that it equals $r + 1 + \lceil H/4 \rceil$.  However, since $r > H$ then $r + 1 + \lceil H/4 \rceil < 2r$.  Relabeling $r$ accordingly gives an MDP with effective dimension $r$ and the lower bound still holds after redefining the universal constant $\gamma$.

We analyze the reward and transition matrices separately.

\emph{Reward matrix.} By construction, the reward is equal to zero for the first half of the episode (i.e. $\RewFun(s_h, a_h, \exostate_h) = 0$ for all $h \leq \frac{H}{2}$.  Therefore, every row of $\Fmatrew$ corresponding to a state with time component $h \leq \frac{H}{2}$ is the zero vector.

For stages $h > \frac{H}{2}$ the state is of the form $(h, \ind_{\exostate}, b)$ with $b \in \{-1,1\}$.  The reward does not depend on the exogenous state or action and is $\RewFun((h,\ind_\exostate,b),a,\exostate) = b$.  Thus the corresponding reward feature vector is:
\[
\phi_r((h,\ind_\exostate, b), a) = b \mathbf{1}_d.
\]
Consequently, we obtain that $\rank(\Fmatrew) = 1$, since the nonzero rows are multiples of $\mathbf{1}_d$.

\emph{Transition matrix.} We next consider the transition information matrix $\Fmattrans$.  Before analyzing the feature vectors start by defining vectors $u_0, u_1, \ldots, u_r \in \RR^{d}$ as follows.  The vector $u_0$ corresponds to the ``all positive'' configuration of $Z = \{1, \ldots, 1\}$ and consists of repeated $(1,0)$ pairs:
\[
u_0 = [1,0,\ldots,1,0].
\]
For each coordinate $i \in \{1, \ldots, r\}$, let $u_i$ be the vector whose $(2i-1, 2i)$ pair is $(-1,1)$ and whose other pairs are zero, i.e.:
\[
[u_i]_{(2i-1, 2i)} = (-1, 1) \qquad [u_i]_{(2j-1, 2j)} = (0,0) \,\, \forall j \neq i.
\]

We start by considering the first stage.  Here we have a transition independent of the action via $\SysFun(s_0, a, i) = (0, \ind_i, \emptyset)$.  The corresponding transition feature vector is $\phi_p((0,i,\emptyset) \mid s_0, a) = e_i$, the standard basis vector.  Thus, the rows starting from $s_0$ span the subspace governed by $E = \mathrm{span}\{e_1, \ldots, e_{H/2}\}$.

Now we consider stages $1 \leq h \leq \frac{H}{2} - 1$.  Starting from a state $(h, \ind_\exostate, \emptyset)$, the next state depends on whether the stage index matches the exogenous index $\exostate$. If $h+1 \neq \exostate$ then the transition is deterministic and does not depend on the newly sampled exogenous state.  In this case the corresponding $\phi_p(s' \mid s, a(Z))$ is either $\mathbf{1}_d$ or $\mathbf{0}_d$.

Otherwise, the next reward bit is updated to $b=[a_h]_{\exostate_h}$ and the next state is either $s_+=(h+1,\ind_\exostate,+1)$ or $s_-=(h+1,\ind_\exostate,-1)$. A direct calculation identical to the time-homogeneous case (see \cref{app:lower_bound_proof}) gives
\[
\phi_p(s_+\mid s,a(Z))
= u_0 + \sum_{i : Z_i = -1} u_i,
\qquad
\phi_p(s_-\mid s,a(Z))
= u_0 + \sum_{i : Z_i = 1} u_i.
\]
Thus all such rows lie in
$B \defn \mathrm{span}\{u_0, u_1, \ldots, u_r\}.$
As in the time-homogeneous instance,
$\dim(B)=r+1$.

The remaining transitions for $h \geq H/2$ do not depend on the exogenous state, so their feature vectors are either $\mathbf{1}_d$ or $\mathbf{0}_d$, and hence lie in $B$.  The row space of
$\Fmattrans$ is therefore the sum of the two subspaces given by $B$ and $E$.  Hence $r+1 \leq \rank(\Fmattrans)
\leq (r+1) + \frac{H}{2}$.

{The nonzero rows of $\Fmatrew$ are multiples of $\mathbf 1_d$, and $\mathbf 1_d\in B\subseteq\mathrm{rowspan}(\Fmattrans)$. Thus adding the reward block does not increase the rank. Consequently, the joint effective dimension satisfies
\[
r_{\mathrm{eff}}
=
\rank\!\begin{pmatrix}
\Fmattrans\\
\Fmatrew
\end{pmatrix}
=
\rank(\Fmattrans)
\le (r+1)+\frac{H}{2}.
\]
In particular, under the assumption that $r>H$, we have $r_{\mathrm{eff}}=O(r)$, so the $\Omega(H^{3/2}r\sqrt K)$ lower bound can equivalently be stated in terms of $r_{\mathrm{eff}}$ up to universal constants.}

Lastly, following a similar argument as in \cref{rmk:dim_padding}, we can pad the exogenous state size $d$ to an arbitrarily large number without affecting the joint effective dimension, so the statistical limit depends on the effective dimension rather than on $d$.

\subsection{Upper Bounds}
\label{app:proofs_unobserved_upper}

\subsubsection{Proof of \cref{thm:linear_mixture_mdp_low_rank_non_homogeneous}}

{Let $\Fmattrans$ and $\Fmatrew$ denote the transition and reward information matrices defined in~\cref{sec:effective_dimension}, and consider the joint information matrix $\Fmat$ with $r=\rank(\Fmat)$. Consider any rank-$r$ factorization $\Fmat=AB^\top$. For each transition row $(s',s,a)$ and reward row $(s,a)$, let $A^{(p)}_{s',s,a}\in\RR^r$ and $A^{(r)}_{s,a}\in\RR^r$ denote the corresponding rows of $A$, viewed as column vectors. In the time-inhomogeneous setting, define the stage-dependent common parameter $\theta_h\defn B^\top\pvecxih\in\RR^r$. By the definition of the information matrices and the factorization $\Fmat=AB^\top$, we have
\[
\transition_h(s'\mid s,a)=e_{s'\mid s,a}^\top\Fmattrans\pvecxih=\bigl(A^{(p)}_{s',s,a}\bigr)^\top\theta_h,\qquad
r_h(s,a)=e_{s,a}^\top\Fmatrew\pvecxih=\bigl(A^{(r)}_{s,a}\bigr)^\top\theta_h.
\]
Thus the time-inhomogeneous Exo-MDP is an $r$-dimensional time-inhomogeneous linear mixture MDP with known reduced features $\widetilde\phi_p(s'\mid s,a)=A^{(p)}_{s',s,a}$ and $\widetilde\phi_r(s,a)=A^{(r)}_{s,a}$, and stage-dependent unknown parameters $\{\theta_h\}_{h=1}^H$. Running the \UV algorithm on this rank-reduced linear mixture representation gives the stated performance by applying Theorem 5.3 of~\citet{zhou2021nearly}.}

\section{Proofs for Observed Exogenous States}
\label{app:proofs_observed}

\subsection{Lower bounds}
\label{app:proofs_observed_lower}

In this subsection we prove a lower bound for the full observation setting. Fix an integer parameter $r\ge 1$, we construct a
family of fully observed time-homogeneous Exo-MDP instances whose effective
dimension is $r+2$, and show that their expected regret over $K$ episodes of
horizon $H$ is lower bounded by $\Omega \bigl(H\sqrt{rK})$ for $K\geq r$ and $H>3$. Since the
effective dimension of our construction is $r_{\mathrm{eff}}=r+2$, this implies
a lower bound of order $\Omega(H\sqrt{r_{\mathrm{eff}}K})$ up to universal
constant factors (after relabeling $r$ if desired). We first analyze the one-step case $H=1$ (an ``Exo-Bandit'') and then lift the
construction to an Exo-MDP with general horizon $H$.

\paragraph{Step 1: the ``Exo-Bandit'' problem with lower bound $\Omega(\sqrt{rK})$.}
\label{subsec:H1}
Similar to \cref{app:lower_bound_proof} we start off with constructing an Exo-Bandit (an Exo-MDP with $H = 1$).

\paragraph{Setup of the ``Exo-Bandit'' problem.}
Assume a single endogenous state $s_1$. The exogenous state space is
$\ExoStateSpace = [r] \times \{\pm 1\}$, and we write $x = (j,b)$ with $j \in [r]$ and $b \in \{\pm1\}$ for a sample exogenous state $x \in \ExoStateSpace$.
Since we focus on step $H = 1$, the endogenous dynamics are trivial and do not depend on the exogenous state with $\SysFun(s_1,a,x) = s_1$ for all $a \in \{\pm1\}^r, x \in \ExoStateSpace$.

The action set is the hypercube $\ActionSpace = \{\pm1\}^r$.
For $a \in \ActionSpace$ we write $a = (a(1),\dots,a(r))$ with $a(j) \in \{\pm1\}$.
The (known) reward function is
\begin{equation}
\label{eq:rw-H1}
\RewFun(s_1,a,(j,b))
\;=\; \frac{1 + a(j) b}{2}
\;\in [0,1].
\end{equation}
Thus the reward is $1$ if $a(j)$ matches $b$ and $0$ otherwise.

The unknown environment parameter is a sign vector $u = (u_1,\dots,u_r) \in \{\pm1\}^r$.
For a fixed $\epsilon \in (0,1/4]$, the exogenous state $X = (J,B) \in \ExoStateSpace$ has distribution
\begin{equation}
\label{eq:pv}
\PXi^u(J=j,B=b)
= \frac{1}{2r}\bigl(1+\epsilon u_j b\bigr),
\qquad j\in[r],\ b\in\{\pm1\}.
\end{equation}
Equivalently, $J$ is uniform on $[r]$, and conditional on $J=j$, the bit $B$ has mean
$\mathbb{E}[B\mid J=j] = \epsilon u_j$.

For a fixed instance $u$ and action $a \in \ActionSpace$, the expected one-step reward is
\[
r(a)
\defn \mathbb{E}_u\bigl[\RewFun(s_1,a,X)\bigr],
\qquad X \sim \PXi^u,
\]
where we omit the dependence of the reward on the fixed state $s_1$ for brevity.
A direct computation gives
\begin{align}
\label{eq:value-H1-simple}
r_u(a)
&= \sum_{j=1}^r \sum_{b=\pm1}
\frac{1}{2r}\bigl(1+\epsilon u_j b\bigr)
\cdot \frac{1 + a(j)b}{2} \\
&= \frac{1}{2}
\;+\; \frac{\epsilon}{2r}\sum_{j=1}^r a(j)u_j
\;=\; \frac{1}{2} + \frac{\epsilon}{2r}\,\langle a, u\rangle. \nonumber
\end{align}
Hence the optimal action is $a^*(u) = u$ with optimal value
\[
r_u(a^*)
= \max_{a\in\ActionSpace} r_u(a) = r_u(u)
= \frac{1}{2} + \frac{\epsilon}{2}.
\]
Moreover, for any $a \in \ActionSpace$ the suboptimality gap in expectation is
\begin{equation}
\label{eq:gap-H1-simple}
r_u(a^*)- r_u(a)
= \frac{\epsilon}{2}\left(1 - \frac{1}{r}\langle a,u\rangle\right).
\end{equation}

\paragraph{Regret Analysis.}
Let $\mathcal{M}_u$ denote the Exo-Bandit instance associated with parameter $u \in \{\pm1\}^r$.
Fix a learning algorithm $\mathcal{B}$ that, in episode $k$, selects an action $a_k \in \ActionSpace$ as a measurable function of the past exogenous observations and its internal randomness.

In the one-step case $H=1$, each episode consists of a single interaction, so the regret of $\mathcal{B}$ on instance $\mathcal{M}_u$ after $K$ episodes is
\[
\Regret_u(K, \mathcal B)
\defn \sum_{k=1}^K r_u(a^*) - r_u(a_k),
\]
where $r_u(a)$ and $r_u(a^*)$ are given by
\cref{eq:value-H1-simple}–\cref{eq:gap-H1-simple}, and the expectation is taken with respect to both the exogenous process and the randomness of $\mathcal{B}$.

In order to derive a lower bound on the expected regret, we place a prior on the hard instances.  Let $U$ be uniform on $\{\pm1\}^r$ and let $\mathcal{M}_U$ denote the random instance determined by $U$.
For an algorithm $\mathcal{B}$, the Bayes regret under this prior is
\[
\BayesReg_K(\mathcal{B})
\defn \mathbb{E}\bigl[\Regret_U(K, \mathcal B)\bigr],
\]
where the expectation is over both $U$ and the exogenous draws.
The minimax regret is
\[
\Reg_K^\star
\defn \inf_{\mathcal{B}}
\sup_{u\in\{\pm1\}^r}
\mathbb{E}\bigl[\Regret_u(K, \mathcal B)\bigr].
\]
Since for any prior on $U$ and any algorithm $\mathcal{B}$,
\[
\sup_{u\in\{\pm1\}^r} \mathbb{E}\bigl[\Regret_u(K, \mathcal B)\bigr]
\geq \mathbb{E}\bigl[\Regret_U(K, \mathcal B)\bigr]
= \BayesReg_K(\mathcal{B}),
\]
a Bayes lower bound that holds for all algorithms immediately implies the same lower bound for $\Reg_K^\star$.

In order to derive a lower-bound on regret for this set of instances we will use the following information-theoretic lemma.

\begin{lemma}
\label{lem:binary-mi}
Let $U\in\{\pm1\}$ be uniform and let $Y$ be any random variable.
Define $m(Y):=\mathbb{E}[U\mid Y]\in[-1,1]$.
Then
\[
\mathbb{E}\bigl[m(Y)^2\bigr] \;\le\; 4\,I(U;Y),
\]
where $I(\cdot;\cdot)$ denotes mutual information.
\end{lemma}

\begin{proof}
For each $y$ we have
\[
\Pr(U=+1\mid Y=y) = \frac{1+m(y)}{2},
\qquad
\Pr(U=-1\mid Y=y) = \frac{1-m(y)}{2}.
\]
Hence
\[
I(U;Y)
= \mathbb{E}\left[
\frac{1+m(Y)}{2}\log(1+m(Y))
+\frac{1-m(Y)}{2}\log(1-m(Y))
\right].
\]
Define
\[
F(z)
:= \frac{1+z}{2}\log(1+z)
+\frac{1-z}{2}\log(1-z)
-\frac{z^2}{4},
\qquad z\in[-1,1].
\]
A direct computation yields
$F''(z)=\tfrac{1}{1-z^2}-\tfrac12\ge0$ on $[-1,1]$, so $F$ is convex and minimized at $z=0$, where $F(0)=0$.
Thus $F(z)\ge0$ for all $z\in[-1,1]$.
Substituting $z=m(Y)$ and taking expectations gives
\[
I(U;Y)
\;\ge\;
\frac{1}{4}\,\mathbb{E}\bigl[m(Y)^2\bigr],
\]
which rearranges to the desired inequality.
\end{proof}

We now state and prove the lower bound for the one-step Exo-Bandit.

\begin{lemma}
\label{thm:H1-simple}
Consider the one-step family of instances $\{\mathcal{M}_u: u\in\{\pm1\}^r\}$ described above, and let $U$ be uniform on $\{\pm1\}^r$.
There exists a universal constant $c>0$ such that for any algorithm $\mathcal{B}$ and all $K\ge1$,
\[
\BayesReg_K(\mathcal{B})
\;\ge\; c\,\min\{\sqrt{rK},\,K\}.
\]
In particular, for all $K\ge r$,
\[
\BayesReg_K(\mathcal{B}) \;\ge\; c\sqrt{rK}.
\]
Consequently, for each $r$ and all $K\ge r$,
\[
\Reg_K^\star
\;\ge\; c\sqrt{rK}.
\]
\end{lemma}

\begin{proof}
Fix an arbitrary algorithm $\mathcal{B}$.
For episode $k$, using the gap expression~\cref{eq:gap-H1-simple} and the prior on $U$ we obtain
\begin{equation}
\label{eq:ep-regret-H1-simple}
\mathbb{E}\bigl[r_u(a^*) - r_u(a_k) \bigr]
= \frac{\epsilon}{2}
\left(1 - \frac{1}{r}\mathbb{E}\bigl[\langle a_k,U\rangle\bigr]\right),
\end{equation}
where the expectation is over $U$, the exogenous draws, and the internal randomness of $\mathcal{B}$.

Let $\mathcal{F}_{k-1} := \sigma(X_1,\dots,X_{k-1})$ denote the history of exogenous states before episode $k$, and define the posterior mean of the instance $U$ via
\(
\mu_{k-1}
\defn \mathbb{E}[U\mid \mathcal{F}_{k-1}]
\in[-1,1]^r.
\)
Conditioning on $\mathcal{F}_{k-1}$ and using that $a_k$ is $\mathcal{F}_{k-1}$-measurable,
\[
\mathbb{E}\bigl[\langle a_k,U\rangle \mid \mathcal{F}_{k-1}\bigr]
= \langle a_k,\mu_{k-1}\rangle.
\]
By Cauchy–Schwarz and $\|a_k\|_2=\sqrt{r}$,
\[
\langle a_k,\mu_{k-1}\rangle
\;\le\; \sqrt{r}\,\|\mu_{k-1}\|_2.
\]
Taking expectations and applying Jensen's inequality,
\begin{equation}
\label{eq:align-bound-H1-simple}
\mathbb{E}\bigl[\langle a_k,U\rangle\bigr]
\leq \sqrt{r}\,\sqrt{\EE[\norm{\mu_{k-1}}_2^2]}.
\end{equation}

We next bound $\EE[\norm{\mu_{k-1}}_2^2]$ using Lemma~\ref{lem:binary-mi}.
Let $U_j$ be the $j$-th coordinate of $U$, we write $\mu_{k-1} = (\mu_{k-1,1},\dots,\mu_{k-1,r})$ where
\[
\mu_{k-1,j} := \mathbb{E}[U_j\mid \mathcal{F}_{k-1}],
\qquad j\in[r].
\]
For each $j$, Lemma~\ref{lem:binary-mi} with $U=U_j$ and $Y=(X_1,\dots,X_{k-1})$ gives
\[
\mathbb{E}[\mu_{k-1,j}^2]
\;\le\; 4\,I(U_j;X_1,\dots,X_{k-1}),
\]
where $I(\cdot;\cdot)$ denotes mutual information.
Summing over $j$,
\begin{equation}
\label{eq:mu2-mi-sum-simple}
\EE[\norm{\mu_{k-1}}_2^2]
= \sum_{j=1}^r \mathbb{E}[\mu_{k-1,j}^2]
\;\le\; 4\sum_{j=1}^r I(U_j;X_1,\dots,X_{k-1}).
\end{equation}

It remains to bound $I(U_j;X_1,\dots,X_{k-1})$ for each coordinate $j$.
For a single draw $X=(J,B)$, note that $J$ is uniform on $[r]$ and independent of $U_j$, and conditional on $J=j$ the bit $B$ has distribution
\[
\Pr(B=b\mid J=j, U_j=u)
= \frac{1}{2}\bigl(1+\epsilon u b\bigr),
\qquad u\in\{\pm1\},\ b\in\{\pm1\}.
\]
Let $P^+$ (resp.\ $P^-$) denote the conditional law of $B$ given $J=j$ and $U_j=+1$ (resp.\ $U_j=-1$); then
\[
P^+(B=b) = \frac{1}{2}(1+\epsilon b),
\qquad
P^-(B=b) = \frac{1}{2}(1-\epsilon b).
\]
A direct computation shows
\begin{equation}
\label{eq:KL-B-simple}
\mathrm{KL}(P^+\|P^-)
= \epsilon\log\frac{1+\epsilon}{1-\epsilon}
\;\le\; 3\epsilon^2
\qquad\text{for }\epsilon\in(0,1/4],
\end{equation}
where we used $\log\frac{1+\epsilon}{1-\epsilon}\le 3\epsilon$ in this range.
Given $P^+(B=b)+P^-(B=b)=1$, we have $\mathrm{KL}(P^-\|P^+) = \mathrm{KL}(P^+\|P^-)$.

Since $J$ is independent of $U_j$, we have
\[
I(U_j;X_1,\dots,X_{k-1})
= I(U_j;J_1,\dots,J_{k-1},B_1,\dots,B_{k-1})
= I(U_j;B_1,\dots,B_{k-1}\mid J_1,\dots,J_{k-1}).
\]

Because the exogenous draws are i.i.d.\ conditional on $U$, we have
\[
I(U_j;X_1,\dots,X_{k-1})
\le \sum_{t=1}^{k-1} I(U_j;X_t)
= (k-1)\,I(U_j;X),
\]
where $X\sim\PXi^U$ is a single draw.  Since $J$ is independent of $U_j$,
\[
I(U_j;X)=I(U_j;B\mid J)
=\Pr(J=j)\,I(U_j;B\mid J=j)
=\frac{1}{r}I(U_j;B\mid J=j).
\]
Using \cref{eq:KL-B-simple} (so $I(U_j;B\mid J=j)\le 3\epsilon^2$), we obtain
\begin{equation}
\label{eq:MI-upper-simple}
I(U_j;X_1,\dots,X_{k-1})
\le \frac{3(k-1)\epsilon^2}{r}.
\end{equation}

Combining~\cref{eq:mu2-mi-sum-simple} and~\cref{eq:MI-upper-simple},
\[
\mathbb{E}\|\mu_{k-1}\|_2^2
\;\le\; 4 \sum_{j=1}^r \frac{3(k-1)\epsilon^2}{r}
= 12(k-1)\epsilon^2.
\]
Substituting into~\cref{eq:align-bound-H1-simple} we obtain

\[
\mathbb{E}\bigl[\langle a_k,U\rangle\bigr]
\leq \sqrt{r}\,\sqrt{12(k-1)\epsilon^2}.
\]

Using this equation we are ready to derive the regret analysis. Plugging the previous inequality into~\cref{eq:ep-regret-H1-simple} yields the per-episode Bayes regret bound
\begin{equation}
\label{eq:episode-lb-H1-simple}
\mathbb{E}\bigl[r_u(a^*) - r_u(a_k)\bigr]
\;\ge\; \frac{\epsilon}{2}\left(1 - \sqrt{\frac{12(k-1)\epsilon^2}{r}}\right).
\end{equation}
Summing~\cref{eq:episode-lb-H1-simple} over $k=1,\dots,K$ and writing $(x)^+ := \max\{x,0\}$,
\[
\BayesReg_K(\mathcal{B})
= \EE[\Regret_U(K, \mathcal B)]
\geq \frac{\epsilon}{2}\sum_{k=1}^K
\left(1-\sqrt{\frac{12(k-1)\epsilon^2}{r}}\right)^+.
\]
Let $\gamma := 12\epsilon^2/r$.
For all $k$ with $k-1\le 1/(4\gamma)$ we have $\sqrt{\gamma(k-1)}\le 1/2$, hence the summand is at least $1/2$.
Therefore
\[
\sum_{k=1}^K\left(1-\sqrt{\gamma(k-1)}\right)^+
\;\ge\; \frac{1}{2}\min\left\{K,\ \frac{1}{4\gamma}+1\right\}
\;\ge\; \frac{1}{2}\min\left\{K,\ \frac{1}{4\gamma}\right\}
\;\ge\; \frac{1}{8}\min\left\{K,\ \frac{1}{\gamma}\right\}.
\]
We conclude that
\begin{equation}
\label{eq:regret-pre-tune-H1-simple}
\BayesReg_K(\mathcal{B})
\;\ge\; \frac{\epsilon}{16}\,
\min\left\{K,\ \frac{r}{12\epsilon^2}\right\}.
\end{equation}

Now set
\[
\epsilon := \min\left\{\frac{1}{4},\ \sqrt{\frac{r}{12K}}\right\}.
\]
If $\sqrt{r/(12K)}\le 1/4$, then $\epsilon=\sqrt{r/(12K)}$ and $r/(12\epsilon^2)=K$, so~\cref{eq:regret-pre-tune-H1-simple} gives
\[
\BayesReg_K(\mathcal{B})
\;\ge\; \frac{\epsilon}{16}K
= \frac{1}{16}\sqrt{\frac{rK}{12}}
\;\ge\; c_1\sqrt{rK}
\]
for some universal constant $c_1>0$.
Otherwise $\epsilon=1/4$, and since $\sqrt{r/(12K)}>1/4$ we have $r/(12K) > 1/16$, i.e.\ $r > 3K/4$.
In this regime, $\sqrt{rK}\ge\sqrt{\tfrac{3}{4}}K$, so $\min\{\sqrt{rK},K\}=K$.
From \cref{eq:regret-pre-tune-H1-simple},
\[
\BayesReg_K(\mathcal{B})
\;\ge\; \frac{1}{64}
\min\left\{K,\ \frac{r}{12(1/4)^2}\right\}
= \frac{1}{64}K
\;\ge\; c_2\min\{\sqrt{rK},K\}
\]
for some universal constant $c_2>0$.
Taking $c := \min\{c_1,c_2\}$ completes the proof of the Bayes bound.
The minimax bound then follows from the Bayes–minimax relation above.
\end{proof}

\paragraph{Step 2: an Exo-MDP instance with lower bound $\Omega(H\sqrt{rK})$.}
\label{subsec:Hgeneral}

We now lift the one-step construction to general horizon $H$ so that the learner
still makes exactly one meaningful decision per episode, while (i) the per-episode
value gap scales as $\Theta(H\epsilon)$ and (ii) the total information per episode
about the unknown parameter remains $O(\epsilon^2)$ (i.e., does not grow with $H$).  We construct an MDP $\MDP_u$ indexed by a latent vector $u \in \{ \pm 1\}^r$ as follows.


\paragraph{Exogenous state space and law.}
Let the exogenous state space be
\[
\ExoStateSpace \;=\; \{0\}\ \cup\ ([r]\times\{\pm1\}),
\qquad d:=|\ExoStateSpace|=2r+1,
\]
and write $x=(j,b)$ for $j\in[r]$ and $b\in\{\pm1\}$, and $x=0$ for the dummy
symbol.

Fix $\epsilon\in(0,1/4]$ and an unknown parameter
$u=(u_1,\dots,u_r)\in\{\pm1\}^r$.  Define the (time-homogeneous) exogenous law
$\PXi^u$ by
\begin{equation}
\label{eq:pv-rare}
\PXi^u(X=0)=1-\frac{1}{2H},
\qquad
\PXi^u(X=(j,b)) =\frac{1}{4Hr}\bigl(1+\epsilon u_j b\bigr),
\quad j\in[r],\ b\in\{\pm1\}.
\end{equation}
Intuitively, with probability $1-\frac{1}{2H}$ the exogenous draw is the dummy
symbol $0$, which is independent of $u$ and hence uninformative.  With the
remaining probability $\frac{1}{2H}$ we observe an informative pair
$(J,B)\in[r]\times\{\pm1\}$.

\paragraph{States, actions, dynamics, and rewards.}
We introduce a single nonterminal state $s_1$ at the beginning of each episode,
together with two absorbing layers indexed by the initial action:
\[
\{\bar s_a : a\in\{\pm1\}^r\}\quad\text{(``alive'')},
\qquad
\{\bar d_a : a\in\{\pm1\}^r\}\quad\text{(``dead'')}.
\]
The action set at $s_1$ is $\ActionSpace=\{\pm1\}^r$, and at all other states
only a dummy action is available.  For $a\in\ActionSpace$ we write
$a=(a(1),\dots,a(r))$.

\emph{Dynamics.} For all $a\in\ActionSpace$ and all $x\in\ExoStateSpace$,
\begin{align*}
\SysFun(s_1,a,x) &= \bar s_a,\\
\SysFun(\bar s_a,\text{dummy},0) &= \bar s_a,\\
\SysFun(\bar s_a,\text{dummy},(j,b))
&=
\begin{cases}
\bar s_a,& b=a(j),\\
\bar d_a,& b\neq a(j),
\end{cases}\\
\SysFun(\bar d_a,\text{dummy},x) &= \bar d_a.
\end{align*}

In other words, after choosing $a$ at the initial state $s_1$, the process deterministically enters the corresponding ``alive'' state $\bar s_a$.  Thereafter, the agent has no further meaningful choices (only a dummy action).  If the exogenous draw is the dummy symbol $0$, the process stays alive. If the exogenous draw is an informative pair $(j,b)$, then the process stays alive if $b=a(j)$ and otherwise transitions to the corresponding ``dead'' state $\bar d_a$.  Once dead, the process remains dead for the rest of the episode.

\emph{Rewards.} At time $h=1$ the reward is identically zero,
$\RewFun(s_1,a,x)=0$.  For $h\ge 2$, define
\begin{equation}
\label{eq:rw-H-rare}
\RewFun(\bar s_a,\text{dummy},x)=1,\quad
\RewFun(\bar d_a,\text{dummy},x)=0,
\qquad \forall x\in\ExoStateSpace.
\end{equation}

\paragraph{Episode value.}
Fix an instance $u$ and an initial action $a\in\ActionSpace$. Let $\pi_a$ be the
policy that selects $a$ at $s_1$ and then plays the dummy action thereafter.
Define the per-step ``death'' probability while alive:
\begin{align}
\label{eq:lambda-def}
\lambda_u(a) &:= \Pr_u \bigl(\SysFun(\bar s_a,\text{dummy},X)=\bar d_a\bigr)
= \Pr_u \bigl(X=(j,b)\ \text{with}\ b\neq a(j)\bigr) \nonumber\\
&= \sum_{j=1}^r \PXi^u(X=(j,-a(j)))
= \sum_{j=1}^r \frac{1}{4Hr}\bigl(1-\epsilon u_j a(j)\bigr)
= \frac{1}{4H}-\frac{\epsilon}{4Hr}\langle a,u\rangle.
\end{align}
Thus $\lambda_u(a)\in[(1-\epsilon)/(4H),(1+\epsilon)/(4H)]$.

Since the process starts alive at time $h=2$, and each step independently kills
with probability $\lambda_u(a)$ while alive, we have
\[
\Pr_u(S_h=\bar s_a) = (1-\lambda_u(a))^{h-2},\qquad h=2,\dots,H.
\]
Using \cref{eq:rw-H-rare}, the horizon-$H$ value is
\begin{equation}
\label{eq:value-H-rare}
V^{\pi_a}_1(s_1,\MDP_u)
=\EE_u\!\left[\sum_{h=1}^H \RewFun(S_h,A_h,X_h)\,\Big|\,A_1=a\right]
=\sum_{h=2}^H \Pr_u(S_h=\bar s_a)
=\sum_{t=0}^{H-2}(1-\lambda_u(a))^{t}.
\end{equation}
Let $V(\lambda):=\sum_{t=0}^{H-2}(1-\lambda)^t$ so that
$V^{\pi_a}_1(s_1,\MDP_u)=V(\lambda_u(a))$.

\paragraph{Optimal action and value gap.}
From \cref{eq:lambda-def}, $\lambda_u(a)$ is minimized by maximizing
$\langle a,u\rangle$, hence the optimal action is $a^*(u)=u$.

\begin{lemma}
\label{lem:gap-H-rare}
Assume $H> 3$ and $\epsilon\in(0,1/4]$.  There exists a universal constant
$c_{\mathrm{gap}}>0$ such that for all $u\in\{\pm1\}^r$ and all $a\in\{\pm1\}^r$,
\begin{equation}
\label{eq:gap-H-rare}
V^{*}_1(s_1,\MDP_u) - V^{\pi_a}_1(s_1,\MDP_u)
\;\ge\;
c_{\mathrm{gap}}\,H\epsilon\left(1-\frac{1}{r}\langle a,u\rangle\right),
\end{equation}
where $V^{*}_1(s_1,\MDP_u):=V^{\pi_{u}}_1(s_1,\MDP_u)$.
\end{lemma}

\begin{proof}
Write $\lambda_*(u):=\lambda_u(u)=(1-\epsilon)/(4H)$ and $\lambda(a):=\lambda_u(a)$.
Since $V(\lambda)$ is decreasing in $\lambda$ and $\lambda(a)\ge \lambda_*(u)$,
the mean value theorem gives $\tilde\lambda\in[\lambda_*(u),\lambda(a)]$ such that
\[
V(\lambda_*(u)) - V(\lambda(a)) = -V'(\tilde\lambda)\cdot(\lambda(a)-\lambda_*(u)).
\]

We compute $-V'(\lambda)=\sum_{t=1}^{H-2} t(1-\lambda)^{t-1}.$ Because $\epsilon\le 1/4$, we have
$\tilde\lambda\le (1+\epsilon)/(4H)\le 5/(16H)$.
Let $M:=\lfloor (H-2)/2\rfloor$.  For all $1\le t\le M$ we have
$(1-\tilde\lambda)^{t-1}\ge (1-\tilde\lambda)^{M}$, hence
\[
-V'(\tilde\lambda)
\ge (1-\tilde\lambda)^{M}\sum_{t=1}^{M} t
= (1-\tilde\lambda)^{M}\cdot \frac{M(M+1)}{2}.
\]
Since $M\le H/2$ and $\tilde\lambda\le 5/(16H)$,
\[
(1-\tilde\lambda)^{M}\ \ge\ \left(1-\frac{5}{16H}\right)^{H/2}\ \ge\ c_0
\]
for a universal constant $c_0>0$.  Also $M(M+1)=\Theta(H^2)$ for $H\ge 3$.
Therefore there exists a universal constant $c_1>0$ such that
\[
-V'(\tilde\lambda)\ \ge\ c_1 H^2.
\]

Next, using \cref{eq:lambda-def},
\[
\lambda(a)-\lambda_*(u)
=\frac{\epsilon}{4H}\left(1-\frac{1}{r}\langle a,u\rangle\right).
\]
Combining the above yields \cref{eq:gap-H-rare} with $c_{\mathrm{gap}}:=c_1/4$.
\end{proof}

\paragraph{Regret and Bayes regret.}
Assume throughout that $H>3$. Let $\mathcal B$ be any learning algorithm that,
in episode $k$, selects an action $a_k\in\ActionSpace$ at $h=1$ as a measurable
function of the past observations and its internal randomness. The regret on
instance $u$ after $K$ episodes is
\[
\Regret_u(K,\mathcal B)
:= \sum_{k=1}^K \Bigl(V^{*}_1(s_1,\MDP_u) - V^{\pi_{a_k}}_1(s_1,\MDP_u)\Bigr).
\]
Let $U\sim\mathrm{Unif}(\{\pm1\}^r)$ and define the Bayes regret
$\BayesReg_K(\mathcal B):=\EE[\Regret_U(K,\mathcal B)]$.

Taking expectations in \cref{eq:gap-H-rare} (over $U$ and the algorithm's
randomness) gives, for each episode $k$,
\begin{equation}
\label{eq:ep-gap-step2}
\EE\!\left[V^{*}_1(s_1,\MDP_U) - V^{\pi_{a_k}}_1(s_1,\MDP_U)\right]
\;\ge\;
c_{\mathrm{gap}}\,H\epsilon\left(1-\frac{1}{r}\EE[\langle a_k,U\rangle]\right).
\end{equation}
Thus it remains to upper bound $\EE[\langle a_k,U\rangle]$.

By Lemma~\ref{lem:binary-mi} and the information-theoretic argument from
Step~1: it suffices to bound $I(U_j;\mathcal F_{k-1})$ for each coordinate $j$,
where $\mathcal F_{k-1}$ is the $\sigma$-field generated by all observations
prior to episode $k$.

Let $X_{h,k'}$ denote the exogenous state observed at step $h\in[H]$ of episode
$k'\in[K]$. Before episode $k$, the learner has observed $D_k := H(k-1)$  i.i.d.\ exogenous draws $\{X_{h,k'}:h\in[H],\,k'<k\}$ from $\PXi^U$. Since the
endogenous transitions and rewards are deterministic functions of $(S_h,A_h,X_h)$
and $X_h$ is fully observed, the entire history up to episode $k-1$ is a function
of these exogenous draws and the algorithm's internal randomness (which is
independent of $U$). Hence by the data processing inequality,
\[
I(U_j;\mathcal F_{k-1}) \;\le\; I\bigl(U_j;X_1,\dots,X_{D_k}\bigr),
\]
where we relabel the $D_k$ exogenous draws as $X_1,\dots,X_{D_k}$.

By the chain rule for mutual information and i.i.d.\ sampling,
\[
I\bigl(U_j;X_1,\dots,X_{D_k}\bigr)
\;\le\; \sum_{t=1}^{D_k} I(U_j;X_t)
= D_k\,I(U_j;X),
\]
where $X\sim\PXi^U$ is a single draw from \cref{eq:pv-rare}.

We now bound $I(U_j;X)$. Under \cref{eq:pv-rare}, with probability
$1-\frac{1}{2H}$ we have $X=0$, which is independent of $U$ and hence carries no
information. With the remaining probability $\frac{1}{2H}$ we have
$X=(J,B)\in[r]\times\{\pm1\}$, where conditional on $J=j$,
\[
\Pr(B=b\mid J=j,U_j=u)=\frac12(1+\epsilon ub),\qquad b\in\{\pm1\}.
\]
By the same KL computation as in \cref{eq:KL-B-simple}, we have
$I(U_j;B\mid J=j)\le 3\epsilon^2$. Since $\Pr(X\neq 0)=\frac{1}{2H}$ and
$\Pr(J=j\mid X\neq 0)=\frac{1}{r}$, we obtain
\[
I(U_j;X)\le
\Pr(X\neq 0,\ J=j)\cdot I(U_j;B\mid J=j)\le
\frac{1}{2H}\cdot \frac{1}{r}\cdot 3\epsilon^2
=\frac{3\epsilon^2}{2Hr}.
\]

Combining with $D_k=H(k-1)$ yields
\begin{equation}
\label{eq:MI-step2}
I(U_j;\mathcal F_{k-1})
\;\le\; \frac{3(k-1)\epsilon^2}{2r}.
\end{equation}

Applying Lemma~\ref{lem:binary-mi} coordinate-wise and summing as in Step~1 gives
\[
\EE\|\EE[U\mid\mathcal F_{k-1}]\|_2^2
\;\le\;
4\sum_{j=1}^r I(U_j;\mathcal F_{k-1})
\;\le\;
4\sum_{j=1}^r \frac{3(k-1)\epsilon^2}{2r}
=6(k-1)\epsilon^2.
\]
Using the same Cauchy--Schwarz as in Step~1 then yields
\begin{equation}
\label{eq:align-step2}
\EE[\langle a_k,U\rangle]
\;\le\;
\epsilon\sqrt{6r(k-1)}.
\end{equation}
Plugging \cref{eq:align-step2} into \cref{eq:ep-gap-step2} gives
\[
\EE\!\left[V^{*}_1(s_1,\MDP_U) - V^{\pi_{a_k}}_1(s_1,\MDP_U)\right]
\;\ge\;
c_{\mathrm{gap}}\,H\epsilon\left(1-\epsilon\sqrt{\frac{6(k-1)}{r}}\right).
\]
Summing over $k=1,\dots,K$ and applying the same truncation argument as in Step~1
yields a universal constant $c_0>0$ such that
\begin{equation}
\label{eq:bayes-pre-tune-step2}
\BayesReg_K(\mathcal B)
\;\ge\;
c_0\,H\epsilon\,
\min\left\{K,\ \frac{r}{\epsilon^2}\right\}.
\end{equation}

Now choose
\[
\epsilon := \min\left\{\frac{1}{4},\ \sqrt{\frac{r}{K}}\right\}.
\]
If $\sqrt{r/K}\le 1/4$, then $\epsilon=\sqrt{r/K}$ and \cref{eq:bayes-pre-tune-step2}
gives $\BayesReg_K(\mathcal B)\ge c\,H\sqrt{rK}$ for a universal constant $c>0$.
Otherwise $\epsilon=1/4$ and \cref{eq:bayes-pre-tune-step2} gives
$\BayesReg_K(\mathcal B)\ge c\,HK$.  Hence there exists a universal constant
$c>0$ such that
\begin{equation}
\label{eq:bayes-final-step2}
\BayesReg_K(\mathcal B)
\ \ge\ c\,\min\{H\sqrt{rK},\,HK\}.
\end{equation}
In particular, when $K\ge r$ we have $\sqrt{rK}\le K$ and thus
$\BayesReg_K(\mathcal B)\ge c\,H\sqrt{rK}$.

By the Bayes--minimax relation, for every algorithm $\mathcal B$,
\[
\sup_{u\in\{\pm1\}^r}\EE[\Regret_u(K,\mathcal B)]
\ \ge\ \BayesReg_K(\mathcal B),
\]
and therefore the minimax regret satisfies
\[
\Reg^\star_K
:=\inf_{\mathcal B}\sup_{u\in\{\pm1\}^r}\EE[\Regret_u(K,\mathcal B)]
\ \ge\ c\,\min\{H\sqrt{rK},HK\}.
\]

\paragraph{Effective dimension of the construction.}
Lastly we verify that the effective dimension of $\MDP_u$ is $r+2$.  Let $d=|\ExoStateSpace|=2r+1$ and index coordinates by $0$ and $(j,b)$.
Consider the transition feature vector for the pair $(\bar s_a,\text{dummy})$
corresponding to transitioning to the dead state $\bar d_a$:
\[
\phi_{p}(\bar d_a\mid \bar s_a,\text{dummy})
:= \bigl(\mathbf 1\{\SysFun(\bar s_a,\text{dummy},x)=\bar d_a\}\bigr)_{x\in\ExoStateSpace}
\in\RR^d.
\]

By the dynamics, $\phi_{p}(\bar d_a\mid \bar s_a,\text{dummy})(0)=0$ and for
$x=(j,b)$ we have $\phi_{p}(\bar d_a\mid \bar s_a,\text{dummy})(j,b)=\mathbf 1\{b\neq a(j)\}$.

Define $Q\in\RR^{r\times d}$ by
\[
Q_{j,0}:=0,\qquad
Q_{j,(j',b)}:=b\,\mathbf 1\{j=j'\},
\quad j\in[r],\ (j',b)\in[r]\times\{\pm1\},
\]
and denote by $q_j^\top$ its $j$th row.  Then $q_1,\dots,q_r$ are linearly
independent.  Also define
\[
v := \bigl(\indicator_{x\neq 0}\bigr)_{x\in\ExoStateSpace}\in\RR^d,
\]
which is $0$ at coordinate $0$ and $1$ on all $(j,\pm1)$ coordinates.  A direct
calculation shows
\[
\phi_{p}(\bar d_a\mid \bar s_a,\text{dummy})
= \frac12\,v - \frac12\sum_{j=1}^r a(j)\,q_j.
\]
Hence the row span of the transition block contains $\{q_1,\dots,q_r,v\}$, which
are linearly independent, so $\rank(\Fmattrans)\ge r+1$.

Moreover, $\Fmattrans$ also contains the feature vector for staying alive at
$(\bar s_a,\text{dummy})$:
\[
\phi_p(\bar s_a\mid \bar s_a,\text{dummy})
= \mathbf 1_d - \phi_{p}(\bar d_a\mid \bar s_a,\text{dummy}),
\]
and therefore $\mathbf 1_d$ lies in the row span of $\Fmattrans$ as well.  Since
each $q_j$ and $v$ has coordinate $0$ equal to $0$, while $\mathbf 1_d$ has
coordinate $0$ equal to $1$, we have
$\mathbf 1_d\notin \mathrm{span}\{q_1,\dots,q_r,v\}$.  Thus $\rank(\Fmattrans)\ge r+2$.

On the other hand, in this construction every nonzero transition feature vector
is either $\mathbf 1_d$, $\phi_{p}(\bar d_a\mid \bar s_a,\text{dummy})$, or
$\phi_{p}(\bar s_a\mid \bar s_a,\text{dummy})$, all of which lie in
$\mathrm{span}\{\mathbf 1_d,v,q_1,\dots,q_r\}$.  Hence $\rank(\Fmattrans)\le r+2$,
and we conclude
\[
\rank(\Fmattrans)=r+2.
\]
{For rewards, from \cref{eq:rw-H-rare} the reward feature vector for $(\bar s_a,\text{dummy})$ is $\mathbf 1_d$ and for $(\bar d_a,\text{dummy})$ is $\mathbf 0_d$. Since $\mathbf 1_d$ already lies in the row span of $\Fmattrans$, adding the reward block does not increase the rank. Consequently, $\rank(F) = r + 2$.  Thus the effective dimension of this family is $r_{\mathrm{eff}}=r+2$.}

\subsection{Upper Bounds}
\label{app:proofs_observed_upper}

Before presenting the full proof of \cref{thm:pto_regret}, we start by establishing a uniform concentration result over the row space of the joint information matrix. Recall that
\[
\Fmat =
\begin{pmatrix}
    \Fmattrans \\
    \Fmatrew
\end{pmatrix},
\qquad
r := \rank(\Fmat).
\]
We let $\L :=\operatorname{rowspan}(\Fmat)$ denote the row space of the joint information matrix $\Fmat$. Moreover, for a function $u : \ExoStateSpace \to \mathbb{R}$ we write $\norm{u}_\infty = \max_{x \in \ExoStateSpace} |u(x)|$ and $\Var_p(u) := \Var_{\ExoState \sim p}(u(\ExoState))$. For ease of notation, we also set $p = \pvecxi$ to denote the true exogenous distribution. We use $D_k = H(k-1)$ to denote the number of observations of the exogenous state available at the start of episode $k$. Our first lemma establishes concentration of $\widehat p_k$ relative to $p$.

\begin{lemma}
\label{lem:uniform_concentration}
Define $\beta := r\log(1+2HK) +\log(\frac{2K}{\delta})+1.$  With probability at least $1 - \delta$, for all $k = 2, \ldots, K$ and every $u \in \L$ we have that:
\begin{equation*}
\abs{(\widehat p_k - p)^\top u} \leq \sqrt{\frac{2 \Var_p(u) \beta}{D_k}} + \frac{5 \norm{u}_\infty \beta}{D_k}.
\end{equation*}
\end{lemma}
\begin{proof}
Fix an episode $k$ and consider the unit ball $\mathcal{B} = \{u \in \L : \norm{u}_\infty \leq 1\}$.  Since we know that the dimension of $\L$ is equal to $r$, $\mathcal{B}$ admits a $D_k^{-1}$-net $\mathcal N_k$ satisfying $|\mathcal N_k| \leq (1+2D_k)^{r}$ (see Section 4.2 of \cite{vershynin2018high}).

For each $v \in \mathcal N_k$, applying Bernstein's inequality (see Theorem 2.8.4 of \cite{vershynin2018high}) we get with probability at least $1 - 2e^{-\beta}$ that
\begin{equation}
\label{eq:conc_1}
\abs{(\widehat p_k - p)^\top v} \leq \sqrt{\frac{2 \Var_p(v) \beta}{D_k}} + \frac{4 \beta}{3D_k}.
\end{equation}
First note that by the choice of $\beta$ we have that:
\[
2 \sum_{k=2}^K |\mathcal N_k| e^{- \beta} \leq 2K(1+2HK)^{r} e^{-\beta} \leq \delta.
\]
Taking a union bound over $k = 2, \ldots, K$ and each $v \in \mathcal{N}_k$ we have with probability at least $1 - \delta$, \cref{eq:conc_1} holds.

Now, fix a $u \in \mathcal B$ and set $v \in \mathcal N_k$ satisfying $\norm{u-v}_\infty \leq D_k^{-1}$.  Then we have that:
\[
\abs{(\widehat p_k - p)^\top (u - v)} \leq \norm{\widehat p_k - p}_1 \norm{u-v}_\infty \leq \frac{2}{D_k}.
\]
Moreover, the standard deviation satisfies a triangle inequality. In particular,
\[
\sqrt{\Var_p(v)} \leq \sqrt{\Var_p(u)} + \sqrt{\Var_p(v-u)} \leq \sqrt{\Var_p(u)} + \norm{v-u}_\infty \leq
\sqrt{\Var_p(u)} + \frac{1}{D_k}.
\]
The first inequality follows from Minkowski's inequality, and the second inequality from the fact that $\Var_p(v-u) \leq \norm{v-u}_\infty^2$, and the final inequality from construction of $\mathcal N_k$.  Substituting these two bounds into \cref{eq:conc_1} yields that:
\[
\big|(\widehat p_k-p)^\top u\big|
\leq
\sqrt{\frac{2\Var_p(u)\beta}{D_k}}
+
\frac{5\beta}{D_k},
\]
where we use $D_k,\beta\geq1$. Scaling by $\norm{u}_\infty$ proves the required result.
\end{proof}

We next use \Cref{lem:uniform_concentration} to compare the value of an arbitrary deterministic policy $\pi$ under the true and plug-in MDPs.

\begin{lemma}
\label{lem:plugin_value_comparison}
Suppose the event in \Cref{lem:uniform_concentration} holds. Fix an episode $k\geq2$ and a deterministic policy $\pi$. If $k \geq 16 \beta + 1$, then we have that
\[
\abs{ V_1^\pi(s_1,\MDP_p) - V_1^\pi(s_1,\MDP_{\widehat p_k})} \leq 2H\sqrt{\frac{\beta}{k-1}} + \frac{5H\beta}{k-1}.
\]
\end{lemma}
\begin{proof}
Fix a realization of the observed data for which the event in
\Cref{lem:uniform_concentration} holds, and fix an episode $k\geq2$ and
a deterministic policy $\pi$. With the data, $\widehat p_k$, and $\pi$
held fixed, including when $\pi$ was selected using the observed data,
consider an auxiliary trajectory
$(S_h,A_h,\ExoState_h)_{h=1}^H$ under the true model $\MDP_p$, generated
using fresh i.i.d.\ exogenous draws from $p$ independent of the observed
data. All expectations, conditional expectations, and variances in this
proof are with respect to this auxiliary trajectory.

For each stage $h\in[H]$, define:
\begin{align*}
    u_h(x) := \RewFun(S_h,\Action_h,x) + V_{h+1}^\pi\big( \SysFun(S_h,\Action_h,x), \MDP_{\widehat p_k} \big), \quad
    v_h  :=\Var_p(u_h), \quad
    \mathcal V_\pi  = \E\left[\sum_{h=1}^H v_h\right].
\end{align*}

We first show that the function $u_h$ belongs to the row space $\L$. For any stage $h$,
\begin{align*}
u_h(x) & = \RewFun(S_h,\Action_h,x) + \sum_{s' \in \StateSpace} V_{h+1}^\pi(s',\MDP_{\widehat p_k}) \indicator_{\SysFun(S_h,\Action_h,x)=s'}.
\end{align*}
The first term is a row of $\Fmatrew$, while each indicator in the second term is a row of $\Fmattrans$. Hence $u_h$ is a linear combination of rows of the joint information matrix $\Fmat$, and therefore $u_h \in \L$.  Moreover, since the rewards are in $[0,1]$ we have $\norm{u_h}_\infty \leq H - h+1$.

Define the one-step model error $\delta_h := (p - \widehat p_k)^\top u_h$.  Since $u_h\in\L$, \Cref{lem:uniform_concentration} implies
\begin{equation}
\label{eq:plugin_bellman_error}
\abs{\delta_h}
\leq
\sqrt{\frac{2v_h\beta}{D_k}}
+
\frac{5(H-h+1)\beta}{D_k}.
\end{equation}

Next we note that via the Bellman equations, $V_h^\pi(S_h, \MDP_{\widehat p_k}) = \widehat p_k^\top u_h$.  Therefore, 
$p^\top u_h = V_h^\pi(S_h, \MDP_{\widehat p_k})+\delta_h.$
Taking conditional expectation under the true model gives
\[
\E\left[ \RewFun(S_h,\Action_h,\ExoState_h) + V_{h+1}^\pi(S_{h+1},\MDP_{\widehat p_k}) \mid S_h,\Action_h \right] = V_h^\pi(S_h,\MDP_{\widehat p_k})+\delta_h.
\]
Taking expectations and summing over $h=1,\ldots,H$ yields
\[
V_1^\pi(s_1,\MDP_p)
=
V_1^\pi(s_1,\MDP_{\widehat p_k})
+
\E\left[
\sum_{h=1}^H \delta_h
\right],
\]
where the value-function terms telescope and we use the fact that $V_{H+1}^\pi(\cdot,\MDP_{\widehat p_k}) = 0.$
Hence,
\[
\abs{
V_1^\pi(s_1,\MDP_p)
-
V_1^\pi(s_1,\MDP_{\widehat p_k})
}
\leq
\E\left[
\sum_{h=1}^H \abs{\delta_h}
\right].
\]

Applying \cref{eq:plugin_bellman_error}, we obtain
\begin{align*}
\E\left[
\sum_{h=1}^H \abs{\delta_h}
\right]
&\leq
\E\left[
\sum_{h=1}^H
\sqrt{\frac{2v_h\beta}{D_k}}
\right]
+
\frac{5\beta}{D_k}
\sum_{h=1}^H(H-h+1)
\\
&\leq
\sqrt{\frac{2H\beta\mathcal V_\pi}{D_k}}
+
\frac{5\beta}{D_k}
\sum_{h=1}^H(H-h+1).
\end{align*}
where the second inequality follows from Cauchy--Schwarz and Jensen's inequality and the definition of $\mathcal V_\pi$.  Finally,
$
\sum_{h=1}^H(H-h+1)
=
\frac{H(H+1)}{2}
\leq H^2.$  Combining the preceding bounds gives
\begin{equation}
\label{eq:mart_conc_2}
\abs{V_1^\pi(s_1,\MDP_p)
-
V_1^\pi(s_1,\MDP_{\widehat p_k})
}
\leq
\sqrt{\frac{2H\beta \mathcal V_\pi}{D_k}}
+
\frac{5H^2\beta}{D_k},
\end{equation}

It remains to control the variance term $\mathcal V_\pi$. Define $\epsilon_h := u_h(\ExoState_h) - p^\top u_h$.  Then $(\epsilon_h)_{h=1}^H$ is a martingale difference sequence with  $\E[\epsilon_h^2\mid S_h,\Action_h]=v_h,$ and therefore $\mathcal V_\pi = \E\left[\left(\sum_{h=1}^H\epsilon_h\right)^2\right].$
Moreover,
\[
\RewFun(S_h,\Action_h,\ExoState_h) = V_h^\pi(S_h,\MDP_{\widehat p_k}) - V_{h+1}^\pi(S_{h+1},\MDP_{\widehat p_k}) + \delta_h+\epsilon_h.
\]
Summing over $h$ gives
\[
\sum_{h=1}^H\epsilon_h = \sum_{h=1}^H \RewFun(S_h,\Action_h,\ExoState_h) - V_1^\pi(s_1,\MDP_{\widehat p_k}) - \sum_{h=1}^H\delta_h. 
\]

Since $V_1^\pi(s_1,\MDP_{\widehat p_k})$ is deterministic conditional on the observations collected before episode $k$, it does not affect the variance. Therefore,
\begin{align*}
\mathcal V_\pi &= \Var\left( \sum_{h=1}^H \epsilon_h \right) = \Var\left( \sum_{h=1}^H \RewFun(S_h,\Action_h,\ExoState_h) - \sum_{h=1}^H\delta_h \right)\\
& \leq 2\Var\left( \sum_{h=1}^H \RewFun(S_h,\Action_h,\ExoState_h) \right) + 2\Var\left( \sum_{h=1}^H\delta_h \right).
\end{align*}

Since the cumulative reward lies in $[0,H]$, its variance is at most $H^2/4$. Moreover,
\[
\Var\left(\sum_{h=1}^H\delta_h\right) \leq \E\left[\left(\sum_{h=1}^H\delta_h\right)^2\right] \leq H\E\left[\sum_{h=1}^H\delta_h^2\right],
\]
where the last inequality follows from Cauchy-Schwarz. Consequently,
\[
\mathcal V_\pi \leq \frac{H^2}{2} + 2H\E\left[\sum_{h=1}^H\delta_h^2\right].
\]

Using the preceding bound on $\delta_h$ from \cref{eq:plugin_bellman_error} and $(a+b)^2\leq2a^2+2b^2$,
\[
\delta_h^2 \leq \frac{4v_h\beta}{D_k} + \frac{50(H-h+1)^2\beta^2}{D_k^2}.
\]
Therefore, after summing up the previous inequality and using the definition of $\mathcal V_\pi$:
\[
\mathcal V_\pi \leq \frac{H^2}{2} + \frac{8H\beta}{D_k}\mathcal V_\pi + \frac{100H^4\beta^2}{D_k^2}.
\]
Now, we assume that $k \geq 16\beta + 1$, or equivalently that $D_k \geq 16 H \beta$. From here we have $\frac{8H\beta}{D_k} \leq \frac{1}{2}$ and $\frac{100H^4 \beta^2}{D_k^2} \leq \frac{100}{256}H^2$.
Rearranging this expression gives $\mathcal V_\pi \leq 2H^2$.  Substituting this into \cref{eq:mart_conc_2} yields
\[
\abs{V_1^\pi(s_1,\MDP_p) - V_1^\pi(s_1,\MDP_{\widehat p_k})} \leq \sqrt{\frac{4H^3\beta}{D_k}} + \frac{5H^2\beta}{D_k}.
\]
Finally, since $D_k=H(k-1)$,
\[
\abs{V_1^\pi(s_1,\MDP_p) - V_1^\pi(s_1,\MDP_{\widehat p_k})} \leq 2H\sqrt{\frac{\beta}{k-1}} + \frac{5H\beta}{k-1},
\]
which proves the result.
\end{proof}

With the previous two lemmas in hand, we are now ready to show the main theorem.

\ObsUpperBound*

\begin{proof}
Condition on the event in \Cref{lem:uniform_concentration}, let $\pi^\star$ be a deterministic Markov optimal policy for
the true model $\MDP_p$. By the optimality of $\pi_k$ in the plug-in model $\MDP_{\widehat p_k}$, we have $ V_1^{\pi^\star}(s_1,\MDP_{\widehat p_k}) \leq V_1^{\pi_k}(s_1,\MDP_{\widehat p_k}).$
Therefore,
\begin{align*}
V_1^{\pi^\star}(s_1,\MDP_p)
-
V_1^{\pi_k}(s_1,\MDP_p)
&=
V_1^\star(s_1,\MDP_p)
-
V_1^{\pi^\star}(s_1,\MDP_{\widehat p_k})\\
&\qquad+
V_1^{\pi^\star}(s_1,\MDP_{\widehat p_k})
-
V_1^{\pi_k}(s_1,\MDP_{\widehat p_k})\\
&\qquad+
V_1^{\pi_k}(s_1,\MDP_{\widehat p_k})
-
V_1^{\pi_k}(s_1,\MDP_p)\\
& \leq
\abs{V_1^{\pi^\star}(s_1,\MDP_p)
      -V_1^{\pi^\star}(s_1,\MDP_{\widehat p_k})}\\
&\qquad+
\abs{V_1^{\pi_k}(s_1,\MDP_{\widehat p_k})-V_1^{\pi_k}(s_1,\MDP_p)}.
\end{align*}

Suppose first that $k-1\geq16\beta$. Applying \Cref{lem:plugin_value_comparison} to $\pi^\star$ and $\pi_k$ gives
\[
V_1^\star(s_1,\MDP_p)-V_1^{\pi_k}(s_1,\MDP_p)\leq4H\sqrt{\frac{\beta}{k-1}}+\frac{10H\beta}{k-1}.
\]
Since $k-1\geq16\beta$, we have
\(
\sqrt{\frac{\beta}{k-1}}\leq\frac14,
\)
and therefore
\(
\frac{\beta}{k-1}\leq\frac14\sqrt{\frac{\beta}{k-1}}.
\)
Consequently,
\[
V_1^\star(s_1,\MDP_p)-V_1^{\pi_k}(s_1,\MDP_p)\leq\frac{13H}{2}\sqrt{\frac{\beta}{k-1}}.
\]

When $1\leq k-1<16\beta$, the per-episode regret is trivially bounded by $H$. Since in this case $\sqrt{\frac{\beta}{k-1}}>\frac14,$ we again have
\(
H \leq \frac{13H}{2} \sqrt{\frac{\beta}{k-1}}. \)
Thus, for every $k\geq2$,
\[
V_1^\star(s_1,\MDP_p) - V_1^{\pi_k}(s_1,\MDP_p) \leq \frac{13H}{2} \sqrt{\frac{\beta}{k-1}}.
\]
The first episode contributes at most $H$ to the regret, so 
\begin{align*}
\Regret(K)
&\leq H + \frac{13H\sqrt{\beta}}{2} \sum_{j=1}^{K-1}\frac{1}{\sqrt{j}}\\
&\leq H + 13H\sqrt{\beta (K-1)}\\
&\leq 14H\sqrt{\beta K}.
\end{align*}

Finally, recalling that
\(
\beta = r\log(1+2HK) + \log\left(\frac{2K}{\delta}\right) + 1
\)
we conclude that
\[
\Regret(K) = O\left( H\sqrt{rK\log(1+HK) + K\log\left(\frac{2K}{\delta}\right)} \right).
\]
This proves the claimed regret bound.
\end{proof}

\begin{figure}[!t]
\caption{Here we plot the total cost function $\TotalCost_1^\BaseStock(s_1)$ as we vary the base-stock value $\BaseStock$ in Scenario II.
The $x$-axis denotes the base-stock value $\BaseStock \in [0,10]$ and the $y$-axis $\TotalCost_1^\BaseStock(s_1)$.  }
\label{fig:sub1}
\centering
\includegraphics[width=.5\textwidth]{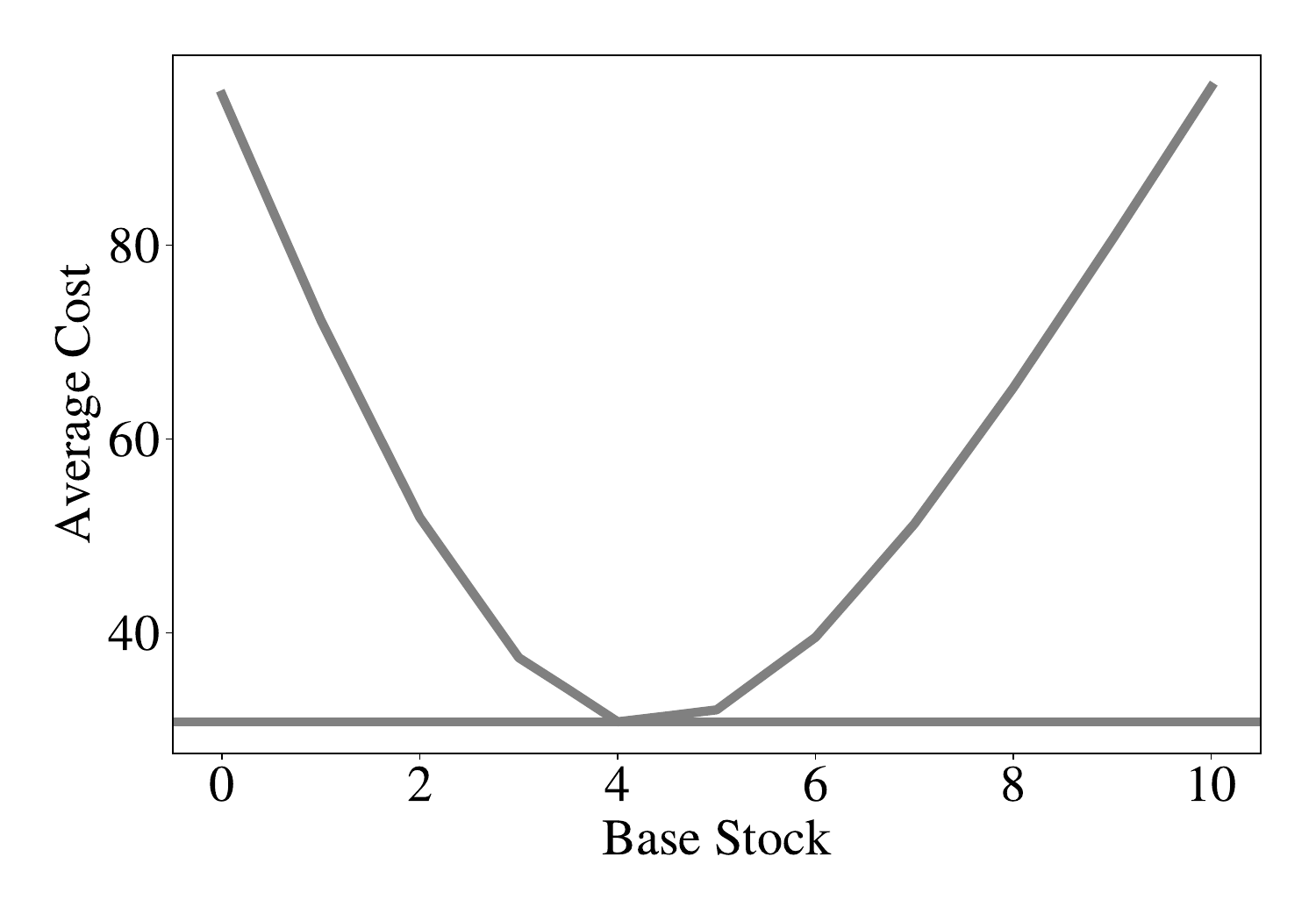}
\end{figure}

\section{Case Study on Inventory Control}\label{app:inventory_control}

Here we provide supplemental information on our case study on inventory control with lost sales and positive lead time.  In \cref{app:cor:inventory_control} we present the regret guarantees for \UV and Plug-In. In \cref{app:baseline_policy_convex} we present base-stock algorithms, and in \cref{app:online_base_stock} the regret-guarantees for our online base-stock policy benchmark algorithm.  Further simulation details are in \cref{app:experiment_details}.

\subsection{Regret Guarantees for \UV and Plug-In on Inventory Control}\label{app:cor:inventory_control}

{
Let $\Fmat$ be the joint information matrix of a single-product stochastic inventory control problem with lead time $L > 0$, where the demand distribution has support $\{0,1,\dots,d-1\}$. We show that $\rank(\Fmat)=d$, and then apply \cref{thm:linear_mixture_mdp_low_rank}.

Since the transition of the outstanding orders $(\Order_{h-L},\dots,\Order_{h-1})$ is a shift operation apart from the arriving order $\Order_{h-L}$, it suffices to consider the inventory component of the state. Fix a state whose inventory level is zero and whose arriving order is $d-1$; for instance, under the ordering $s=(\Inventory_h,\Order_{h-L},\dots,\Order_{h-1})$, take $s=(0,d-1,0,\dots,0).$  The current order $\Order_h$ can be chosen arbitrarily. For each demand realization $\Demand_h=j\in\{0,\dots,d-1\}$, the next inventory level is $\Inventory_{h+1}=d-1-j.$  Therefore, restricting the transition block of the joint information matrix to this state-action pair, the entry indexed by next-inventory level $i\in\{0,\dots,d-1\}$ and demand $j\in\{0,\dots,d-1\}$ is
\begin{align*}
\Fmat[s,a]_{i,j} =
\begin{cases}
1, & \text{if } i=d-1-j,\\
0, & \text{otherwise}.
\end{cases}
\end{align*}
Thus $\Fmat[s,a]$ is a $d\times d$ permutation matrix and consequently has rank $d$. Since $\Fmat[s,a]$ is a submatrix of the joint information matrix $\Fmat$, we have $\rank(\Fmat)\ge d$. On the other hand, $\Fmat$ has exactly $d$ columns, one for each exogenous demand value, so $\rank(\Fmat)\le d$. Hence $\rank(\Fmat)=d$.

Therefore the joint information matrix has full column rank, so the effective dimension is $r=d$ and there is no further rank reduction in this inventory control model. Applying \cref{thm:pto_regret,thm:linear_mixture_mdp_low_rank} therefore yields regret guarantees that scale with $d$, the cardinality of the support of the exogenous demand distribution.
}

\subsection{Base-Stock Policies}\label{app:baseline_policy_convex}

{Let $\TotalCost_1^\BaseStock$ denote the $H$-stage total cost function for the base-stock policy $\pi_h^\BaseStock$ starting from an initial state of no inventory, i.e., $s_1 = (\Inventory_1,\Order_{1-L},\dots,\Order_{0}) = (0,0,\dots,0)$. Formally, we have:}
\begin{equation}
\label{eq:value_base_stock}
\TotalCost_1^{\BaseStock} = \EE_{\ExoState_1,\dots ,\ExoState_H, \pi_\BaseStock}\left[\sum_{h=1}^H \CostFun(S_h, \Action_h, \ExoState_h)\mid s_1 = (0,0,\dots,0)\right],
\end{equation}
where $S_h = (\Inventory_h, \Order_{h-1}, \ldots, \Order_{h-L})$ and $\Action_h = \Order_h$.  We start off by noting that by existing results, the total cost function $\TotalCost_1^\BaseStock$ is convex in the base-stock level $\BaseStock$.

\begin{lemma}[Theorem 8 of \citet{janakiraman2004lost}]
\label{lem:basestock_H_convexity}
For any distribution of the demand $\PXi$, the total cost function $\TotalCost_1^{\BaseStock}$ for a base-stock policy $\pi^\BaseStock$ is convex in $\BaseStock$.
\end{lemma}

\begin{algorithm}[!t]
\caption{One-dimensional stochastic convex bandit algorithm from~\citet{agarwal2011stochastic} applied to online inventory control}
\label{alg:convex_opt}
\begin{algorithmic}[1]
\Require Total number of episodes $K$
\State Set $l_1 \gets 0$ and $r_1 \gets d$ and let $\sigma \gets H/2$
\For{epoch $\tau = 1,2,\dots$}
\For{round $i = 1,2,\dots$}
\State Let $\gamma_i \gets 2^{-i}$
\State For each $\BaseStock \in \{\BaseStock_l, \BaseStock_c, \BaseStock_r\}$, run the base-stock policy for $\frac{2\log(K)}{\gamma_i^2}$ episodes
\If{$\max\{LB_{\gamma_i}(\BaseStock_l), LB_{\gamma_i}(\BaseStock_r)\} \geq \max\{UB_{\gamma_i}(\BaseStock_l), UB_{\gamma_i}(\BaseStock_r)\}$}
\If{$LB_{\gamma_i}(\BaseStock_l) \ge LB_{\gamma_i}(\BaseStock_r)$}
\State Let $l_{\tau+1} \gets \BaseStock_l$ and $r_{\tau+1} \gets r_\tau$
\Else
\State Let $l_{\tau+1} \gets l_\tau$ and $r_{\tau+1} \gets \BaseStock_r$
\EndIf

\ElsIf{$\max\{LB_{\gamma_i}(\BaseStock_l), LB_{\gamma_i}(\BaseStock_r)\}
\ge UB_{\gamma_i}(\BaseStock_c) + \gamma_i$}

\If{$LB_{\gamma_i}(\BaseStock_l) \ge LB_{\gamma_i}(\BaseStock_r)$}
\State Let $l_{\tau+1} \gets \BaseStock_l$ and  $r_{\tau+1} \gets r_\tau$
\Else
\State Let $l_{\tau+1} \gets l_\tau$ and $r_{\tau+1} \gets \BaseStock_r$
\EndIf
\State Continue to epoch $\tau + 1$
\EndIf
\EndFor
\EndFor
\State \Return $(\pi^k)_{k \in [K]}$
\end{algorithmic}
\end{algorithm}
\subsection{Online Base-Stock Algorithm}\label{app:online_base_stock}


Since we know that $\TotalCost_1^\BaseStock$ is convex in the base-stock level $\BaseStock$, we can use existing work on establishing algorithms for stochastic online convex optimization with bandit feedback from~\citet{agarwal2011stochastic}.  This directly yields an algorithm for our scenario by using the following observation.  Once the policy $\pi^\BaseStock$ is fixed, each $H$-stage evaluation of the policy in a given episode can be treated as a single sample for its expected $H$-stage total cost $\TotalCost_1^\BaseStock$.

Algorithm 1 in \citet{agarwal_reinforcement_nodate} (restated in \Cref{alg:convex_opt} specified for our context) is epoch based where the feasible region of optimal base-stock parameters $[0,U]$, where $U$ corresponds to the maximum demand, is refined over time.  {(Note that here it suffices to take $U = d$).} In each epoch, the algorithm aims to discard a portion of the feasible region determined to contain provably suboptimal points.  To do so, the algorithm estimates the performance of base-stock policies at three different points within the working feasible region.  At the end of an epoch, the feasible region is reduced whenever the confidence intervals are sufficiently separated to certify that part of the current region is suboptimal.  In the algorithm we let $LB_{\gamma_i}(\BaseStock)$ and $UB_{\gamma_i}(\BaseStock)$ denote lower and upper confidence bound estimates on the performance of base-stock policy $\pi^\BaseStock$, defined as $\hat{\TotalCost}_1^{\BaseStock} \pm \gamma_i$, where $\hat{\TotalCost}_1^{\BaseStock}$ is the empirical average cost of base-stock policy $\BaseStock$ evaluated over $2 \log(K) / \gamma_i^2$ trajectories.

\begin{corollary}
\label{cor:base_stock_algo}
Let $\BaseStock^*$ denote the best-performing base-stock level, $
\BaseStock^* = \argmin_{\BaseStock} \TotalCost_1^{\BaseStock}.$
Then applying Theorem 1 of \citet{agarwal2011stochastic} establishes that \Cref{alg:convex_opt} yields with probability at least $1 - 1/K$ that:
\begin{align*}
\sum_{k=1}^K \TotalCost_1^{\BaseStock^k} - \TotalCost_1^{\BaseStock^*} \leq 108H\sqrt{K \log(K)} \log_{4/3}\left(\frac{K}{8 \log(K)}\right) = \tilde{O}(H \sqrt{K}).
\end{align*}
\end{corollary}
\noindent Note that the factor $H$ arises since \citet{agarwal2011stochastic} assumes the costs are in $[0,1]$ whereas $\TotalCost_1^\BaseStock \in [0,H]$.  This also defines regret relative to the performance of the best performing base-stock policy.  However, as we will see in \cref{sec:experiments} there are scenarios where this gap leads to $O(K)$ gap in total cost relative to the optimal policy.

\begin{table}[!t]
\caption{Configuration parameters for the inventory control scenarios in
\cref{fig:inventory_control} and \cref{tab:performance}.}
\label{tab:inventory_scenarios}
\centering

{
\begin{tabular}{|c|c|c|c|c|}
\hline
\textbf{} & \textbf{Scenario I} & \textbf{Scenario II} & \textbf{Scenario III} & \textbf{Scenario IV} \\ \hline
Horizon ($H$) & $10$ & $1$ & $1$ & $10$ \\ \hline
Lead Time ($L$) & $2$ & $0$ & $0$ & $0$ \\ \hline
Holding Cost ($\HoldingCost$) & $3$ & $3$ & $3$ & $3$ \\ \hline
Lost Sales ($\LostSales$) & $8$ & $8$ & $8$ & $8$ \\ \hline
Max Inventory & $8$ & $10$ & $20$ & $10$ \\ \hline
Demand Process & time-homogeneous & time-homogeneous & time-homogeneous & time-homogeneous \\ \hline
\end{tabular}
}
\smallskip
{
\begin{tabular}{|c|c|c|c|}
\hline
\textbf{} & \textbf{Scenario V} & \textbf{Scenario VI} & \textbf{Scenario VII} \\ \hline
Horizon ($H$) & $6$ & $3$ & $3$ \\ \hline
Lead Time ($L$) & $0$ & $1$ & $1$ \\ \hline
Holding Cost ($\HoldingCost$) & $1$ & $10$ & $10$ \\ \hline
Lost Sales ($\LostSales$) & $20$ & $10$ & $10$ \\ \hline
Max Inventory & $20$ & $10$ & $10$ \\ \hline
Demand Process & time-inhomogeneous & Markovian & Markovian \\ \hline
Additional Parameters
& $\epsilon=0.05$
& $\rho = 0.55$
& $\rho = 0.9$ \\ \hline
\end{tabular}
}
\end{table}

{\subsection{Description of Scenarios}
\label{app:inventory_scenario_descriptions}

We consider seven inventory scenarios designed to evaluate the algorithms under three different demand settings: stationary i.i.d. demand, time-inhomogeneous demand, and temporally correlated demand. The parameter values for all scenarios are summarized in \cref{tab:inventory_scenarios}.

\paragraph{Stationary demand (Scenarios I--IV).}
We first consider time-homogeneous demand, where demand in each period is drawn independently from a Poisson distribution truncated to the finite support $\{0,\ldots,d-1\}$, where $d-1$ is the maximum order size.  Scenario I has mean demand $3$, while Scenarios II--IV have mean demand $7$. Across these scenarios, we vary the lead time, horizon, and maximum order size to evaluate the algorithms under different inventory-system configurations.

\paragraph{Time-inhomogeneous demand (Scenario V).}
We next consider an independent but time-varying demand process that alternates between low- and high-demand periods. Let $\PXi^h$ denote the demand distribution in period $h$. During a low-demand period,
\[
\PXi^h(x)
=
\begin{cases}
1-\epsilon, & x=0,\\
\displaystyle \frac{\epsilon}{d}, & x\in\{1,\ldots,d-1\},
\end{cases}
\]
whereas during a high-demand period,
\[
\PXi^h(x)
=
\begin{cases}
1-\epsilon, & x=d-1,\\
\displaystyle \frac{\epsilon}{d-1}, & x\in\{0,\ldots,d-2\}.
\end{cases}
\]
We alternate deterministically between the low- and high-demand distributions across periods and set $\epsilon=0.05$. This scenario evaluates the effect of explicitly accounting for time-inhomogeneity in the exogenous process.

\paragraph{Correlated demand (Scenarios VI--VII).}
Finally, we evaluate robustness to violations of the independence assumption by considering a Markovian demand process. Let $p_{\rm low}$ and $p_{\rm high}$ denote Poisson distributions with means $\mu_{\rm low}=2$ and $\mu_{\rm high}=13$, respectively, truncated and renormalized to the support $\{0,\ldots,d-1\}$. The initial demand distribution is
\[
\PXi(x)
=
\frac{1}{2}p_{\rm low}(x)
+
\frac{1}{2}p_{\rm high}(x).
\]
For subsequent periods, demand follows the transition kernel
\[
\PXi(x' \mid x)
=
\begin{cases}
\rho\, p_{\rm low}(x') + (1-\rho)\,p_{\rm high}(x'),
& x \le \lfloor d/2\rfloor,\\[3pt]
(1-\rho)\,p_{\rm low}(x') + \rho\,p_{\rm high}(x'),
& x > \lfloor d/2\rfloor.
\end{cases}
\]
Thus, following a relatively low demand realization, the next-period distribution places weight $\rho$ on the low-demand distribution, whereas following a relatively high demand realization it places weight $\rho$ on the high-demand distribution. We refer to $\rho$ as the \emph{stickiness parameter}. Scenarios VI and VII use $\rho=0.55$ and $\rho=0.9$, respectively, to compare moderate and strong temporal dependence.
}


\section{Additional environment: Airline Revenue Management.}
\label{app:arm_simulations}

In addition to the inventory control problem we also evaluate the algorithms on the airline revenue management problem \citep{littlewood1972,sinclair2023hindsight}. There are $K\in\mathbb{N}$ resources, and each resource $k\in[K]$ has an initial capacity $B_k$. Customers are segmented into $M\in \mathbb{N}$ types. Customers of type $\exostate_t=j\in[M]$ request $A_j \in \RR_+^{K}$ resources and yield a revenue of $r_j$. If the algorithm decided to accept customers of type $\exostate_t$, the relevant resources are consumed and revenue earned; otherwise the resource and revenue stay the same. We note that a no-customer can be modeled via no resource consumption and a revenue of zero.  The goal of the decision-maker is to maximize the expected revenue.

\paragraph{Exo-MDP formulation.}
This is modeled as an Exo-MDP where
\[
\StateSpace = [0,B_1]\times\cdots\times[0,B_K],
\qquad
\ExoStateSpace = [M],
\qquad
\ActionSpace = \{0,1\}^M.
\]
The endogenous state $\state_t$ corresponds to the remaining capacity of the $K$ different resources, exogenous space $\ExoStateSpace$ correspond to the customer type, and $\ActionSpace$ to the accept/reject decisions for each of the customer types (with $a_{\exostate_t}$ the decision applied to the realized type at time $t$). The reward is defined by 
\[
\RewFun(s_t,a,\exostate_t)
=
\begin{cases}
r_{\exostate_t}, & a_{\exostate_t}=1 \ \text{and}\ s_t - A_{\exostate_t}\ge 0,\\
0, & \text{otherwise},
\end{cases}
\]
where $s_t-A_{\exostate_t}\ge 0$ is satisfied componentwise. The transition update accounts for consumed resources if a request is accepted:
\[
\SysFun(s_t,a,\exostate_t)
=
\begin{cases}
s_t - A_{\exostate_t}, & a_{\exostate_t}=1 \ \text{and}\ s_t - A_{\exostate_t}\geq 0,\\
s_t, & \text{otherwise}.
\end{cases}
\]

\paragraph{Simulation Parameters.} We consider two different parameter set-ups.

\noindent \emph{Scenario I: Baseline instance.}
The first set-up is from \citet{sinclair2023hindsight} with $K=3$ resources and $M=3$ customer types (where one type corresponds to a ``no arrival’’ customer) over $H = 4$ steps.
The starting capacity is $B=[8,4,4]$ and arrivals are i.i.d.\ with exogenous distribution $\pvecxi=(1/3,1/3,1/3)$. Customer type $1$ requests resources $A_1=[2,3,2]$ and yields revenue $r_1=1$, while customer type $2$ requests $A_2=[3,0,1]$ and yields revenue $r_2=2$. The remaining type corresponds to ``no customer’’ and has zero requests and zero revenue.

\smallskip
\noindent \emph{Scenario II: Peak-shift instance.}
To model nonstationary exogenous states, we also consider a time-varying arrival distribution over a horizon of length $H=6$, where premium customers become more likely later in the episode. Specifically, we consider $K = 2$ resources with $M = 4$ customer types (where again, one type corresponds to a ``no arrival'' customer).  The arrival probability vector varies with time according to
\[
\pvecxih = \begin{bmatrix}
0.45 & 0.45 & 0.00 & 0.10 \\
0.45 & 0.45 & 0.00 & 0.10 \\
0.40 & 0.50 & 0.00 & 0.10 \\
0.30 & 0.55 & 0.00 & 0.15 \\
0.10 & 0.10 & 0.60 & 0.20 \\
0.05 & 0.05 & 0.75 & 0.15
\end{bmatrix},
\]
over $M=4$ types (including a ``no arrival’’ type) where each row corresponds to different steps $h$. Revenues are $r=[2.0,3.5,6.0]$ for the three purchasing types, and resource consumption is encoded by $A_1 = [1,0]$, $A_2 = [1,1]$ and $A_3 = [0,1]$, so that the first type uses only resource $1$, the second uses both resources, and the third (premium) uses only resource $2$, with starting capacity $B=[8,6]$.

\paragraph{Discussion.}
In \cref{tab:arm_performance} we compare the performance of our algorithms with the $Q$-Learning algorithm of \citet{jin2018q} at the final episode $K = 500$. We highlight two main insights.
First, the algorithms that exploit the exogenous structure substantially outperform \QLearning in both scenarios. The two \PlugIn approaches achieve near-optimal performance, while the linear algorithms \UCRL and \HFUCRL also perform well despite being designed for the more general setting in which the exogenous states are unobserved. The additional improvement of the \PlugIn algorithms is consistent with the statistical advantage of directly observing the exogenous customer arrivals.
Second, the \PlugIn methods appear relatively insensitive to whether the model is specified as homogeneous or non-homogeneous. In the non-homogeneous Scenario~II, the homogeneous \PlugIn algorithm remains near-optimal, while in the homogeneous Scenario~I, \PlugInNH performs comparably to \PlugIn. Taken together, these results suggest that the empirical cost of allowing for time-varying exogenous distributions when the environment is homogeneous, or imposing homogeneity when the variation is modest, is relatively small in these instances.  This appears to be driven in part by the fact that the optimal stationary policy in Scenario II is close in value to the optimal non-stationary policy.

\begin{table}[!tb]
{
\caption{Total observed reward at the final episode $K = 500$ for policy $\pi_K$ with a 95\% confidence interval. In parentheses we show the relative performance to the total expected reward of the optimal policy, $(V^*_1 - \widehat{R}^{\pi_K}) / V_1^*$, where $\widehat{R}^{\pi_K}$ denotes the empirical average final-episode reward. Some empirical final-episode averages exceed the optimal expected value due to Monte Carlo sampling error.}
\label{tab:arm_performance}
\setlength\tabcolsep{0pt}
\smallskip
\begin{tabular*}{\columnwidth}{@{\extracolsep{\fill}}lcc}
\toprule
Algorithm  & Scenario I & Scenario II\\
\midrule
Optimal Policy ($V_1^*$) & $3.22$ & $18.95$  \\
\midrule
\PlugIn & $3.31 \pm 0.18,(0\%)$ & $19.02 \pm 0.54,(0\%)$ \\
\PlugInNH & $3.33 \pm 0.19,(0\%)$ & $19.43 \pm 0.47,(0\%)$ \\
\midrule
\UCRL & $3.19 \pm 0.18,(1\%)$ & $19.27 \pm 0.49,(0\%)$ \\
\HFUCRL & $3.15 \pm 0.17,(2\%)$ & $18.84 \pm 0.52,(1\%)$ \\
\midrule
\QLearning & $3.01 \pm 0.17,(7\%)$ & $15.59 \pm 0.62,(18\%)$ \\
\Random & $1.69 \pm 0.17,(48\%)$ & $9.38 \pm 0.61,(50\%)$ \\
\bottomrule
\end{tabular*}
}
\end{table}

\section{Simulation Details}
\label{app:experiment_details}

\paragraph{Computing infrastructure.} The experiments were conducted on a personal computer with an Apple M2, 8-core processor and 16.0GB of RAM. No GPUs were needed for the experiments.  Each simulation took approximately two hours.

\paragraph{Experiment setup.} {Each experiment was run with $250$ iterations where the various plots and metrics were computed with respect to the mean of the various quantities.
Hyperparameters for the confidence intervals (the constant scaling in front of the dominating terms) were tuned via grid search in $10^{-i}$ for $i \in \{-5, \ldots, 0\}.$}

\end{APPENDICES}

\end{document}